%% file: 00-main.tex
\documentclass[11pt, a4paper, logo, copyright]{googledeepmind}

\usepackage[authoryear, compress, round]{natbib}
\usepackage{wrapfig}
\usepackage{xspace}
\usepackage{epigraph}
\usepackage{hyperref}
\usepackage{cleveref}
\usepackage{indentfirst}
\usepackage{enumitem}
\usepackage{tabularx}
\usepackage{makecell}
\setlist[itemize]{topsep=0pt}

\title{An Approach to Technical AGI Safety and Security}

\correspondingauthor{rohinmshah@google.com}

\renewcommand{\today}{}

\author[1]{Rohin Shah}
\author[*,1]{Alex Irpan}
\author[*,1]{Alexander Matt Turner}
\author[*,1]{Anna Wang}
\author[*,1]{Arthur Conmy}
\author[*,1]{David Lindner}
\author[*,1]{Jonah Brown-Cohen}
\author[*,1]{Lewis Ho}
\author[*,1]{Neel Nanda}
\author[*,1]{Raluca Ada Popa}
\author[*,1]{Rishub Jain}
\author[*,1]{Rory Greig}
\author[*,1]{Samuel Albanie}
\author[*,1]{Scott Emmons}
\author[*,1]{Sebastian Farquhar}
\author[*,1]{Sébastien Krier}
\author[*,1]{Senthooran Rajamanoharan}
\author[*,1]{Sophie Bridgers}
\author[*,1]{Tobi Ijitoye}
\author[*,1]{Tom Everitt}
\author[*,1]{Victoria Krakovna}
\author[*,1]{Vikrant Varma}
\author[*,2]{Vladimir Mikulik}
\author[*,1]{Zachary Kenton}
\author[]{Dave Orr}
\author[]{Shane Legg}
\author[]{Noah Goodman}
\author[]{Allan Dafoe}
\author[]{Four Flynn}
\author[]{Anca Dragan}

\affil[1]{Google DeepMind}
\affil[2]{Work done while at Google DeepMind}
\affil[*]{Core contributor, alphabetical order}

\usepackage{tcolorbox}
\newtcolorbox{formattedquote}{
    colback=blue!3!white,
    colframe=blue!20!white,
    fontupper=\footnotesize,
    boxsep=-2pt 
}

\begin{abstract}
    Artificial General Intelligence (AGI) promises transformative benefits but also presents significant risks. We develop an approach to address the risk of harms consequential enough to significantly harm humanity. We identify four areas of risk: misuse, misalignment, mistakes, and structural risks. Of these, we focus on technical approaches to misuse and misalignment. For misuse, our strategy aims to prevent threat actors from accessing dangerous capabilities, by proactively identifying dangerous capabilities, and implementing robust security, access restrictions, monitoring, and model safety mitigations. To address misalignment, we outline two lines of defense. First, model-level mitigations such as amplified oversight and robust training can help to build an aligned model. Second, system-level security measures such as monitoring and access control can mitigate harm even if the model is misaligned. Techniques from interpretability, uncertainty estimation, and safer design patterns can enhance the effectiveness of these mitigations. Finally, we briefly outline how these ingredients could be combined to produce safety cases for AGI systems.
\end{abstract}

\begin{document}

\maketitle

\newpage

\input{00-extended-abstract}

\newpage

\tableofcontents

\newpage

\input{01-introduction}

\input{02-evidence-dilemma}

\input{03-assumptions/intro}
\input{03-assumptions/paradigm}
\input{03-assumptions/no-human-ceiling}
\input{03-assumptions/timelines}
\input{03-assumptions/acceleration}
\input{03-assumptions/continuity}
\input{03-assumptions/benefits}

\input{04-risk-areas/intro}
\input{04-risk-areas/misuse-risks}
\input{04-risk-areas/misalignment-risks}
\input{04-risk-areas/mistakes}
\input{04-risk-areas/structural-risks}

\input{05-addressing-misuse}

\input{06-addressing-misalignment/intro}
\input{06-addressing-misalignment/amplified-oversight}

\input{06-addressing-misalignment/guiding-model-behavior}

\input{06-addressing-misalignment/robust-training}

\input{06-addressing-misalignment/hardening}

\input{06-addressing-misalignment/safer-design-patterns}
\input{06-addressing-misalignment/interpretability}

\input{06-addressing-misalignment/stress-tests}
\input{06-addressing-misalignment/safety-cases}
\input{06-addressing-misalignment/discussion}

\input{07-conclusion}

\section*{Acknowledgements}

Thanks to Been Kim, Kevin Robinson, Lewis Smith, Mark Kurzeja, Mary Phuong, Michael Dennis, Rif A. Saurous, Rory Greig, Sarah Cogan, Verena Rieser and many other researchers at Google DeepMind for comments on a draft of this paper.

\bibliography{refs}

\end{document}

%% file: 00-extended-abstract.tex
\section*{Extended Abstract}

%

AI, and particularly AGI, will be a transformative technology. As with any transformative technology, AGI will provide significant benefits while posing significant risks. This includes risks of \textbf{severe harm}: incidents consequential enough to significantly harm humanity. This paper outlines our approach to building AGI that avoids severe harm.\footnote{Severe risks require a radically different mitigation approach than most other risks, as we cannot wait to observe them happening in the wild, before deciding how to mitigate them. For this reason, we limit our scope in this paper to severe risks, even though other risks are also important.}

Since AGI safety research is advancing quickly, our approach should be taken as exploratory. We expect it to evolve in tandem with the AI ecosystem to incorporate new ideas and evidence.

Severe harms necessarily require a precautionary approach, subjecting them to an \textit{evidence dilemma}: research and preparation of risk mitigations occurs before we have clear evidence of the capabilities underlying those risks. We believe in being proactive, and taking a cautious approach by anticipating potential risks, even before they start to appear likely. This allows us to develop a more exhaustive and informed strategy in the long run.

Nonetheless, we still prioritize those risks for which we can foresee how the requisite capabilities may arise, while deferring even more speculative risks to future research. Specifically, we focus on capabilities in foundation models that are enabled through learning via gradient descent, and consider Exceptional AGI (Level 4) from~\citet{morris2023levels}, defined as an AI system that matches or exceeds that of the 99th percentile of skilled adults on a wide range of non-physical tasks. This means that our approach covers conversational systems, agentic systems, reasoning, learned novel concepts, and some aspects of recursive improvement, while setting aside goal drift and novel risks from superintelligence as future work.

We focus on technical research areas that can provide solutions that would mitigate severe harm. However, this is only half of the picture: technical solutions should be complemented by effective governance. It is especially important to have broader consensus on appropriate standards and best practices, to prevent a potential race to the bottom on safety due to competitive pressures. We hope that this paper takes a meaningful step towards building this consensus.

\subsection*{Background assumptions} \label{sec:extended_abstract_background_assumptions}
In developing our approach, we weighed the advantages and disadvantages of different options. For example, some safety approaches could provide more robust, general theoretical guarantees, but it is unclear that they will be ready in time. Other approaches are more ad hoc, empirical, and ready sooner, but with rough edges.

To make these tradeoffs, we rely significantly on a few background assumptions and beliefs about how AGI will be developed:
\begin{enumerate}
    \item \textbf{No human ceiling:} Under the current paradigm (broadly interpreted), we do not see any fundamental blockers that limit AI systems to human-level capabilities. We thus treat even more powerful capabilities as a serious possibility to prepare for.\footnote{Even simply automating current human capabilities, and then running them at much larger scales and speeds, could lead to results that are qualitatively beyond what humans accomplish. Besides this, AI systems could have native interfaces to high-quality representations of domains that we do not understand natively: AlphaFold~\citep{jumper2021highly} has an intuitive understanding of proteins that far exceeds that of any human.}
    \begin{itemize}
        \item \textbf{Implication:} Supervising a system with capabilities beyond that of the overseer is difficult, with the difficulty increasing as the capability gap widens, especially at machine scale and speed. So, for sufficiently powerful AI systems, our approach does not rely purely on human overseers, and instead leverages AI capabilities themselves for oversight.
    \end{itemize}

    \item \textbf{Timelines:} We are highly uncertain about the timelines until powerful AI systems are developed, but crucially, we find it plausible that they will be developed by 2030.
    \begin{itemize}
        \item \textbf{Implication}: Since timelines may be very short, our safety approach aims to be “anytime”, that is, we want it to be possible to quickly implement the mitigations if it becomes necessary. For this reason, we focus primarily on mitigations that can easily be applied to the current ML pipeline.
    \end{itemize}

    \item \textbf{Acceleration:} Based on research on economic growth models, we find it plausible that as AI systems automate scientific research and development (R\&D), we enter a phase of accelerating growth in which automated R\&D enables the development of greater numbers and efficiency of AI systems, enabling even more automated R\&D, kicking off a runaway positive feedback loop.
     \begin{itemize}
        \item \textbf{Implication:} Such a scenario would drastically increase the pace of progress, giving us very little calendar time in which to notice and react to issues that come up. To ensure that we are still able to notice and react to novel problems, our approach to risk mitigation may involve AI taking on more tasks involved in AI safety. While our approach in this paper is not primarily targeted at getting to an AI system that can conduct AI safety R\&D, it can be specialized to that purpose.
    \end{itemize}
    \item \textbf{Continuity:} While we aim to handle significant acceleration, there are limits. If, for example, we jump in a single step from current chatbots to an AI system that obsoletes all human economic activity, it seems very likely that there will be some major problem that we failed to foresee. Luckily, AI progress does not appear to be this discontinuous. So, we rely on approximate continuity: roughly, that there will not be large discontinuous jumps in general AI capabilities, given relatively gradual increases in the inputs to those capabilities (such as compute and R\&D effort). Crucially, we do not rely on any restriction on the rate of progress with respect to calendar time, given the possibility of acceleration.
     \begin{itemize}
        \item \textbf{Implication:} We can iteratively and empirically test our approach, to detect any flawed assumptions that only arise as capabilities improve.
        \item \textbf{Implication:} Our approach does not need to be robust to arbitrarily capable AI systems. Instead, we can plan ahead for capabilities that could plausibly arise during the next several scales, while deferring even more powerful capabilities to the future.
    \end{itemize}
    
\end{enumerate}

\subsection*{Risk areas} \label{sec:extended_abstract_risk_areas}

\begin{figure}[t]
    \centering
    \includegraphics[width=\linewidth]{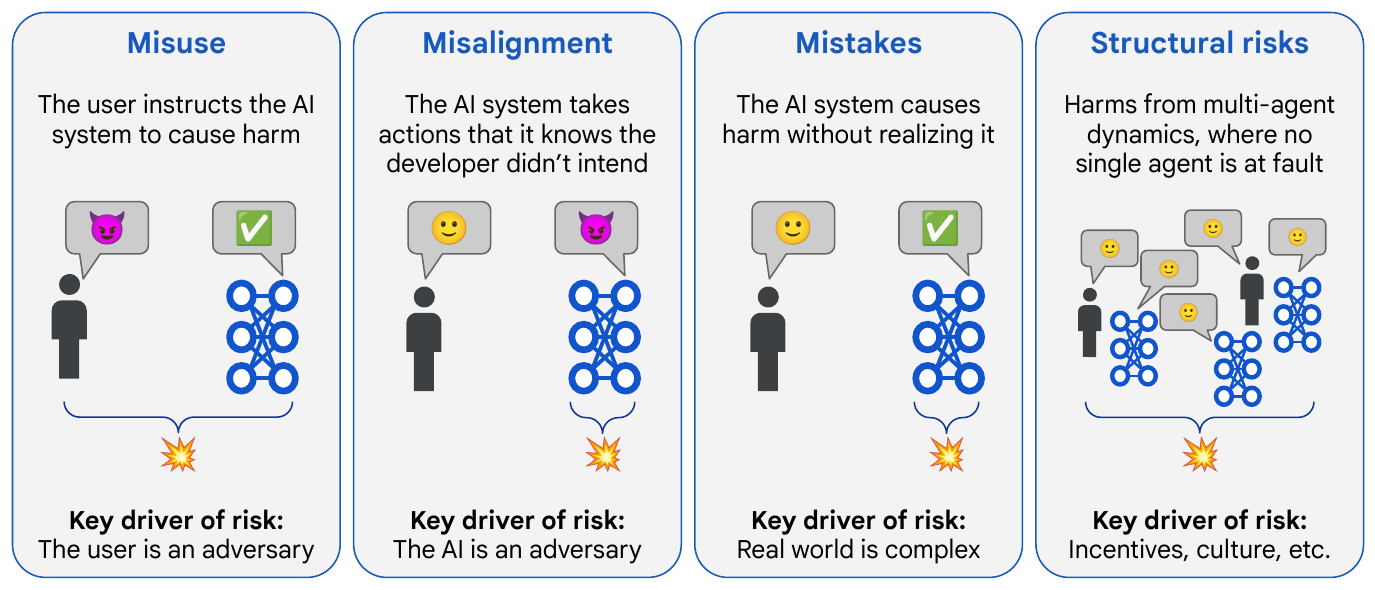}
    \caption{\textbf{Overview of risk areas.} We group risks based on factors that drive differences in mitigation approaches. For example, misuse and misalignment differ based on which actor has bad intent, because mitigations to handle bad human actors vary significantly from mitigations to handle bad AI actors.}
    \label{fig:risk-areas-extended-abstract}
\end{figure}

When addressing safety and security, it is helpful to identify broad groups of pathways to harm that can be addressed through similar mitigation strategies. Since the focus is on identifying similar mitigation strategies, we define areas based on abstract structural features (e.g. which actor, if any, has bad intent), rather than concrete risk domains such as cyber offense or loss of human control. This means they apply to harms from AI in general, rather than being specific to AGI.

We consider four main areas:
\begin{enumerate}
    \item \textbf{Misuse:} The user intentionally instructs the AI system to take actions that cause harm, against the intent of the developer. For example, an AI system might help a hacker conduct cyberattacks against critical infrastructure.
    \item \textbf{Misalignment:} The AI system knowingly\footnote{This definition relies on a very expansive notion of what it means to know something, and was chosen for simplicity and brevity. A more precise but complex definition can be found in \Cref{sec:misalignment-risks}. For example, we include cases where the model has learned an ``instinctive'' bias, or where training taught the model to ``honestly believe'' that the developer's beliefs are wrong and actually some harmful action is good, as well as cases where the model causes harm without attempting to hide it from humans. This makes misalignment (as we define it) a broader category than related notions like deception.} causes harm against the intent of the developer.\footnote{By restricting misalignment to cases where the AI knowingly goes against the developer's intent, we are considering a specific subset of the wide set of concerns that have been considered a part of alignment in the literature. For example, most harms from failures of multi-multi alignment~\citep{critch2020ai} would be primarily structural risks in our terminology.} For example, an AI system may provide confident answers that stand up to scrutiny from human overseers, but the AI knows the answers are actually incorrect. Our notion of misalignment includes and supersedes many concrete risks discussed in the literature, such as deception, scheming, and unintended, active loss of control.
    \item \textbf{Mistakes:} The AI system produces a short sequence of outputs that directly cause harm, but the AI system did not know that the outputs would lead to harmful consequences that the developer did not intend. For example, an AI agent running the power grid may not be aware that a transmission line requires maintenance, and so might overload it and burn it out, causing a power outage.
    \item \textbf{Structural risks:} These are harms arising from multi-agent dynamics – involving multiple people, organizations, or AI systems – which would not have been prevented simply by changing one person's behaviour, one system's alignment, or one system's safety controls.
\end{enumerate}

Note that this is not a categorization: these areas are neither mutually exclusive nor exhaustive. In practice, many concrete scenarios will be a mixture of multiple areas. For example, a misaligned AI system may recruit help from a malicious actor to exfiltrate its own model weights, which would be a combination of misuse and misalignment. We expect that in such cases it will still be productive to primarily include mitigations from each of the component areas, though future work should consider mitigations that may be specific to combinations of areas.

When there is no adversary, as with \textbf{mistakes}, standard safety engineering practices (e.g. testing) can drastically reduce risks, and should be similarly effective for averting AI mistakes as for human mistakes. These practices have already sufficed to make severe harm from human mistakes extremely unlikely, though this is partly a reflection of the fact that severe harm is a very high bar. So, we believe that severe harm from AI mistakes will be significantly less likely than misuse or misalignment, and is further reducible through appropriate safety practices. For this reason, we set it out of scope of the paper. Nonetheless, we note four key mitigations for AI mistakes: improving AI capabilities, avoiding deployment in situations with extreme stakes, using shields that verify the safety of AI actions, and staged deployment.

\textbf{Structural risks} are a much bigger category, often with each risk requiring a bespoke approach. They are also much harder for an AI developer to address, as they often require new norms or institutions to shape powerful dynamics in the world. For these reasons, these risks are out of scope of this paper.

This is not to say that nothing is done for structural risks – in fact, much of the technical work discussed in this paper will also be relevant for structural risks. For example, by developing techniques to align models to arbitrary training targets, we provide governance efforts with an affordance to regulate models to fuzzy specifications.

Our strategy thus focuses on \textbf{misuse} and \textbf{misalignment}. For misuse, the strategy is put in practice through our Frontier Safety Framework \citep{gdm2025fsf}, which evaluates whether the model has the capability to cause harm, and puts in place security and deployment mitigations if so. Our strategy for misalignment begins with attaining good oversight, which is a key focus area across the AGI safety field. In addition, we put emphasis on where to get oversight, i.e. on what tasks oversight is needed in order for the trained system to be aligned. 

\subsection*{Misuse} \label{sec:extended_abstract_misuse}

\begin{figure}[t]
    \centering 
    \includegraphics[width=\linewidth]{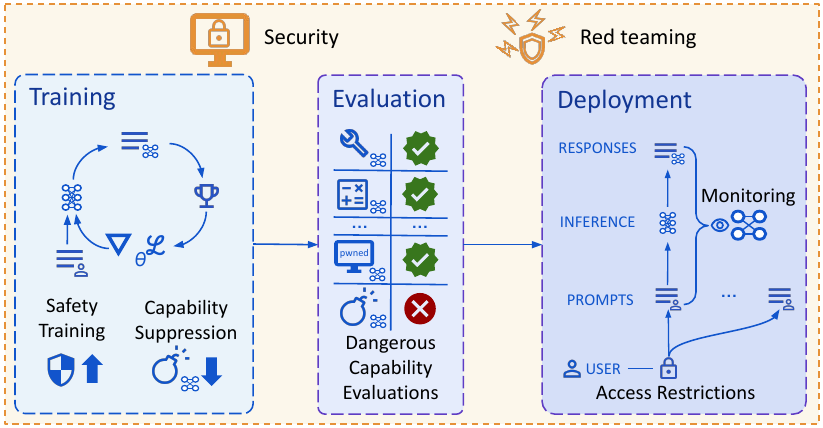}
    \caption{\textbf{Overview of our approach to mitigating misuse.} We aim to block bad actors' access to dangerous capabilities, through a combination of security for model weights, model-level mitigations (``training'' box), and system-level mitigations (``deployment'' box). Dangerous capability evaluations assess whether mitigations are necessary, while red teaming assesses their sufficiency.}
    \label{fig:misuse-overview-extended-abstract}
\end{figure}

Misuse occurs when a human deliberately uses the AI system to cause harm, against the developer’s wishes. To prevent misuse, we put in place security and deployment mitigations that prevent bad actors from getting enough access to dangerous capabilities to cause severe harm.

Given a (mitigated) AI system, we can assess the mitigations by trying to misuse the models ourselves (while avoiding actual harm), and seeing how far we can get. Naively, if we are not able to cause (proxy) harm with the models, then an external actor will not be able to cause (actual) harm either. This naive argument should be supplemented by additional factors, e.g. that a bad actor may put in much more effort than we do. Conversely, if we are able to cause proxy harm, then we need to strengthen mitigations.

As a special case of the above, when there are no mitigations in place, we are checking whether the model even has the capabilities to cause severe harm in the first place, and only introduce mitigations if we identify such capabilities. This is the approach we put into practice at Google DeepMind through our Frontier Safety Framework \citep{gdm2025fsf}.

\subsubsection*{Risk assessment} \label{sec:extended_abstract_risk_assessment}

\noindent \textbf{Misuse threat modelling} identifies concrete plausible ways that threat actors could cause severe harm through powerful capabilities of frontier AI systems. Questions of importance include: what capabilities the AI system has, what the threat actor does with these capabilities, what kinds of threat actors might be involved, and so forth. The goal is to produce realistic descriptions of the most plausible pathways to harm as well as their expected damages, so that we know what model capabilities could serve as a proxy for increased risk of severe harm in the real world.

\noindent \textbf{Dangerous capability evaluations} are a concrete way to measure the degree to which those capabilities exist in the AI system. We define a suite of tasks that we believe capture the capabilities that are representative of potential misuse, and see if we can use our models to score highly on the task suite.

Typically, we would use these evaluations to argue that misuse is implausible because the model lacks the requisite capabilities, and so no mitigations are required. To enable this, we define capability thresholds at which misuse could occur, and map them to some concrete, measurable score on the dangerous capability evaluations. Mitigations can be prepared on appropriate timescales based on forecasts of when the threshold will be achieved.

\subsubsection*{Mitigations} \label{sec:extended_abstract_mitigations}

\noindent \textbf{Deployment mitigations} aim to prevent bad actors from accessing dangerous capabilities through use of intended APIs. Mitigations start at the model level and include:
\begin{enumerate}
    \item \textbf{Safety post-training:} Developers can teach models not to fulfil harmful requests during post-training. Such an approach would need to additionally ensure that the model is resistant to jailbreaks.
    \item \textbf{Capability suppression:} It would be ideal to remove the dangerous capability entirely (aka “unlearning”), though this has so far remained elusive technically, and may impair beneficial use cases too much to be used in practice.
\end{enumerate}

Our approach considers further deployment mitigations at the system level, which, combined with model mitigations, provide defense in depth: 
\begin{enumerate}
    \item \textbf{Monitoring:} Monitoring involves detecting when a threat actor is attempting to inappropriately access dangerous capabilities, and responding in a way that prevents them from using any access to cause severe harm. Detection can be accomplished by using classifiers that output harm probability scores, leveraging the internal activations of the model, or auditing generated content manually.
    \item \textbf{Access restrictions:} Models with dangerous capabilities can be restricted to vetted user groups and use cases, reducing the surface area of dangerous capabilities that an actor can attempt to inappropriately access. Restrictions could also be placed if a user often triggers safety monitors.
\end{enumerate}

\noindent \textbf{Security mitigations} aim to prevent bad actors from stealing an AI system with dangerous capabilities. While many such mitigations are appropriate for security more generally, there are novel mitigations more specific to the challenge of defending AI models in particular. For example, one desideratum is to limit the number of people who can unilaterally access model weights. To achieve this, we need interfaces that enable standard ML workflows, so that model development can be done through the interfaces, which can then be hardened.

\noindent \textbf{Societal readiness mitigations} use AI systems to harden societal defenses; for example, it aims to prepare for AI cyber-offense capabilities by enabling rapid patching of vulnerabilities in critical infrastructure. Such mitigations can help ensure that even with access to dangerous capabilities, bad actors would not be able to cause severe harm.

\subsubsection*{Assurance against misuse} \label{sec:extended_abstract_assurance}

\noindent \textbf{Misuse stress tests.} Once mitigations are in place, our approach dictates creating a detailed argument for why a set of misuse mitigations, once applied, would be sufficient for reducing risk to adequate levels. This enables us to identify some key assumptions that this argument is predicated on, and carry out stress tests to identify flaws in these assumptions. For example, a dedicated red team may discover novel jailbreaks that can evade the safety finetuning mitigations.

Since a bad actor might put in much more effort, the red team is given extra advantages to compensate. For example, they could have full knowledge of the mitigations, or we could loosen the mitigation thresholds to make the system easier to attack. In addition, it is valuable to decouple the red team from mitigation development, to reduce the risk of shared blind spots. This could be done by having a separate internal expert red team, or by having external partners conduct stress tests.

\noindent \textbf{Misuse safety cases.} A safety case is a structured argument, supported by a body of evidence, that a system is safe. Our approach enables two types of safety cases:
Inability: The system does not possess the capability to cause harm. This is justified based on the results of dangerous capability evaluations, and how those capabilities enable risks.
Red teamed: For models with dangerous capabilities, we need to run stress tests to understand how robust our mitigations are. This evidence then feeds into an argument that mitigations are sufficient for reducing risk down to adequate levels. Currently, red teaming methods (both automated and manual) enable us to quantify how difficult it is to access the dangerous capability, but more research is needed to relate that to overall risk.

\subsection*{Misalignment} \label{sec:extended_abstract_risk_misalignment}

\begin{figure}[t]
    \centering 
    \includegraphics[width=\linewidth]{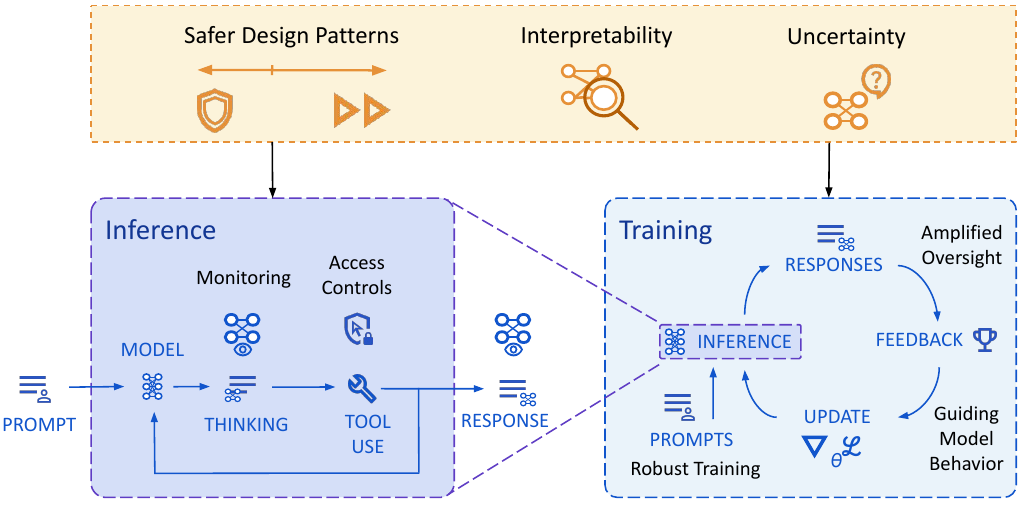}
    \caption{\textbf{Overview of our approach to mitigating misalignment.} The first line of defense aims to train an aligned model (``training'' box). Should this fail, the second line of defense aims to prevent severe harm even if the model is misaligned (``inference'' box). Note that inference mitigations continue to apply after training (that is, during deployment). In addition, while it is useful to monitor the AI system during training, we may want to avoid using the monitor as a training signal, as this could lead to the AI system hiding the bad behavior rather than removing it. Enablers such as safer design patterns, interpretability, and  uncertainty can strengthen other mitigations. While not depicted in this diagram, we also expect to conduct evaluations and stress tests to assess the necessity and sufficiency of mitigations, similarly to our approach in misuse (\Cref{fig:misuse-overview-extended-abstract}).}
    \label{fig:alignment-overview-extended-abstract}
\end{figure}

Our approach has two lines of defense. First, we aim to use model level mitigations to ensure the model does not pursue misaligned goals. We follow the standard ML playbook: (1) improve the ability to get good oversight (\textbf{amplified oversight}), (2) use this to improve the model (\textbf{guiding model behavior}), and (3) figure out where oversight is needed to train a model that will robustly do what we want (\textbf{robust training}).

Second, we consider how to mitigate harm even if the model is misaligned (often called ``AI control''), through the use of system level mitigations. We take significant inspiration from computer security in designing this line of defense, treating the AI system as an untrusted insider, motivating mitigations such as \textbf{access control}. \textbf{Monitoring} is particularly important, since it stands to benefit from increasing AI capabilities.

We also consider research areas that can strengthen the defenses above. \textbf{Uncertainty} is a key component for both robust training and monitoring. \textbf{Interpretability} can allow us to understand how the AI system works, enabling a variety of mitigations. The area of \textbf{safer design pattern}s explores design decisions that can enable safety at the cost of some other desideratum, with the goal of measuring and expanding the Pareto frontier to choose an appropriate point.

\subsubsection*{Training an aligned model} \label{sec:extended_abstract_training_an_aligned_model}

These model-level mitigations aim to create a model that is pursuing the goal that we intend.

\noindent \textbf{Amplified oversight.} A key challenge in almost any alignment plan is that it can be very difficult to tell whether a given model output is good or bad, once the model has capabilities beyond that of its overseers. As an example, imagine a human overseer trying to determine how good a move from AlphaGo is. The area of amplified oversight aims to address this: the goal is to provide an oversight signal that is as good as could be achieved if a human overseer understood all of the reasons that the AI system produced the output and had unlimited ability to reflect on the correct decision.

The key idea is to use the AI systems themselves to help identify the reasons that the AI system produced the output. For example, we could put two copies of the model in a setting where each model is optimized to point out flaws in the other’s outputs to a human “judge”. Ideally, if one model introduced a subtle flaw in their output that the judge wouldn’t notice by default, the other model would point out and explain the flaw, enabling the judge to penalise the first model appropriately.

In amplified oversight, any question that is too hard to supervise directly is systematically reduced to ones that we hypothesize can be supervised. However, humans may be systematically biased even for fairly simple questions. If this turns out to be a problem in practice, we could seek to model these deviations and automatically correct or account for them when interpreting the oversight.

\noindent \textbf{Guiding model behavior.} Given an oversight signal, we have to apply the oversight to train the model. While existing techniques such as reinforcement learning may suffice, we could potentially improve upon this. For example, new feedback modalities such as natural language critiques of model answers could increase the information provided in an update, and model internals could be used to produce a more targeted update that improves sample efficiency.

\noindent \textbf{Robust training.} Once we have an oversight signal that can distinguish between good and bad behaviour, we need to use this signal to build an AI system that behaves well.  One remaining problem is that the model may still behave poorly outside of the training distribution.

Robust training mitigates these problems by expanding the distribution of inputs on which our AI systems are trained. While exploring the broader question of what data is necessary and sufficient to induce an aligned model (reward model\footnote{Although this is framed more from a traditional reward model training perspective, the same applies to language-based reward models or constitutional methods, as well as to thinking models. In all cases, we ask what tasks the model needs to be exposed to in order to produce aligned actions in every new situation.} or policy), we keep a practical focus on cases that cause failures. Key techniques include active learning, automated adversarial training, and online training after deployment.

\subsubsection*{Defending against a misaligned model} \label{sec:extended_abstract_defending_against_a_misaligned_model}

While amplified oversight and robust training could be a sufficient solution in principle, in practice there will remain some holes. So, we aim for \textit{defense in depth}: even if the AI system is misaligned, we can mitigate the damage through appropriate defenses.

\noindent \textbf{Security.} A key approach is to treat the model similarly to an untrusted insider, motivating mitigations like access control, anomaly detection, logging and auditing. Relative to human employees, AI systems will likely perform more specialized tasks, enabling us to limit the available affordances in a more fine-grained manner. Other relevant techniques include sandboxing and hardening the environments in which AI systems act.

\noindent \textbf{Monitoring.} Increasing AI capabilities have the potential to significantly improve monitoring capability, since AI systems can operate at the necessary scale, while still bringing flexible intelligence traditionally associated only with manual review. Hierarchical supervision will be important: that is, model outputs are first monitored by a cheap system, and the most suspicious or unclear cases are escalated to more capable expensive monitors, such as those defined by amplified oversight.

\subsubsection*{Enabling stronger defenses} \label{sec:extended_abstract_enabling_stronger_defenses}

Our approach calls for investment in research areas that can strengthen our capability at the primary mitigations discussed above.

\noindent \textbf{Uncertainty.} Both robust training and monitoring benefit significantly from, and may even require, models that maintain uncertainty. This could be either in the traditional Bayesian sense, or via reasoning at the language abstraction level. For robust training, uncertainty can enable active learning, to better identify which aspects of the distribution are most useful to expand. For monitoring, uncertainty can play a key role in determining when to escalate in hierarchical supervision.

\noindent \textbf{Safer design patterns.} There are many decisions in the design of AI systems that can make it easier to achieve desired safety properties, but that may sacrifice other desirable properties. It is important to quantitatively study such tradeoffs, and invest in expanding the Pareto frontier of such tradeoffs, to help make the appropriate decision when building AI systems. Examples include:
\begin{enumerate}
    \item \textbf{Training targets:} When we establish the desired behaviour for an AI system, we can include principles that enable more user control, such as:
    \begin{enumerate}[label=\alph*), leftmargin=*]
        \item Informing the user of the AI system’s actions
        \item Checking in with the user before taking any important and irreversible actions
        \item Pausing or acting conservatively if the user provides unexpectedly negative feedback
    \end{enumerate}
    \item \textbf{Creativity:} We can control the amount of behaviour exploration or optimization when training the AI system. Less exploration can prevent the discovery of reward hacking behaviour, but also reduces the opportunity to find creative solutions that we would like.
    \item \textbf{Externalized reasoning:} We could ensure that the AI system produces explanations for its actions that humans approve of, making it more likely for us to provide good oversight.
\end{enumerate}

\noindent \textbf{Interpretability.} Intuitively, making a system safe should be much easier the more you understand how it works. While we can gain some surface understanding of how AI systems work through rigorous analysis of their behaviour, there is the potential to do much better if we could also understand the algorithms implemented by the forward pass of a trained LLM. If successful, this would have wide-ranging implications for alignment, such as:
\begin{enumerate}
    \item Checking for misaligned AI systems that try to appear aligned by passing our tests.
    \item Testing whether an AI explanation of its answer faithfully represents its internal reasoning when producing that answer.
    \item Providing information about the AI’s reasoning to an overseer, to improve amplified oversight.
\end{enumerate}

Unfortunately, interpretability research is still quite nascent and has not yet enabled applications like these. As a result, we still need significant investment in “basic science”.

\subsubsection*{Alignment assurance} \label{sec:extended_abstract_alignment_assurance}

\noindent \textbf{Alignment stress tests.} A stress test involves a red team that tries to show that some assumption in an alignment plan is false. The success or failure of the red team at this goal provides evidence about how sound the assumption is, and whether we can rely on it when aligning our AI systems. Stress tests can also produce model organisms demonstrating a particular failure mode, that can enable further research preventing that failure mode, as well as building up scientific evidence about the likelihood of failures of misalignment.

\noindent \textbf{Control evaluations.} A control evaluation extends misuse red teaming to the misalignment setting. In the misuse setting, the red team acts as a malicious external user, to simulate the adversarial actions we may observe with misuse, and we observe how successful they are. Analogously, in a control evaluation, the red team can finetune or prompt the model to try to cause (a proxy for) harm, and we observe how successful the resulting model is.

\noindent \textbf{Alignment safety cases.} As with misuse, it is important to provide structured arguments that AI systems are not egregiously misaligned.
\begin{enumerate}
    \item \textbf{Inability.} For current models, an inability safety case is appropriate: the AI system could not cause severe harm even if it was trying to do so. This can be demonstrated through capability evaluations, similarly to the case with misuse. However, this class of safety case may stop applying soon.
    \item \textbf{Supervision (aka control).} This type of argument says that if the AI system were to produce outputs that would cause harm, we would notice and take corrective action that prevents the harm. It is important that such arguments are robust to collusion, that is, the possibility that AI-enabled monitors may themselves be misaligned, and so choose not to report cases where the primary AI system takes a bad action. Control evaluations can be used as evidence supporting these safety cases.
    \item \textbf{Incentives.} An incentives-based argument appeals to the training process to argue that the AI system is pursuing the intended goal, and thus should not be misaligned. Goal misgeneralization or distribution shift poses a significant challenge to incentives-based arguments, as it is unclear how to demonstrate sufficient coverage of the situations the AI system may encounter during deployment.
    \item \textbf{Understanding.} This class of safety cases aims to analyse the final model well enough that we can certify that it only produces safe outputs – without exhaustively checking all of the potential inputs and outputs. We are currently far from any such safety case, though in principle advances in interpretability could allow for such a safety case.
\end{enumerate}

\subsection*{Limitations} \label{sec:extended_abstract_limitations}

The focus of this paper is on technical approaches and mitigations to risks from AGI. A key limitation is that this is only half of the picture: a full approach would also discuss governance, which is equally if not more important. For example, for misuse risks, it is key that all frontier developers implement the necessary mitigations to meaningfully reduce risk. If one developer implements mitigations without other developers following suit, an adversary can switch to using the less secure AI system. In addition, we do not address structural risks, since they require bespoke approaches.

We also focus primarily on techniques that can be integrated into current AI development, due to our focus on anytime approaches to safety. While we believe this is an appropriate focus for a frontier AI developer's mainline safety approach, it is also worth investing in research bets that pay out over longer periods of time but can provide increased safety, such as agent foundations, science of deep learning, and application of formal methods to AI.

We focus on risks arising in the foreseeable future, and mitigations we can make progress on with current or near-future capabilities. The assumption of approximate continuity (\Cref{sec:continuity}) justifies this decision: since capabilities typically do not discontinuously jump by large amounts, we should not expect such risks to catch us by surprise. Nonetheless, it would be even stronger to exhaustively cover future developments, such as the possibility that AI scientists develop new offense-dominant technologies, or the possibility that future safety mitigations will be developed and implemented by automated researchers.

Finally, it is crucial to note that the approach we discuss is a \emph{research agenda}. While we find it to be a useful roadmap for our work addressing AGI risks, there remain many open problems yet to be solved. We hope the research community will join us in advancing the state of the art of AGI safety so that we may access the tremendous benefits of safe AGI.

%% file: 01-introduction.tex
\section{Introduction} \label{sec:intro}

Artificial intelligence is poised to become a transformative technology with massive benefits. It could raise living standards across the world~\citep{russell2022if}, transform key sectors such as healthcare~\citep{singhal2025toward} and education~\citep{kumar2023math}, and accelerate scientific discovery~\citep{wang2023scientific}. As with any transformative technology, it also comes with risks.

The high rate of change and novelty of impacts make it hard to forecast and mitigate risks of a novel transformative technology. Premature mitigations and regulations can often backfire, reducing benefits without a compensatory reduction in harms. For example, airline deregulation has had significant economic benefits~\citep{morrison2010economic}, without adverse effects on safety~\citep{moses1990aviation}.

So, for many risks, while it is appropriate to include some precautionary safety mitigations, the majority of safety progress should be achieved through an ``observe and mitigate'' strategy. Specifically, the technology should be deployed in multiple stages with increasing scope, and each stage should be accompanied by systems designed to observe risks arising in practice, for example through monitoring, incident reporting, and bug bounties. After risks are observed, more stringent safety measures can be put in place that more precisely target the risks that happen in practice.

Unfortunately, as technologies become ever more powerful, they start to enable \textbf{severe harms}. An incident has caused severe harm if it is consequential enough to significantly harm humanity. Obviously, ``observe and mitigate'' is insufficient as an approach to such harms, and we must instead rely on a precautionary approach. For example, during the Manhattan Project, scientists worried that nuclear weapons might ignite the atmosphere, and had to rule it out through theoretical analysis~\citep{wiescher2024nuclear,konopinski1946ignition}.

We leave the exact threshold for severe harms unspecified. On one end of the spectrum, any harm that happens regularly is not a severe harm as defined in this paper. For example, a crash of a self-driving car would not count as severe harm, since fatal crashes from manually driven cars occur on a regular basis. On the other end of the spectrum, existential risks~\citep{ord2020precipice} that permanently destroy humanity are clear examples of severe harm. In between these ends of the spectrum, the question of whether a given harm is severe isn't a matter for Google DeepMind to decide; instead it is the purview of society, guided by its collective risk tolerance and conceptualisation of harm.

Given the massive potential impact of AGI, we expect that it too could pose potential risk of severe harm. In this report, we lay out Google DeepMind’s approach to mitigating severe risks from AGI, focusing on technical safety and security mitigations. Importantly, this is a roadmap rather than a solution, as there remain many open research problems to address.

Different severe harms are enabled by different capabilities, and so it is not possible to pinpoint a particular capability level at which we expect to see the various risks discussed in this paper to emerge. Nonetheless, it is useful for illustrative purposes to imagine a concrete level of capability in the AI systems we discuss. Following the terminology of \citet{morris2023levels}, we will consider Exceptional AGI (Level 4),\footnote{While Level 4 was originally called ``Virtuoso AGI'', we instead use the term ``Exceptional AGI'', which we believe better captures this capability level.} defined as an AI system that matches or exceeds that of the 99th percentile of skilled adults on a wide range of non-physical tasks. At this capability level, we identify several types of severe risks (Section~\ref{sec:risk-areas}), and focus our approach towards two main sources of severe risk: misuse and misalignment.

At a high level, our strategy for misuse is to:
\begin{enumerate}
    \item Evaluate whether the model has the capability to cause severe harm (\Cref{sec:capability-risk-assessment}).
    \item Put in place appropriate deployment mitigations (Sections~\ref{sec:model-deployment-mitigations}-\ref{sec:access-restrictions}) and security mitigations (\Cref{sec:model-security}) if so.
    \item Assess the quality of the mitigations by attempting to break them, and seeing how far we can get (Section~\ref{sec:misuse-stress-tests}).
\end{enumerate}

Likewise, our strategy for misalignment is:
\begin{enumerate}
    \item Attain good oversight, using the AI system itself to help with the oversight process (Section~\ref{sec:amplified-oversight}), and use it to train the AI system (Section~\ref{sec:guiding-model-behavior}).
    \item Identify cases where oversight is needed so that the trained model will robustly do what we want (Section~\ref{sec:robust-training-and-monitoring}).
    \item Apply defense in depth to defend against AI systems that are misaligned despite our best efforts, through a combination of monitoring with efficient and scalable oversight (Section~\ref{sec:robust-training-and-monitoring}), and the application of computer security techniques (Section~\ref{sec:hardening}).
    \item Assess the quality of the mitigation through alignment assurance techniques (Sections~\ref{sec:alignment-stress-tests} and~\ref{sec:alignment-safety-cases}).
\end{enumerate}

We also invest in \emph{enablers} that help improve a variety of these steps, including safer design patterns (Section~\ref{sec:safer-design-patterns}) and interpretability (Section~\ref{sec:interpretability}).

\paragraph{Governance.} A full treatment of risk reduction must also consider the broader sociotechnical context. For example, race dynamics~\citep{armstrong2016racing, cave2018ai} can render unilateral actions ineffective: if an incautious AI developer races ahead with no regard to safety, restraint from other developers may not accomplish very much.

Since we focus on Google DeepMind’s approach, these governance issues that require wider societal engagement are out of scope. Nonetheless, it is crucial that we develop broad consensus on appropriate standards and best practices. We hope that this paper takes a meaningful step in this direction.

\paragraph{Related work.} Our discussion here is most similar to Anthropic’s prior discussion~\citep{anthropic2023core}, though we put significantly more emphasis on robust training, monitoring and security. Meanwhile, OpenAI focuses on automating alignment research~\citep{leike2022our}. While we agree that automation of alignment research should be done when feasible (\Cref{sec:acceleration}), we view automation as a way to accelerate alignment research, rather than as the primary aim to which we should aspire.

There have been many prior literature reviews surveying AGI safety or aspects thereof~\citep{everitt2018agi, ji2023ai, shah2020ai, kenton2021alignment, kim2024road}. We do not aim to exhaustively cover the field – instead, we present the research areas and actions that, in our judgement, are most important for a frontier AI company to invest in. For example, we prioritize areas that already have empirical support, and areas that appear necessary for risk reduction.

%% file: 02-evidence-dilemma.tex
\section{Navigating the evidence dilemma} \label{sec:evidence-dilemma}

While severe harms necessitate a precautionary approach, such an approach to a new and quickly evolving technology suffers from an \textit{evidence dilemma}~\citep{ISRSAA2025}. Precautionary mitigations will be based on relatively limited evidence, and so are more likely to be counterproductive. Research efforts aimed at future problems may study mitigations that later turn out to be infeasible, disproportionate, or unnecessary. Nonetheless, as discussed previously, waiting until the evidence is conclusive (``observe and mitigate'') is not a viable option for severe harms.

However, an intermediate approach is possible. Many risks do not seem plausible given currently foreseeable capability improvements, but could arise with capability improvements that are not yet on the horizon. These risks are good candidates to defer to the future, when more evidence will be available, particularly since mitigations are much easier to design when it is known how capabilities will arise. Importantly, AI progress does not usually involve large discontinuous jumps in capability assuming continuous increases in inputs (\Cref{sec:continuity}), though the overall pace of progress may accelerate (\Cref{sec:acceleration}). As long as we accelerate safety and security research in parallel, we thus expect that risks that we currently defer to the future will still be foreseen before they are realized in practice, giving us time to develop mitigations.

In other words, for any given risk, there are two options to address it:
\begin{enumerate}
  \item Propose specific research areas to investigate, and mitigations to develop and implement, that address the risk.
  \item Defer the risk to be addressed when more evidence is available about the relevant capabilities for that risk. The evidence may reveal that the worry was unfounded, and thus no mitigations are needed, or may suggest new approaches needed for the risk in question.
\end{enumerate}
In the next subsection, we briefly outline a number of risks and capabilities within the AGI safety and security space, and explain whether they are in scope for our approach (option 1), or they are being deferred to future work once enough evidence is available (option 2).

\subsection{Applying our approach to concrete risks}

In this paper, we do not focus on concrete capabilities or pathways to harm. We develop approaches that we expect to generalize across capabilities~\citep{christiano2021mundane}. While we do consider four different risk areas (\Cref{sec:risk-areas}), these are grouped based on similarity of mitigations needed to address them, rather than the concrete harmful outcomes that may result.

For example, the first International AI Safety Report discusses risks from loss of control~\citep{ISRSAA2025}. In contrast, we do not discuss loss of control as its own category. Our mitigations for it would be split across misuse, misalignment, and structural risks, corresponding respectively to the Report’s categories of intentional active loss of control, unintentional active loss of control, and passive loss of control~\citep[Figure 2.5]{ISRSAA2025}. This reflects our expectation that the appropriate mitigations for intentional active loss of control will more closely match those for (say) defending against misuse of AI by terrorists to develop weapons, than to those for unintentional active loss of control. Similarly, we expect that as a structural risk, passive loss of control or gradual disempowerment~\citep{kulveit2025gradual} will require a bespoke approach, which we set out of scope for this paper.

Nonetheless, to illustrate the difference between the risks our approach intends to cover, and the ones that we are deferring to the future, we briefly cover some concrete risks and their relationship to our approach.

\noindent \textbf{Misuse.} Our approach to misuse proactively identifies dangerous capabilities, measures relevant model capabilities, and then implements security and deployment mitigations that together prevent bad actors from causing extreme harm. We can already see signs of some such dangerous capabilities: for example, AI models may soon be capable of cyberattacks, given they have already found at least one novel real-world vulnerability~\citep{allamanis2024bigsleep}. The general approach should largely scale well to future dangerous capabilities, as long as threat modeling identifies the risks before they lead to severe harm.

However, our approach does rely on assumptions about AI capability development: for example, that dangerous capabilities will arise in frontier AI models produced by centralized development. This assumption may fail to hold in the future. For example, perhaps dangerous capabilities start to arise from the interaction between multiple components~\citep{drexler2019reframing}, where any individual component is easy to reproduce but the overall system would be hard to reproduce. In this case, it would no longer be possible to block access to dangerous capabilities by adding mitigations to a single component, since a bad actor could simply recreate that component from scratch without the mitigations.

\noindent \textbf{Conversational systems, agentic systems, and reasoning capabilities.} Different capability profiles will enable different concrete risks. Chatbots~\citep{adiwardana2020towards} and reasoners~\citep{jaech2024openai} are on the same spectrum and the risks are similar, since reasoners can be thought of as chatbots with somewhat higher capabilities. Language agents~\citep{xi2025rise,wang2023voyager} introduce new risks because they add new affordances to language models. It would be easier for a misaligned AI system to exfiltrate its own weights if it can use network scanning tools. Embodied agents could add further risk, for example by enabling misaligned AI systems to develop weapons that it can use as threats. The scale of deployment also increases risks, as it becomes harder to practically oversee the AI system’s actions.

Despite the variation in concrete risks, our high-level approach remains the same. It aims to ensure we can effectively oversee what AI systems are doing on any given input by leveraging the AI systems themselves (amplified oversight), and to scale that oversight to large-scale deployments (robust training and monitoring). These approaches can be specialized to many different capabilities, though the details will vary. For example, reasoning models may naturally serve the role of an asynchronous, expensive monitor in between cheap classifiers and manual auditing in a hierarchical monitoring setup (\Cref{sec:robust-training-and-monitoring}). As another example, for agentic systems it is important to put significant effort into access controls on the tools that the system is allowed to use, whereas this is not as crucial for conversational systems.

\noindent \textbf{Superhuman concepts and capabilities.} A more challenging possibility is that AI systems develop superhuman concepts or capabilities that unaided humans do not easily understand. This is also within the scope of our approach, relying particularly on the research area of amplified oversight (\Cref{sec:amplified-oversight}), which aims to continue to provide effective oversight in this regime.

While there are significant theoretical challenges in the worst case~\citep{christiano2021eliciting}, the overseer has a number of advantages 
that suggest that they can oversee an AI system with moderately stronger capabilities or concepts, even if it would not work with superintelligence.

\noindent \textbf{Goal drift.} Theoretical models of intelligence typically imply that intelligent agents should be able to learn their environment online in a sample-efficient manner~\citep{legg2007universal, hutter2024introduction}. In the context of language models, we could imagine powerful inference-time reasoning that allows the model to improve the concepts in its world model, provoking a drift in its goals due to an ontological crisis~\citep{de2011ontological}. More generally, in-context learning could mean that the model’s capabilities and behavioral tendencies could change drastically (given sufficient inference-time compute), potentially invalidating any training-based safety.

While inference-time reasoning can certainly improve capabilities~\citep{jaech2024openai}, it currently appears to be much less powerful than would be needed for an ontological crisis. \cite{cotra2020forecasting} identified meta-learning (for which in-context learning is one example) as a key bottleneck where AI systems would likely significantly lag behind human capability for a long time. The ARC-AGI benchmark~\citep{chollet2024arc} tests this same skill with the express goal of producing a benchmark that is easy for humans and hard for AIs. In addition, current architectures face a parallelism tradeoff~\citep{merrill2023parallelism}, where to enable highly parallel pretraining they give up the expressive power to do substantial sequential reasoning, even with arbitrarily long contexts, suggesting that the necessary level of powerful inference-time reasoning may not arise without significant architectural innovation.

As a result, our approach does not aim to directly address the risk of goal drift, though we expect many of the techniques we suggest can be extended to cover this setting as well. It is important to continue to monitor AI capability at meta-learning to ensure that we notice when this risk does become more plausible.

Importantly, our approach \textit{does} aim to handle goal misgeneralization~\citep{shah2022goal}, in which an AI system may appear to pursue a different goal under distribution shift (rather than meta-learning), either because its goal changes for different inputs, or because we were wrong about what its goal was, and this was only revealed to us after a distribution shift.

\noindent \textbf{Recursive improvement.} We find it plausible that AI systems will enable recursive AI improvement, where AI systems are used to conduct AI research enabling the creation of better AI systems, leading to a positive feedback loop that can drive an intelligence explosion~\citep{yudkowsky2013intelligence} (\Cref{sec:acceleration}). For the foreseeable future, we expect that this will occur via AI systems doing similar types of AI research as human researchers, as reflected in capability evaluations like RE-bench~\citep{wijk2024re}. In contrast, we do not expect that, for example, the AI system will edit its own weights, and defer risks related to such capabilities to future safety and security research.

One challenge with recursive improvement is that a misaligned AI system could sabotage the research it does~\citep{benton2024sabotageevaluations}. This is one concrete instantiation of a misalignment risk, and our approach to misalignment naturally handles it in the same way that it handles other risks from reasoners and agents.

The larger challenge is that recursive improvement can drastically increase the rate of AI progress, potentially developing novel capabilities that our approach does not address, before researchers have a chance to develop mitigation strategies for those capabilities. The natural solution here is to also apply AI to the safety and security research process, accelerating it to keep pace with progress in AI capabilities. In the foreseeable future, AI is poised to accelerate individual subtasks such as experiment implementation~\citep{jimenez2023swe}, dataset creation~\citep{perez2022discovering}, and reward modeling~\citep{bai2022constitutional, zhang2024generative, guan2024deliberative}. These tasks are shared between capabilities and safety research, so AI progress is likely to accelerate both at roughly similar rates. In the longer term, it is hard to predict whether recursive improvement will differentially benefit capabilities research or safety and security research; if it is significantly skewed towards accelerating capabilities research, then we may need greater investment in applying AI to improve safety and security research. How this should be done will heavily depend on the capability profile of future AI systems, and so we defer it to future work.

\noindent \textbf{Superintelligence.} \citet{bostrom2014superintelligence} defines superintelligence as human-like general intelligence taken to an incredible, unimaginable degree. This would necessarily include capabilities that our approach does not address, such as strong in-context learning that can produce novel concepts (a capability that humans possess). If it were to be developed, superintelligence would likely have other capabilities we have not considered that pose risks of their own, and that will require novel mitigations. Since we do not yet know what these capabilities will be, we defer them to future safety and security research.

It is important to note that, as discussed at the beginning of this section, even with foreseeable capabilities we expect safety and security research to be significantly accelerated. If superintelligence is ever developed, we expect that at some earlier point, human researchers may have become obsolete due to automated researchers. So, a key part of safety and security will be \textit{bootstrapping}; that is, AI systems are designed by earlier AI systems that have already been aligned, and do a better job at subsequent alignment than humans would.

Importantly, only our first line of defense (training an aligned model) enables bootstrapping. Intuitively, an aligned model will ``try its best'' when aligning the next model. However, our second line of defense only aims to defend against severe harm from a misaligned model. It would be harder to defend against misaligned models sabotaging our bootstrapping efforts, since this would involve novel research that may be hard to verify. In addition, defending against severe harm does not entail that the AI system will ``try its best'' at a task such as aligning the next model, and so it may not give us the benefits of bootstrapping.

%% file: 03-assumptions/intro.tex
\newcommand\currentParadigmAssumption{current paradigm continuation assumption\xspace}
\newcommand\noHumanCeilingAssumption{no human ceiling assumption\xspace}
\newcommand\uncertainTimelinesAssumption{uncertain timelines assumption\xspace}
\newcommand\potentialForAcceleratingImprovementAssumption{potential for accelerating improvement assumption\xspace}
\newcommand\continuityAssumption{approximate continuity assumption\xspace}

\section{Assumptions about AGI development}\label{sec:assumptions}

In this section, we describe five core assumptions that underpin our approach to technical safety:
\begin{enumerate}
    \item The \textit{\currentParadigmAssumption} (\Cref{sec:paradigm}): frontier AI systems continue to be developed within the current paradigm for the foreseeable future.
    \item The \textit{\noHumanCeilingAssumption} (\Cref{sec:superhuman}): AI capabilities will not cease to advance once they achieve parity with the most capable humans.
    \item The \textit{\uncertainTimelinesAssumption} (\Cref{sec:timelines}): the timeline for the future development of highly capable AI remains unclear.
    \item The \textit{\potentialForAcceleratingImprovementAssumption} (\Cref{sec:acceleration}): AI automation of scientific research and development could precipitate accelerating growth via a positive feedback loop.
    \item The \textit{\continuityAssumption} (\Cref{sec:continuity}): there will not be large, discontinuous jumps in AI capability given continuous inputs in the form of computation and R\&D effort.
\end{enumerate}
For each assumption, we summarize the evidence providing its support and outline the implications for our approach to technical safety.

%% file: 03-assumptions/paradigm.tex
\subsection{Current paradigm continuation} \label{sec:paradigm}

\noindent Our first core assumption is the \textit{\currentParadigmAssumption}, which we define to be the assumption that the current paradigm for developing frontier AI systems will continue to represent the dominant approach to increasing AI capability for the foreseeable future.
The key influence of this assumption is that we target an \textit{anytime} approach to AGI safety, that is, an approach that could be put into place at any point.
Since we believe that the current paradigm represents the most plausible R\&D trajectory for achieving highly capable systems in the foreseeable future,
we must be adequately prepared if and when highly capable AI is developed within the current paradigm.
We also focus on the current paradigm as a matter of pragmatism.
Put simply, it would be infeasible to develop an effective risk mitigation strategy in the absence of a basic model of how AI will be developed. 

For the purposes of this document, we define the current paradigm as a suite of technical approaches that build on three foundational ideas.
These include:
(1) The importance of \textit{scaling up of computation and data} (enabled by innovation in both hardware and software).
(2) The central role of \textit{learning} and \textit{search} as mechanisms to leverage this computation and data to impart capabilities to AI systems.
(3) The significance of \textit{algorithmic innovations} that enable more efficient use of computation and data for learning and search.
These ideas are exemplified by breakthrough results achieved with Large Language Models (LLMs)~\citep{brown2020language} and deep reinforcement learning with inference-time search~\citep{silver2016mastering} where deep learning~\citep{lecun2015deep} has played a central role.
However, we also consider the ecosystem of affordances surrounding these artifacts---such as tool use through API calls and agents constructed from multimodal LLMs via scaffolding---to form part of the current paradigm, and our approach aims to be robust to changes in this ecosystem.

\noindent \textbf{Implications for our approach to AI safety:}
Since we assume that learning and search will form key components of future AI systems, we require an accurate oversight signal that can assess whether a given AI action is good or bad.
This motivates our focus on the development of \textit{Amplified Oversight} algorithms that can continue provide this oversight signal even as AI systems become more capable than their human overseers (\Cref{sec:amplified-oversight}).
However, even with Amplified Oversight, there remain two key challenges in using this oversight signal to train AI systems that behave as intended.
First, we will only be able to provide oversight on actions produced within the training distribution.
Second, we anticipate that Amplified Oversight signal will be costly to obtain.
These challenges motivate the development of \textit{robust training} strategies that expand the training distribution with a particular focus on cases that induce failures, as well as \textit{monitoring} strategies (\cref{sec:robust-training-and-monitoring}).
Conceptually, monitoring comprises a triage system in which a cheaper system is used to oversee model outputs and escalates only the most suspicious cases to full Amplified Oversight.
Note that both kinds of strategies derive considerable benefit from models that can accurately estimate \textit{uncertainty}, another building block of our approach.
Concretely, robust training strategies can leverage uncertainty to select the most informative cases for training, while monitoring models can leverage uncertainty to better inform their escalation decisions.

\noindent \textbf{Summary of supporting arguments:}
At a high level, our argument amounts to the observation the current paradigm has a strong track record to date, no obviously insurmountable barriers to continued progress in the near future and no clear competitors showing similar promise or progress.
However, for the sake of greater specificity, we frame our argument by first identifying the two objections that, if true, would falsify the \currentParadigmAssumption.
\textit{Objection 1: ``The current paradigm will hit a wall''}---progress within the current paradigm will slow to an indefinite crawl
and do so prior to reaching AI capabilities that pose the risks of severe harm discussed in this document (\cref{sec:risk-areas}).
\textit{Objection 2: ``There will be a paradigm shift in which the current paradigm is abandoned.''}---a new paradigm will emerge to supplant current paradigm, and the field of AI development will undergo a paradigm shift~\citep{Kuhn1962TheSO}.

To argue against \textit{Objection 1}, we present evidence suggesting that AI capability will continue to progress within the current paradigm in the near future.
For the purposes of this argument, we consider evidence relating to a five year future time horizon (2030).
We do so partly as a matter of feasibility---this is the time horizon over which we can make reasonably informed estimates about key variables that we believe drive progress within the current paradigm (though we anticipate that the current paradigm will continue beyond this).
We also do so partly as a matter of pragmatism---planning over significantly longer horizons is challenging in a rapidly developing R\&D environment.
As such, we anticipate that some revisions to our assumptions, beliefs and approach may be appropriate in the future.

Concretely, our argument against (1) consists of three claims.
Our first claim is that long run AI capability improvements to date have been driven through large-scale increases in compute, data and algorithmic efficiency.
Our second claim is that the primary ``inputs'' (computation, data, algorithmic efficiency) that have historically driven AI capability advances can plausibly continue to grow at their historical rates into the near future.
Our third claim is that this scaling of inputs is likely to continue to yield meaningful gains in AI capability.

Our argument against \textit{Objection 2} shares common ground with our argument against \textit{Objection 1} and rests on three additional claims.
Our fourth claim is that the current paradigm enjoys a significant ``first mover advantage'' with respect to research expertise and capital investment over any alternative paradigm that emerges as a competitor in the near future.
Our fifth claim, drawing on our first three claims, is that, since we expect the current paradigm to continue to deliver progress, there will be reduced incentive for research investments in alternative paradigms.
Our sixth and final claim is that there is little evidence of mature alternatives that show close to the same degree of promise or rapid progress as the current paradigm.

Note that while we believe that there is solid evidence supporting the current paradigm assumption, we acknowledge the extreme difficulty of predicting the future.
Consequently, we must remain prepared to revise our approach to AGI in the face of evidence against the current paradigm assumption.
In more detail, if \textit{Objection 1} is proven correct, many of the research investments underpinning our technical approach will diminish in relevance. 
However, from the perspective of severe harm mitigation, our approach will fail relatively gracefully in this scenario.
Since AI capability advances would likely slow to a crawl, the need for technical solutions that mitigate the risk of severe harms from AI via the pathways we describe in~\cref{sec:risk-areas} are significantly diminished.
If \textit{Objection 2} is proven correct, the implications for severe harm mitigation are more significant.
Many of the plans that we outline in this document will no longer be appropriate.
Should this occur, we will need to fundamentally reconsider our approach.

\subsubsection{Compute, data, and algorithmic efficiency as drivers of AI capability progress} \label{sec:drivers-of-ai-progress}

Our first three claims argue that (1) compute, data, and algorithmic efficiency are the key factors behind prior AI capability progress, (2) these inputs will plausibly continue to increase near historical rates over the next five years, and (3) this increase will continue to deliver further AI capability progress.

\noindent \textit{Claim 1: Long-run AI capability improvements have been driven through large-scale increases in compute, data and algorithmic efficiency.}  

More than two decades ago, ~\citet{moravec1998will} reflected on AI progress and observed that  \textit{``...the performance of AI machines tends to improve at the same pace that AI researchers get access to faster hardware.''}
More recently, \citet{sutton2019bitter} assessed the trajectory of AI research and drew similar conclusions about the critical role played by the improving price-performance of computation, noting that \textit{``The biggest lesson that can be read from 70 years of AI research is that general methods that leverage computation are ultimately the most effective, and by a large margin. The ultimate reason for this is Moore's law, or rather its generalization of continued exponentially falling cost per unit of computation.''}
In support of this position, \citet{sutton2019bitter} cites projects spanning domains such as chess, go, speech recognition and computer vision in which general methods based on learning and search ultimately outcompeted specialized approaches once sufficient computation became available.
Indeed, many of the most highly-cited results in the machine learning literature over the last 15 years draw heavily on this trend.
In the field of natural language processing, breakthrough performance was achieved by scaling up computation, data and model capacity~\citep{mikolov2013efficient,amodei2016deep,radford2019language,brown2020language}.
In reinforcement learning, large-scale computation enabled the combined strengths of deep neural networks and Monte Carlo tree search to achieve superhuman performance at the game of go \citep{silver2016mastering}.
While this performance was initially achieved by training on human game play, later variants achieved the same feat purely through self-play~\citep{silver2017mastering} and likewise surpassed human expert performance on other board games~\citep{silver2018general}.
In speech recognition, increased model capacity and data has driven major gains in performance~\citep{amodei2016deep}.
Similarly in computer vision, the use of hardware accelerators enabled researchers to scale up Convolutional Neural Networks (CNNs) of increasingly high capacity~\citep{krizhevsky2012imagenet,simonyan2014very,he2016deep,hu2020squeeze} to take advantage of larger training datasets made available through projects like ImageNet~\citep{deng2009imagenet}.

Note that while many frontier systems have leveraged greater computation and data than their predecessors, it has not been the case that all approaches have benefited equally from the additional resources.
In line with the perspective of \cite{sutton2019bitter}, Transformer architectures~\citep{vaswani2017attention}---imposing fewer inductive priors than other architectures such as CNNs---have grown in prominence as computation has continued to scale.
The influence of reduced inductive priors has been most evident in the field of computer vision, where Transformers were found to outperform CNNs only once pretraining data reached scales considerably larger than the one million images of ImageNet~\citep{dosovitskiy2020image}.

Beyond qualitative evidence that growing computation and data is driving AI progress, a growing body of research has demonstrated that it is possible to explicitly quantify the relationship between scaling of computation and data and AI capability with respect to a few key metrics of interest.
Building on prior theoretical work~\citep{amari1992four} and empirical studies illustrating reliable performance improvements with increased data~\citep{banko2001scaling,amodei2016deep,sun2017revisiting}, \citet{hestness2017deep} characterize the relationship between dataset size, model size and generalization metrics (such as cross entropy loss on a held out validation set).
Their results suggest relatively smooth power law relationships across each of the tasks that they investigate (translation, language modeling, and image classification).
\citet{kaplan2020scaling} build on these findings and conduct a large-scale study that reveals a smooth power law relationship between compute, dataset size, model capacity and language modeling loss, with ``scaling law'' trends spanning six orders-of-magnitude. 
As the resources dedicated to training have grown, there has been increasing focus on using these resources efficiently.
\citet{hoffmann2022training} revisited the power law coefficients derived by \cite{kaplan2020scaling} to produced revised estimates, ultimately 
enabling the training of more capable models for fixed FLOP budgets.
More recently, scaling laws have played a key role in guiding development choices for training frontier LLMs~\citep{achiam2023gpt,team2023gemini,dubey2024llama}.
While many scaling studies focus on predicting properties such as model loss, there are also efforts that directly link computation to downstream model capabilities (such as aggregate performance across many tasks).
\cite{owen2024predictable} conduct such a study, concluding that aggregate benchmark performance is indeed fairly predictable from compute scaling. Similarly, \citet{yuan2023scaling} observe that the improvements in pretraining loss tracked by scaling laws  directly translate into improved performance on reasoning benchmarks.

The view that scaling compute and data is a key driver of AI development can also be observed in the behavior of AI developers.
\cite{amodei2018ai} estimate that the total compute used for the largest training runs grew by a factor of approximately 300,000 between 2012 and 2018.
Over a longer time span (2010 - 2024),~\cite{epoch2024trainingcomputeoffrontieraimodelsgrowsby45xperyear} estimate an annual average growth rate of approximately 4$\times$ in the total computation used by frontier training runs.

In addition to leveraging train-time compute for improved learning, a number of studies have observed and quantified the smooth improvements in AI capability induced by scaling test-time computation for improved search.
\cite{jones2021scaling} conducted studies across a family of AlphaZero agents~\citep{silver2018general} and demonstrated that training compute could be smoothly and predictably exchanged for test-time compute (used to perform Monte Carlo tree search) in a manner that maintained game play performance.
\citet{epoch2023tradingoffcomputeintrainingandinference} build on \cite{jones2021scaling} by collating evidence across a number of domains suggesting that increased test-time compute often yields predictable gains in performance.
In the coding domain, \cite{li2022competition} observe that using additional compute to sample repeatedly from the model scales log-linearly with correct solution generation. 
While naive as a search strategy, scaling up simple repeated sampling of LLM responses has proven to be remarkably effective at producing at least one correct solution~\citep{brown2024largelanguagemonkeysscaling}.
More broadly, scaling up test-time search through strategies such as Best-of-N sampling~\citep{nakano2021webgpt} and beam search against a process reward model~\citep{lightman2023letsverify} appear to yield smooth performance improvements on reasoning tasks~\citep{snell2024scaling}.

Beyond scaling up data and computation, a third key factor driving AI progress has been algorithmic innovation.
This acts as a ``force multiplier'' on data and computation by increasing the efficiency with which they are brought to bear on a problem.
In a study of six areas of algorithms research (SAT solvers, chess, go, physics simulation, factorization and mixed integer programming), \cite{grace2013algorithmic} estimates that gains from algorithmic progress have been between 50\% and 100\% as large as those driven by hardware progress.
\cite{hernandez2020measuring} estimate that the number of floating point operations required to train a classifier to match AlexNet~\citep{krizhevsky2012imagenet} accuracy on ImageNet fell by a factor of 44 between 2012 and 2019.
This corresponds to a doubling of algorithmic efficiency every 16 months for 7 years (faster than Moore's law).
Demonstrating this progress, \cite{karpathy2022deep} re-implemented the hand-written digit recognition neural network of~\citep{lecun1989backpropagation} using algorithmic innovations accumulated in the intervening period, achieving a 60\% reduction in error rate.
\cite{erdil2022algorithmic} sought to decompose AI progress in a way that can be attributed to the separate factors of scaling (compute, model size and data) and algorithmic innovation, and concluded that algorithmic progress doubles effective compute budgets roughly every nine months.
\cite{ho2024algorithmic} further extend this approach to study algorithmic improvements in the pretraining of language models for the period of 2012 - 2023.
During this period, the authors estimate that the compute required to reach a set performance threshold halved approximately every eight months.

To summarize, we have discussed several sources of evidence in support of \textit{Claim 1}.
First, we highlighted qualitative evidence in the form of anecdotal observations (backed by empirical data) from senior researchers suggesting that increased computation has driven AI progress.
We then described a prominent trend: many AI capability breakthroughs over the last 10-15 years have been achieved by scaling up computation and data (supported by improvements in algorithmic efficiency).
Next, we discussed a number of papers aiming to more precisely quantify the relationship between compute, data, algorithmic efficiency and improving AI capabilities with respect to appropriate metrics.
Collectively, we believe this literature indicates that scaling compute, data and algorithmic efficiency have been the key drivers of performance gains to date. 
As further evidence, we cited analysis suggesting that AI developers have dramatically increased the computation and data employed for training AI systems, reflecting their belief that these inputs are key to achieving enhanced AI capability.
In aggregate, we believe that \textit{Claim 1} is strongly supported by the evidence above.

\noindent \textit{Claim 2: It is plausible that the increases in computation, data and algorithmic efficiency will continue to grow at recent historical rates for the next five years.}

As discussed above, we believe the key drivers of frontier AI progress to date have been the scaling of compute, data and algorithmic efficiency.
We next consider whether this scaling can plausibly continue.
To contextualize an assessment of plausibility, we first observe that the scaling of compute for the largest training runs (a factor of approximately $4\times$ per year~\citep{epoch2024trainingcomputeoffrontieraimodelsgrowsby45xperyear}) appears to be unprecedented.
As noted by~\citet{epoch2024canaiscalingcontinuethrough2030}, this rate of growth exceeds that of the most explosive technological expansions in history.
In particular, it exceeds the peak rates of growth of solar capacity installation ($1.5\times$ per year, 2001-2010), mobile phone adoption ($2\times$ per year, 1980-1987) and human genome sequencing ($3.3\times$ per year, 2008-2015).
It is therefore important to assess whether it is even technically feasible for such scaling to continue.

In the most detailed study of the topic conducted to date,~\cite{epoch2024canaiscalingcontinuethrough2030} investigate whether there are technical barriers preventing the compute of frontier training runs from scaling in line with its recent trajectory of $4\times$ per year.
In particular, this analysis examines the plausibility of a 2030 training run of 2e29 FLOPs, equivalent to a 10,000$\times$ scaling up of training compute from GPT-4~\citep{achiam2023gpt}.
\cite{epoch2024canaiscalingcontinuethrough2030} focus on four potential bottlenecks.
The first potential bottleneck is power supply (specifically power supply in the US, where the largest publicly known training runs have taken place thus far). 
The authors determine that the necessary power (specifically data center campuses in the 1 - 5 GW range) will likely be available, and thus power supply will not limit scaling at the current trajectory.
Moreover, distributed training~\citep{douillard2023diloco,douillard2024dipaco} could enable significantly greater power to be accessed by frontier training runs.
A second potential bottleneck to scaling up compute is the availability of hardware accelerators (such as GPUs and TPUs).
While there is significant uncertainty to estimating future capacity, \cite{epoch2024canaiscalingcontinuethrough2030} determine that it is highly plausible that there will be sufficient accelerator capacity for a 2e29 FLOP training run in 2030.
More specifically, the authors project capacity for 100M H100-equivalents to be available for a training run (capable of supporting a 2e29 FLOP training run under appropriate assumptions).
The third bottleneck is data scarcity---whether there is likely to be sufficient data to continue scaling frontier training runs.
Broadly, the authors conclude that training data will not be a binding constraint, given the projected growth in online text data and the availability of vast corpora of multimodal data.
The final bottleneck investigated by \cite{epoch2024canaiscalingcontinuethrough2030} is the ``latency wall''.
This represents an effective speed limit imposed by the minimum time required for forward and backward passes in deep neural networks on the assumption that batch size cannot be productively scaled without limit.
However, as with the other constraints, the authors predict that this constraint will not prevent scaling.

While the analysis describes above underscores the technical feasibility of continued compute and data scaling, it does not provide evidence that AI developers will be prepared to spend the hundreds of billions of dollars required to implement this scaling.
To a large degree, it would seem likely that this willingness will depend on the evidence produced by intermediate models that scaling continues to yield AI capability gains that can ultimately produce significant economic value.
However, given that total labor compensation represents over 50\% of global GDP~\citep{ilo_ourworldindata_labor_share_gdp}, it is clear that the economic incentive for automation is extraordinarily large.
As such, \cite{epoch2024canaiscalingcontinuethrough2030} conclude that it is not only technically feasible to continue scaling, but also plausible that AI developers make the investments required to enable this.

Forecasting the future rate of algorithmic efficiency gains is challenging. 
\cite{ho2024algorithmic} note that this rate appears to be closely tied to the level of investment, and the degree to which AI can substitute for human labor.
Nevertheless, we see no clear evidence that algorithmic efficiency gains should diverge from their historical rate of progress (a doubling of effective compute every eight months for LLMs), should compute and data scaling rates maintain their scaling trajectories.
Consequently, we believe that there is solid evidence for \textit{Claim 2}---it is plausible that increases in computation, data and algorithmic efficiency continue to grow at recent historical rates for the next five years.

\noindent \textit{Claim 3: Future scaling of computation and data, coupled with algorithmic advances, will continue to deliver meaningful increases in AI capabilities.}

In support of \textit{Claim 1}, we described many pieces of evidence that long-run AI capability improvements have been driven through large-scale increases in compute, data and algorithmic efficiency.
Reviewing this evidence, we observe that the current paradigm has accumulated a robust track record of transforming the ``inputs''  of compute and data into the ``output'' of AI capability improvements.
We believe it would be surprising if future scaling of these inputs did not continue to deliver meaningful increases in AI capabilities (though such an outcome is certainly possible).
Consequently, we think the balance of evidence provides reasonable support for \textit{Claim 3}, though with some uncertainty.

\subsubsection{Challenges for alternative paradigms}

Our remaining claims concern the difficulties faced by alternative paradigms in displacing the current paradigm.

\noindent \textit{Claim 4: The current paradigm enjoys a significant ``first mover advantage'' with respect to research expertise and capital investment over any alternative paradigm that emerges as a competitor in the future.}

To support \textit{Claim 4}, we first note that a diverse array of AI developers have made substantial infrastructure investments predicated on the continued success of the current paradigm.
As discussed above, these investments have led to an unprecedented $4\times$ per year scaling of training compute over the last 14 years~\citep{epoch2024trainingcomputeoffrontieraimodelsgrowsby45xperyear}.
Indeed, it is estimated that the dollar cost of training frontier systems has grown by a factor of $2.4\times$ per year between 2016 and 2024~\citep{cottier2024rising}.
As evidence of the scale of this investment, NVIDIA---the largest supplier of hardware accelerators for training frontier models---has grown by a factor of approximately 30 in the last 5 years\footnote{In late October, 2024 NVIDIA surpassed Apple as the world's most valuable company for the first time. As noted by~\cite{korinek2024economic}, public markets reflect public expectations of AI, and are therefore a useful, if somewhat noisy indicator for forecasting AGI. Consequently, in addition to providing evidence of tremendous demand for their products, the value of NVIDIA reflects a perception that the company stands to benefit from continued progress of AI in the current paradigm.}.
Note that investments at modern frontier scale require future planning and commitments that are costly to modify.
Since these investments specifically target the current paradigm, it will be difficult for a new paradigm to leverage them with the same efficiency.

A second key factor relates to research and engineering expertise.
The success of the current paradigm has created a strong market demand for particular sets of research and engineering skills---those that can be productively deployed across relevant portions of the research and development stack within the current paradigm.
To meet this demand, educational institutions and companies have trained a labor force with the corresponding skills.
As a concrete example of the strength of the demand, in 2023 Coursera's AI-related courses attracted 6.8 million enrollments \citep{coursera_blog_2023}, with the most popular course (provided by DeepLearning.AI) experiencing almost 400,000 enrollments to date.
Given their expertise, this growing labor force is best placed to contribute novel ideas and technical innovations within the current paradigm, further increasing its advantage over competitors.

\noindent \textit{Claim 5: If the current paradigm continues to deliver progress, there will be reduced incentive for research investments in alternative paradigms.}

In our arguments against \textit{Objection 1}, we described evidence suggesting that it is plausible that the current paradigm continues to deliver AI progress.
If this is the case, there are several reasons why there may be reduced incentive for research investments in alternative paradigms.
First, research investments in alternative, unproven paradigms entail greater risk relative to investments in the current paradigm, particularly as the latter continues to deliver improved AI capabilities.
Second, foundational research in alternative paradigms typically entails longer time horizons.
By contrast, research investments within the current paradigm promise benefits over much shorter periods. 
Third, as noted above, AI developers have made substantial investments in the infrastructure and labor force required to pursue progress within the current paradigm. 
A transition away from the current paradigm risks stranding these assets.

We note here a counterargument to \textit{Claim 5.}
If computation requirements continue to grow exponentially for frontier AI development, capital requirements will also grow.
This would increase the incentive to invest in alternative paradigms that can deliver AI capability improvements at reduced cost.
On balance however, we anticipate that the current paradigm will continue to attract the vast majority of research investment.

\noindent \textit{Claim 6: There is little evidence of mature alternatives that show close to the same degree of promise or rapid progress as the current paradigm.}
  
Our primary argument in support of \textit{Claim 6} is simple. 
We are not aware of mature (or indeed nascent) alternatives that appear to demonstrate near-competitiveness with the current paradigm of frontier AI development.
Two concrete examples of alternative paradigms whose dominance would violate our assumption include systems that make minimal use of the learning paradigm~\citep{buchanan1988fundamentals}, and mind uploading to create emulated people~\citep{hanson2016age}.
We note here that our argument is not decisive---history contains many examples of paradigm shifts~\citep{Kuhn1962TheSO} and we cannot claim exhaustive knowledge of the AI development landscape.

%% file: 03-assumptions/no-human-ceiling.tex
\subsection{No human ceiling for AI capability} \label{sec:superhuman}

\noindent Our second core assumption is the \textit{\noHumanCeilingAssumption}.
We define this to be the assumption that
advances in AI capabilities will not cease once they achieve parity with corresponding capabilities of the most capable human for a given task.
Put differently, we assume there is no ``human ceiling'' that sets an upper limit for AI capability.

\noindent \textbf{Implication for our approach to AI safety.}
The primary consequence of the no human ceiling assumption is that our approach to safety must leverage new AI capabilities as they become available in order to remain effective.
Initially, this will entail early adoption of AI assistance and tooling throughout our R\&D process---a process of augmentation, rather than replacement.
As capabilities continue to improve, we anticipate delegation of increasingly complex research and engineering tasks to AI assistance. 
Ultimately, in order to keep pace with advancing AI capabilities, we believe it will be necessary for the vast majority of the cognitive labor relevant to AI safety to be performed by AI.
This period of transition will correspond to a period of elevated risk, since delegation of tasks means that progress is likely to be fast compared to today.
This will necessitate decision making in periods that are relatively short on human timescales.
Consequently, this period will also elevate the importance of enabling humans to efficiently verify the artifacts of AI safety research~\citep{irving2024automation}.
To meet these challenges, our development of \textit{Amplified Oversight} algorithms aim to ensure that humans can continue to provide meaningful oversight as AI capabilities surpass those of humans (forward ref).
Our investments in \textit{Interpretability} (Section~\ref{sec:interpretability}) aim to further improve our ability to oversee future AI systems by strengthening our understanding of how they work.
As AI capability grows, there will be greater need for \textit{Alignment Stress Testing} (Section~\ref{sec:alignment-stress-tests}) to provide assurances.
More broadly, we target the development of robust \textit{Safety Cases} (Section~\ref{sec:alignment-safety-cases}) to form holistic arguments that guide our deployment of internal AI capability as an accelerant of AI safety research.

\noindent \textbf{Summary of supporting evidence:}
Our argument in support of the \noHumanCeilingAssumption takes the form of three claims.
Our first claim is that superhuman performance has been convincingly demonstrated in several tasks.
This provides a ``proof of concept'' that there exist concrete, clearly-defined tasks for which human capability does not represent a meaningful ceiling for AI.
Our second claim is that AI development exhibits a trend towards more generalist, flexible systems.
Consequently, we expect superhuman capability to emerge across an increasingly large number of tasks in the future.
Our third claim is simply that we observe no principled arguments why AI capability will cease improving upon reaching parity with the most capable humans.

\noindent \textbf{Supporting arguments and evidence}

\noindent \textit{Claim 1: Superhuman performance has been convincingly demonstrated in several tasks.}

Deep Blue~\citep{campbell2002deep}, the chess system that defeated world champion Garry Kasparov in 1997, represents an early example of an AI system achieving superhuman performance on a single task.
Since then, the capability of the most capable AI chess systems has continued to advance~\citep{ccrl4040}, reaching an Elo rating of 3643 in 2024.
For comparison, the highest Elo rating attained by a human chess player to date was 2882, achieved by Magnus Carlsen in 2014~\citep{chessdb_top100}.
This difference in Elo rating corresponds to a 98.8\% expected win rate for the AI system over the strongest human player of all time.
As a second example, the Watson system~\citep{ferrucci2010building} outcompeted two of the strongest human players in the TV quiz show ``Jeopardy!'' in 2011.
Superhuman performance has similarly been convincingly demonstrated in the games of Go~\citep{silver2016mastering,silver2017mastering} and Shogi~\citep{silver2018general}.
Moving beyond games, following~\citep{morris2023levels}, we consider AlphaFold~\citep{jumper2021highly} to represent superhuman performance on the task of predicting the 3D structure of a protein from an amino acid sequence more accurately than the most capable scientists.

Of course, in very narrow domains (such as floating point arithmetic), computer systems have long outperformed humans.
However, we note a trend towards domains of increasing complexity as more computation becomes available.
Go represents a substantially more challenging domain than chess, for example.
AlphaGo~\citep{silver2016mastering} was enabled not only by algorithmic innovation, but also by leveraging vastly more computation than was available to Deep Blue~\citep{campbell2002deep}.

In summary, there exist several tasks for which AI systems have convincingly achieved superhuman performance.
Recently, the collection of such tasks has expanded to include domains of significant complexity, such as the game of go and protein structure prediction. 

\noindent \textit{Claim 2: AI development exhibits a trend towards more generalist, flexible systems.}

Historically, many AI systems have been designed to achieve competence on narrow, specific tasks.
This approach, perhaps best exemplified by Deep Blue's mastery of chess~\citep{campbell2002deep}, enabled tractable progress through specialization.
More recently, however, AI development has witnessed a clear trend towards more general-purpose AI systems.
Systems such as GPT-3~\citep{brown2020language} and Gopher~\citep{rae2021scaling} demonstrated basic capabilities on a range of tasks that include translation, question answering, basic arithmetic and fact checking.
Successor systems such as GPT-4~\citep{achiam2023gpt}, Gemini 1.5~\citep{team2024gemini} and Llama 3.1~\citep{dubey2024llama} have extended this progress, attaining basic competency on tasks such as coding, mathematics and image comprehension.  
In contrast to predecessor systems that often required task-specific fine-tuning to elicit useful functionality~\citep{bommasani2021opportunities}, these systems are often capable of performing tasks in a zero-shot manner, without such adaptation.
There has also been a notable trend towards multimodal models that can process and integrate information across different modalities (for example, text, images, videos and code), further reinforcing the trend towards greater flexibility and generalization.

While these generalist systems are, for the most part, significantly below human expert capability on many tasks, there are clear signs that their capabilities are advancing. The o1 model~\citep{openai2024learning} is a single system that ranked in the 89th percentile on a benchmark of competitive programming questions from CodeForces and ranked among the top 500 students in the US in a qualifier for the USA Math Olympiad (AIME).
The same system also surpassed human PhD-level accuracy on~\citep{rein2023gpqa}, a benchmark of graduate-level physics, biology, and chemistry problems.
We view the o1 model as an example of the trend towards flexible, generalist AI systems while concurrently attaining capabilities that are approaching human expert performance.
More broadly, we observe a clear trend towards generalist, flexible AI systems with increasing levels of capability.

\noindent \textit{Claim 3: We observe no principled arguments why AI capability will cease improving upon reaching parity with the most capable humans.}

Our primary evidence in support of \textit{Claim 3} is somewhat anecdotal: we have searched, unsuccessfully, for principled arguments that AI capability will cease improving upon reaching parity with the most capable humans.  
Of course, we must acknowledge the limits of our knowledge and the possibility that such arguments exist.
We do, however, also note that many examples of superhuman performance observed to date~\citep{silver2016mastering,silver2018general,jumper2021highly} appear to use algorithmic approaches that differ substantially from those used by humans.
This lends credence to the idea that human capabilities do not necessarily place a constraint on the capabilities of AI systems.
Moreover, given the strong correlation between compute and AI capabilities to date~\citep{kaplan2020scaling,hoffmann2022training,achiam2023gpt,owen2024predictable,yuan2023scaling} and the clear possibility of continued scaling~\citep{epoch2024canaiscalingcontinuethrough2030}, we believe the balance of evidence points away from a cessation of AI progress at the level of the most capable humans.

%% file: 03-assumptions/timelines.tex
\subsection{The timeline of AI development remains uncertain} \label{sec:timelines}

Our third core assumption is the \textit{\uncertainTimelinesAssumption}.
We define this to be the assumption that the timeline for the development of AI is uncertain.
Consequently, a broad range of timelines, notably including short timelines, are plausible.

To anchor our timeline estimates, and to compare with existing AI timeline forecasts, we must characterize the level of AI capability that we are forecasting.
We use the definition of ``Exceptional AGI'' introduced by \citet{morris2023levels} for this purpose.
An Exceptional AGI is a system that has a capability matching at least 99th percentile of skilled adults on a wide range of non-physical tasks, including metacognitive tasks like learning new skills.
\citet{morris2023levels} identify this as the threshold at which many concerns relating to the most severe risks from AI are likely to emerge.
We use ``short timelines'' to refer to the development of an Exceptional AGI before the end of the current decade.

\noindent \textbf{Implication for our approach to AI safety.}
AGI safety research can vary in terms of its applicability to current systems. There is both development of anytime mitigations that can be applied whenever necessary, as well as more expansive foundational explorations that could take many years to bear fruit (but which hold the potential for greater safety benefits).
However, given the \uncertainTimelinesAssumption, it is important that frontier AI developers have an anytime safety approach to put in place in the case of short timelines. For this reason, in this paper we focus on mitigations that can be integrated into current frontier AI development, and leave more foundational explorations out of scope.


\noindent \textbf{Summary of supporting evidence:}
Our argument in support of the \uncertainTimelinesAssumption consists of two claims.
Our first claim is that the wide variety of existing AI forecasts lend credence to the claim that a broad spectrum of AI timelines are plausible.
Our second claim is that uncertainty over timelines is warranted, given that forecasting the development of technology is challenging.

\noindent \textbf{Supporting arguments and evidence}

\noindent \textit{Claim 1: Existing AI forecasts support a broad spectrum of timelines.}

Forecasting the arrival of AI that meets the threshold of an Exceptional AGI, or indeed of any future level of capability, is challenging.
Nevertheless, two salient pieces of evidence suggest themselves as building blocks for an AI timeline framework:
(1) the long-running and relatively smooth historical trend of price-performance improvements in computation;
(2) approximate bounds on the computation performed by the human brain. 
Consequently, empirical data spanning these axes has inspired a large number of forecasts for AI timelines.

Note that while many AI forecasts estimate the arrival of some notion of advanced AI capability, they often operationalize this concept slightly differently (for instance, while some focus on economic impact, others focus on skill level).
For each forecast below, we describe the AI capability threshold in question (though note that this capability threshold is not always defined precisely).
We refer the reader to \citet{aiimpacts2022list} and \citet{epoch2023literaturereviewoftransformativeartificialintelligencetimelines} for literature reviews of AI timelines and summarize several notable forecasts below.

Compute-centric forecasting was initially explored by \citet{good1970some} who projected the arrival of an ``intelligent machine'' (of comparable capability to humans) in 1993 ($\pm$ 10 years) by noting an existing estimate of human brain scale (approximately $10^{12}$ neurons) and suggesting that long-running historical trends of miniaturization were set to continue.
\citet{moravec1998will} provided a more explicit model relating human nerve volume to computation.
In particular, by extrapolating from estimated equivalences between the human retina and computer vision algorithms, Moravec predicted from market trends that affordable hardware necessary for matching human brain performance would likely be available in the 2020s.
This argument was extended by \citet{bostrom1998long} to incorporate considerations of economic and military incentives and make a case for the possibility of machine intelligence vastly surpassing human intelligence in the first third of this century.
\citet{kurzweil2005singularity} frames the trajectory of Moore’s law as part of a broader evolutionary process termed ``the law of accelerating returns'' and proposes 2029 as the date at which a particular instantiation of the Turing Test will be passed.
This Kapor-Kurzweil Turing test variant~\citep{KaporKurzweil2002} involves three human judges, three human foils and an AI.
Each judge conducts a sequence of four interviews (each two hours in duration) with the three human foils and the AI (24 hours of interviews are conducted in total).
The test is considered to have been passed if two or more of the three judges are fooled into believing that the AI is a human.
To provide an estimate of plausibility for this timeline, we note that on the Metaculus forecasting platform, the probability assigned by the community prediction to the outcome that the test will be passed has risen from 26\% in May 2020 to 80\% at the time of writing in late October 2024~\citep{metaculus_turing_test}.

\citet{cotra2020forecasting} conducted a particularly detailed timeline forecasting exercise that focused on the arrival of transformative AI (defined to be AI with impact comparable to the industrial revolution). 
Timelines were constructed by coupling estimates for the computation performed by the human brain (structured around ``biological anchors'' such as the learning performed within a human lifetime), predictions for algorithmic progress (relative to 2020), declines in computation cost and forecasts for the scale at which actors may be willing to deploy capital on a single training run.
The resulting forecast for transformative AI corresponds to a median estimate of 2052 and a 15\% probability of arrival by 2036.
Follow-up analysis conducted two years later by the same author shortened these estimates to a median estimate of 2040 and a 15\% probability of arrival by 2030~\citep{cotra2022twoyear}.
\cite{davidson2023compute} also builds on the training requirements estimated by \cite{cotra2020forecasting} and employs a semi-endogenous growth model to account for increasing R\&D spend, yielding a median estimate of 2043 for the arrival of transformative AI.

Contrasting with the ``inside view'' \citep{kahneman2011thinking} approach of \cite{cotra2020forecasting} and other strategies that focus their forecast on an explicit computational model of how AI may develop, \cite{davidson2021semiinformative} develops an ``outside view'' prediction of the arrival of AI that focuses on reference classes and Bayesian reasoning.
This approach, which excludes consideration of the specific ways in which AI development may occur, yields an 8\% probability that AGI will be developed by 2036.
Here, AGI is defined as ``computer program(s) that can perform virtually any cognitive task as well as any human, for no more money than it would cost for a human to do it.''

Another family of methods for future forecasts has sought to conduct questionnaires of expert AI researchers, with notable surveys conducted in 2016 \citep{grace2018will}, 2022 \citep{Grace2022Expert} and 2023 \citep{grace2024thousands}.
These surveys span a broad range of questions about AI development.
Of particular interest, the surveys cover the arrival of ``High-level machine intelligence'' (HLMI), which is defined to have been achieved when unaided machines can accomplish every task better and more cheaply than human workers. 
In 2016, the aggregate forecast estimated a 50\% chance of HMLI by 2061. 
This dropped by one year to 2060 in the 2022 survey, then by 13 years to 2047 in the 2023 survey.

To summarize the discussion above, we observe forecasts of AI development spanning a broad spectrum of timelines across a range of forecasting methodologies.
This diverse collection of forecasts includes support for short timelines, perhaps best exemplified by the \cite{metaculus_turing_test} community prediction indicating the high probability (80\%) that the Kapor-Kurzweil Turing test variant will be passed in 2029.

\noindent \textit{Are physical limits relevant?}
As discussed in \cref{sec:paradigm}, the scaling up of computation has been a critical driver of AI capability.
Since this computation must be implemented in physical hardware, it is worth considering whether physical limits are relevant to a discussion of AI timelines.
In the long run, physics imposes limits to exponential hardware trends such as Moore’s law \citep{moore1965cramming,moore1975progress}. 
By considering the construction of ``the ultimate laptop'', \citet{lloyd2000ultimate} suggests that Moore’s law (formalized as an 18 month doubling) cannot last past 2250.
\citet{krauss2004universal} consider limits on the total computation achievable by any technological civilization in our expanding universe - this approach imposes a (looser) 600-year limit in Moore’s law. 
However, since we are very far from these limits, we do not expect them to have a meaningful impact on timelines to Exceptional AGI.

\noindent \textit{Claim 2: Uncertainty over timelines is warranted since forecasting technology is challenging.}

Forecasting plays a key role in planning, but how much confidence can we place in technological timeline forecasts?
History is replete with examples of misprediction.
Just days after Nobel laureate Ernest Rutherford’s declaration that anyone claiming that nuclear reactions would provide a potent source of energy was talking ``moonshine'', Szilard conceived of the neutron chain reaction \citep{adams2013anniversary}.
In 1960, a second Nobel Laureate, Herbert Simon, wrote that ``machines will be capable, within twenty years, of doing any work that a man can do.'' \citep{simon1960new}. 
To date, AI forecasters have an underwhelming track record. 
\citet{armstrong2015we} analyzes a database of 95 AI predictions conducted prior to 2015. Broadly, they find that expert predictions contradict each other and are indistinguishable from non-expert predictions. 
Addressing two widely-discussed ``folk laws'', they find no support for the ``Maes-Garreau law''~\citep{Kelly_2007} which posits that AI experts exhibit a strong bias towards predicting the arrival of AI towards the end of their lifetime, but do find a clear prevalence of predictions that AI will arrive 15 - 25 years in the future. 
Moreover, there is a wide disparity between the views of prominent AI developers \citep{korzekwa2023views}.
As such, we believe caution is merited when placing significant mass on any particular future trajectory.

%% file: 03-assumptions/acceleration.tex
\subsection{The potential for accelerating capability improvement} \label{sec:acceleration}

\noindent Our fourth core assumption is the \textit{\potentialForAcceleratingImprovementAssumption}.
We define this to be the assumption that the use of AI systems could plausibly lead to a phase of accelerating growth, such that we should consider this a serious factor in our planning.
In more detail, we believe it is plausible that initial automation of scientific research and development would support the development of increasingly capable AI research systems that further accelerate research and development, producing a positive feedback loop.

The possibility of such an outcome has been imaginatively referred to as the question of ``Fizzle or Foom?''\footnote{These terms appear in various contexts relating to growth rates and AI, for instance~\citep{yudkowsky2013intelligence,cotra2023llms}.}
For clarity, we adopt the definitions of~\citet{erdil2024estimating}, in which
``\textit{Foom}'' describes the scenario in which \textit{AI systems produce a proportional improvement in AI software which results in a greater-than-proportional improvement in the subsequent AI-produced R\&D output and technological progress accelerates.}
``\textit{Fizzle}'' describes the scenario in which \textit{AI systems produce a proportional improvement in AI software which results in a smaller-than-proportional improvement in its subsequent AI-produced R\&D and technological progress decelerates.}
In this terminology, our assumption corresponds to the claim that a \textit{Foom} scenario is plausible.

\noindent \textbf{Implication for our approach to AI safety.}
A period of accelerating growth could radically increase the velocity of AI development.
As a consequence, there may be very little calendar time to react to new developments and make good decisions with respect to risk mitigation.
Consequently, our risk mitigation strategy must also ultimately be significantly accelerated through AI assistance to ensure we retain the ability to react to, and address, novel risks as they arise.

\noindent \textbf{Summary of supporting evidence}

Our argument in support of the \potentialForAcceleratingImprovementAssumption takes the form of three claims.
Our first claim is that, while there is a significant diversity of views, there is support within the economics literature for the plausibility of very fast (in particular, hyperbolic) growth.
Our second claim is that, while the picture is mixed, there is some evidence that the returns to software R\&D could be plausibly be high enough to enable a ``software-only singularity'' in which sufficiently high returns to R\&D effort lead to hyperbolic growth~\citep{davidson2023compute}.
Our third claim is that AI researcher surveys are consistent with the possibility of rapidly accelerating growth.

\noindent \textbf{Supporting arguments and evidence}

\noindent \textit{Claim 1: The economics literature supports a diversity of perspectives on accelerating growth, including the plausibility of extremely rapid growth.}

At present, there remain fairly divergent views among researchers on the plausibility of extremely rapid growth, particularly when considering the time period of the next few decades.
\cite{nordhaus2021we} suggests that the key factor determining the feasibility of extremely rapid growth is the degree of substitutability between information and conventional, non-information inputs. 
By examining a suite of diagnostic metrics over empirical economic data to date, Nordhaus concludes that a period of rapid acceleration lies far in the future (100 years or more for economic growth rates reaching 20\%).  \cite{christensen2018uncertainty} survey expert economists, also finding that near-term explosive growth is unlikely (the aggregate median forecast is for 2\% global growth between 2010 and 2100, with a slowdown in the second half of the century).  Over a shorter time frame, \cite{acemoglu2024simple} similarly predicts limited impact (forecasting for instance, that US GDP growth due to AI in the next 10 years will be no greater than 0.9\%).

Contrasting with this perspective, there is also literature that endorses the prospect of substitutability of capital for labor, stretching back to \citep{keynes1930essays}. \cite{hanson2001economic} illustrates that, under such an assumption, explosive growth becomes highly plausible. This outlook is consistent with the earlier work of \cite{hanson2000long}, which observes that modeling growth over long historical timescales points toward the potential for explosive growth towards the middle of the twenty-first century. 
More broadly, \cite{sandberg2013overview} note that both existing endogenous growth models and Hanson's models support the conclusion that if some form of mental capital (which could be embodied in humans or AI) becomes relatively cheap to copy, then it is probable that extremely rapid growth will follow.
\cite{erdil2023explosive} analyze 12 arguments for and against the possibility of ``explosive growth'' (an acceleration of roughly an order of magnitude) caused by AI, offering even odds of such growth by the end of this century but highlighting high levels of uncertainty around this estimate. \cite{trammell2023economic} survey a number of economic models for the effects of AI and show that dramatic increases in growth rate are possible across a range of models. \cite{davidson2023compute} builds on the estimates for the training requirements of transformative AI (AI with impact comparable to the industrial revolution) produced by \cite{cotra2020forecasting} and employs a semi-endogenous growth model to forecast the time between AI capable of automating 20\% of human cognitive labor (weighted by 2020 economic value) and $\sim$100\% of human cognitive labor. 
This yields a median estimate of 3 years for the period of transition, corresponding to an increase of 4 orders of magnitude in ``effective compute'' (a quantity of compute that is modulated to take account of increasing software efficiency over time). 
Recent modeling by~\citet{erdil2025gateintegratedassessmentmodel} that draws on empirical scaling laws and semi-endogenous growth theory and models changes in compute, automation and production supports the plausibility of very rapid growth in Gross World Product (e.g. exceeding 30\% per year in 2045) when adopting parameters from empirical data, existing literature and reasoned judgment.
\citet{moorhouse2025preparingforthei} conduct analysis on the feasibility and consequences of an intelligence explosion and estimate that there is a greater than 50\% chance that AI leads to at least a century of technological progress in a decade.


Two key themes emerge from this literature \citep{davidson2021could}. 
The first is that idea-based models of long-run growth suggest that it possible for AI to induce explosive growth by automating R\&D and inducing an ideation positive feedback loop. The second is that a broad range of economic models predict explosive growth if AI becomes capable of enabling capital to substitute for labor.

\noindent \textit{Claim 2: The returns to software R\&D could plausibly be high enough to support hyperbolic growth.}

From the earliest days of the artificial intelligence field, researchers have recognized the theoretical possibility that achieving some threshold level of AI progress may precipitate a rapid acceleration in further development. 
Writing in 1959, I.J. Good noted that:
``Once a machine is designed that is good enough, say at a cost of \$100,000,000, it can be put to work designing an even better machine. At this point an ``explosion'' will clearly occur; all the problems of science and technology will be handed over to machines and it will no longer be necessary for people to work.''
When considering the form that such a transition period may take, \cite{good1959speculations} speculates that there may be relatively little evidence of the acceleration in advance:
``It seems probable that no mechanical brain will be really useful until it is somewhere near to the critical size. If so, there will be only a very short transition period between having no very good machine and having a great many exceedingly good ones.''
This perspective is echoed by \cite{bostrom1998long} who suggests that, in addition to purely technical considerations, the marginal utility of AI will increase dramatically as it reaches human-level, triggering further increases in funding.  

Central to understanding this possibility is the process of estimating the ``returns to R\&D'', a challenging task that has attracted attention within the economics literature~\citep{bloom2020ideas}.
In the context of AI development, there has been particular interest in understanding the plausibility of an ``intelligence explosion'' of the kind defined by~\cite{good1959speculations} and the dynamics that such an event could entail~\citep{chalmers2010singularity}.
\citet{yudkowsky2013intelligence} formalizes the notion of such an event through an economics lens, introducing the notion of the \textit{returns on cognitive reinvestment}, defined as ``the ability to invest more computing power, faster computers, or improved cognitive algorithms to yield cognitive labor which produces larger brains, faster brains, or better mind designs''.

A particular challenge here is the difficulty of obtaining good estimates for returns to research effort relating to frontier AI development.
Perhaps the most relevant proxy for AI R\&D for which it is possible to obtain reasonable empirical data is software R\&D.
The most detailed study to date of the empirical returns to software R\&D was produced by~\citet{erdil2024estimating} (conducted as part of their extensive methodological survey of approaches for estimating idea production in growth models).
We briefly summarize the key points of their analysis below.

\cite{erdil2024estimating} base their analysis on the semi-endogenous growth model of~\citep{jones1995rd}, which can be written as follows:

\begin{align}
    \frac{1}{A(t)}\frac{dA(t)}{dt} = \theta A(t)^{-\beta} I(t)^{\lambda}.
    \label{eqn:jones}
\end{align}

where $I(t)$ captures some measure of inputs to R\&D, $A(t)$ is some measure of efficiency or ideas, $\beta$ indicates whether there are increasing or diminishing returns to new ideas and $\lambda$ characterizes the returns to scale on the inputs.
In their wide-ranging study of research productivity, \cite{bloom2020ideas} use the ratio $r = \frac{\lambda}{\beta}$ to define the notion of ``returns to research''. This ratio has important consequences for the asymptotic behavior of \cref{eqn:jones}.
If we suppose that R\&D can be fully automated and we assume access to a fixed stock of compute, $c$, then the input to R\&D is simply $I(t) = c A(t)$.
We can substitute this into \cref{eqn:jones} to obtain

$$\frac{dA(t)}{dt} = \theta c^\lambda A(t)^\gamma$$

where $\gamma = \lambda - \beta + 1$. 
We see that, under this model, hyperbolic growth will occur when $\gamma > 1$, or equivalently, when $r = \frac{\lambda}{\beta} > 1$.
Since we assumed a fixed stock of compute, \cite{davidson2023compute} refers to hyperbolic growth in this regime as a ``software-only singularity''.
Note that this model is highly simplistic in two key ways. 
First, as highlighted by~\cite{erdil2024estimating}, this model assumes that the efficiency of AI systems is driven solely by cognitive labor whereas in practice it is likely that experiments (possibly entailing large-scale use of computation) will also play a significant role.
This assumption will lead to overly optimistic estimates of growth.
Second, in practice, compute is also likely to grow (rather than remaining fixed), and as such this growth model can be considered conservative, at least with respect to this assumption.
Acknowledging these limitations of the model, it remains of significant interest to estimate the value of $r$ from empirical data in relevant domains.
To this end, \cite{erdil2024estimating} estimate $r$ in several settings.
In computer chess, where data is most readily available, the authors estimate $r=0.83$ (with a standard error of $0.15$). 
In other domains where data is harder to obtain (computer vision, SAT solvers, linear programming and reinforcement learning), the median estimate of $r$ falls above 1, but not in a manner that is statistically significant.
In summarizing their findings, \cite{erdil2024estimating} observe that the returns to software R\&D may be sufficient to produce hyperbolic growth in software, but the evidence is not conclusive. 
As such, we believe that it remains plausible that the returns to software R\&D could be sufficient to support hyperbolic growth, and we should be prepared for this eventuality.

\noindent \textit{Claim 3: AI researcher surveys support the possibility of rapidly accelerating growth.}

\noindent A series of surveys of large numbers of AI researchers~\citep{grace2018will,Grace2022Expert,grace2024thousands} asked respondents about the probability that technological progress becomes more than an order of magnitude faster within 5 years of achieving ``high-level machine intelligence'' (defined to have been achieved when unaided machines can accomplish every task better and more cheaply than human workers). 
Since 2016, the majority of respondents indicated that they believe that it is one of ``quite likely,'' ``likely,'' or an ``about even chance'' (with the remaining two choices being ``unlikely'' and ``quite unlikely'').
Interestingly, responses to questions relating to the probability of an intelligence explosion have remained fairly stable across time~\citep{grace2024thousands}.
The median probability that surveyed AI researchers assigned to explosive growth two years after achieving high-level machine intelligence was 20\%~\citep{grace2024thousands}. 
While by no means definitive, this result suggests support among AI researchers for the possibility of rapidly accelerating growth.

\noindent \textbf{Summary:} As we discussed in \cref{sec:timelines}, forecasting the timeline (or indeed the eventual likelihood) of accelerating growth is fraught with difficulty.
Nevertheless, considering the evidence above, we believe there are strong arguments for the possibility of rapidly accelerating capability development. 
Such a world would likely function significantly differently from the world of today. 
It could entail, for example, not only full automation of factories, but full automation of the construction of factories, a process accompanied by correspondingly large increases in global energy consumption. 
Moreover, we believe it is plausible that explosive growth could be triggered in the near future (see \Cref{sec:timelines} for our discussion of timelines), with significant implications for AI safety.
As highlighted above, this perspective motivates the need for an anytime safety approach that is responsive to early indicators of such developments.

%% file: 03-assumptions/continuity.tex
\subsection{Approximate continuity} \label{sec:continuity}

\noindent Our fifth core assumption is the \textit{\continuityAssumption}.
We define this to be the assumption that general AI capabilities will scale fairly smoothly and predictably with the availability of computation, R\&D effort, and data, and in particular there will not be large discontinuous ``jumps'' in general AI capabilities with respect to these inputs.\footnote{Sufficiently drastic acceleration would have similar implications as a ``jump'', and so would count as a ``jump'' for our purposes. Note that for the \continuityAssumption, the acceleration must be in rate of general AI capabilities relative to computation, R\&D effort, and data, \emph{not} relative to time.}
Importantly, however, due to the \potentialForAcceleratingImprovementAssumption, we do not make any such assumption on the rate of AI progress with respect to calendar time.

\noindent \textbf{Implication for our approach to AI safety.}
A key implication of the \continuityAssumption is that it enables us to iteratively and empirically test our strategies and to detect any flawed assumptions or beliefs that only arise as capabilities improve.
As a consequence, our technical approach to AI safety does not need to be robust to arbitrarily capable AI systems.
Instead, our approach can focus on currently foreseeable capability improvements (as discussed in \Cref{sec:evidence-dilemma}), without necessarily being robust to AI systems with greater capabilities.
More broadly, our approach is underpinned by the expectation that business-as-usual scaling will not produce a large discontinuity in AI capability. 
If we can anticipate when model capabilities will develop, we can build up testing and control infrastructure in advance to cover the relevant risks, such as improving misuse detection in APIs in response to increased offensive cybersecurity capabilities (\Cref{sec:misuse}), or hardening sandboxing ahead of autonomous weight exfiltration capabilities (\Cref{sec:misalignment}).

Note that the \continuityAssumption does not obviate the need for foresight.
Rather, it justifies why it is acceptable to rely on extrapolations of trends in AI capabilities as a form of foresight, rather than preparing for arbitrarily powerful capabilities that could emerge at any point.
In addition, to ensure that we are extrapolating the right trends, it is important that we achieve a good understanding of the AI capabilities that exist at any given time.
Iterative deployment is particularly valuable for this: even if an AI company doesn't notice a particular capability in their models, the combined efforts of its users are likely to discover the capability.

The more unpredictable performance improvements are, the wider the required safety margins.
While it may be feasible to launch emergency crash efforts to mitigate some surprises after they are observed, this is expensive and unsustainable as a general strategy.
In the limit of unpredictability, it becomes economically infeasible to adequately protect against possible increases in capability. 
That said, we do not require perfect predictability.
Sudden improvements in capability are acceptable so long as they are:
\begin{enumerate}
    \item \textit{Not too large.} The model does not go from completely incapable to reliably capable of performing a task in a single jump. While a moderately large jump can present some risk, our primary risk areas (Section~\ref{sec:risk-areas}) depend on execution of complex plans, which requires reliable performance. 
    \item \textit{Not too frequent.} Surprises may happen occasionally, but are rare enough that it’s feasible to deal with the few that do happen at reasonable expense.
    \item \textit{Not too general.} A sudden improvement in capabilities does not happen across the board, but is localized to a narrow capability. If the model’s performance spikes only on a specific capability, it is probable that other capabilities become bottlenecks to accomplishing real-life tasks, limiting the scope of the risk.
\end{enumerate}

\noindent \textit{Automated ML R\&D.}
Expert-level machine learning (ML) R\&D performance is a particularly notable threshold because of its potential to enable recursive AI improvement.
This may lead to significant acceleration in capability improvement.
If this happens, there is a significant chance that it will be a large discontinuous jump with respect to time.
However, the mechanism for the acceleration of general capabilities is precisely that there is acceleration in the \emph{inputs} to AI progress, particularly (automated) R\&D effort.
So, the \continuityAssumption will still hold in such a setting.
As discussed in \Cref{sec:acceleration}, we will likely need our safety planning and monitoring to also be significantly assisted by AI.
Then, a performance discontinuity in calendar time trends would look more on trend when considering the amount of cognitive labor spent in analyzing and preparing for it.

\noindent \textbf{Summary of supporting evidence:}
Our argument in support of the \continuityAssumption takes the form of four claims.
Our first claim is that, looking at the reference class of large jumps in metrics of technological progress in general, large discontinuous jumps are rare.
Our second claim is that empirical evidence to date suggests that business-as-usual scaling tends not to produce sudden, large jumps in general capabilities, as measured by sufficiently broad benchmarks.
Our third claim is that while unpredictable capability gains can occur on narrow tasks, this occurs relatively infrequently
and performance almost never jumps directly from chance to >90\% in one iteration of scaling. 
Our fourth claim is that specific proposals for how future discontinuities in capabilities could arise do not seem likely. 

\subsubsection{Base rates and conceptual arguments for large discontinuous jumps}

\noindent \textit{Claim 1: Taking the outside view, large discontinuous jumps in highly optimized domains are rare.}

As discussed in \Cref{sec:drivers-of-ai-progress}, significant effort is being applied to improve general AI capabilities, to the tune of billions of dollars per year and thousands of researchers, and this effort will likely increase in the future. 
When so much effort is applied to optimize a variable of interest, we may expect that the research will divide into many smaller streams that tackle particular subproblems.
Then, large discontinuous jumps in any subproblem will often not cause a jump in the overall variable of interest.
When subproblems are primarily complements to each other, then if one subproblem is solved, other subproblems will become a bottleneck for overall progress, limiting the impact on overall progress.
When subproblems can substitute for each other, then overall progress becomes a sum of many contributions.
In practice, we expect progress on substitutable subproblems to be at least partially decorrelated. Consequently, the variance of the total progress reduces, decreasing the chance of a large discontinuous jump in overall progress.
In addition, researchers will attempt to pick the low-hanging fruit first, making it ever harder to identify a key insight that produces a large discontinuous jump even within a subproblem.

This argument relies only on the assumption that significant effort is applied to optimize a variable of interest. It is thus possible to empirically check how well the argument applies in other such domains to establish an ``outside view'' \citep{kahneman2011thinking} reference-class base rate estimate.
\cite{grace2021discontinuous} conduct a search for historical cases of extremely rapid progress on metrics that people were optimizing for. 
The study documents several events that represent a discontinuity in progress equivalent to more than a century at previous rates in at least one metric of interest. 
However, given their extensive search for examples, \cite{aiimpacts2020likelihood} conclude that the base rate probability for an AI progress discontinuity is relatively low, and further suggest that the arguments suggesting that AI is fundamentally different in character from other technologies are not compelling (though not decisively so). 

\noindent \textit{Impacts of radical technical changes.}
This outside view may not apply given a sufficiently large paradigm shift~\citep{Kuhn1962TheSO}, which could create an entirely new set of subproblems and a qualitatively different rate of progress. This would pose a challenge to the \continuityAssumption, but as discussed in \Cref{sec:paradigm}, we would need to fundamentally reconsider our approach in this case anyway. 

It is also possible to have radically different model training techniques, that nonetheless stay within paradigm. 
With such changes, we should be open to radically different performance and scaling behavior. 
While the new technique will likely follow its own scaling law (as, for example, when comparing transformers with LSTMs~\citep{kaplan2020scaling}), it could represent a discontinuous improvement on the current baseline. 
Nonetheless, due to the arguments above, we consider this to be unlikely, though do not rule it out entirely.

\noindent \textit{Features of the search space.} \cite{yudkowsky2013intelligence} argues that there are many aspects of the search space for intelligent AI designs that could lead to compounding returns on cognitive investment ultimately leading to acceleration that is functionally equivalent to a large discontinuous jump with respect to calendar time. \cite{yudkowsky2008recursive} identifies five factors in particular: cascades, cycles, insight, recursion, and unknown unknowns.
Importantly, we agree that the positive feedback loops created by cycles and recursion can lead to drastic acceleration with respect to \textit{time} (\Cref{sec:acceleration}).
We expect that all five factors will play a significant role in overall capabilities progress, but will not lead to jumps with respect to \textit{inputs} for the reasons given above.

\subsubsection{Empirical evidence from existing AI progress}

\noindent \textit{Claim 2: General capabilities tend not to show large, sudden jumps.}

Abrupt jumps in AI capabilities across many tasks produce risks that are more challenging to mitigate than those produced by smooth gains in capabilities. 
Such jumps can be measured using broad benchmarks that cover many behaviours, or by considering scores from many benchmarks, as is standard practice. \citet{owen2024predictable} find that aggregate benchmarks (BIG-Bench~\citep{srivastava2023beyond}, MMLU~\citep{hendrycks2020measuring}) are predictable with up to 20 percentage points of error when extrapolating through one order of magnitude (OOM) of compute. 
\cite{gadre2024language} similarly find that aggregate task performance can be predicted with relatively high accuracy, predicting average top-1 error across 17 tasks to within 1 percentage point using 20$\times$ less compute than is used for the predicted model. 
\cite{ruan2024observational} find that 8 standard downstream LLM benchmark scores across many model families are well-explained in terms of their top 3 principal components. 
Their first component scales smoothly across 5 OOMs of compute and many model families, suggesting that something like general competence scales smoothly with compute. 

\noindent \textit{Claim 3: While jumps in task-specific performance have been observed, dramatic jumps are rare and may be better explained by measurement artifacts.}

The literature has many examples of tasks on which increasing scale appears to make little difference to performance before abruptly showing significant gains, a phenomenon referred to as ``emergence''.\footnote{The phenomenon of emergence has received considerable attention in other disciplines, most notably in critiques of scientific reductionism~\citep{anderson1972more}.} 
\cite{wei2022emergent} define the emergent abilities of LLMs to be \textit{abilities that are not present in smaller-scale models but are present in large-scale models; thus they cannot be predicted by simply extrapolating the performance improvements on smaller-scale models.} 
These occur in both natural tasks~\citep{srivastava2023beyond} and in tasks adversarially selected for unusual scaling trends~\citep{mckenzie2023inverse}. 
Even when overall benchmark performance is roughly predictable, individual task performance often is not \citep{owen2024predictable}. 

\emph{Few dramatic gains.} However, even on individual tasks, emergence is observed only rarely with dramatic gains in metrics. 
On most tasks where emergence has been reported, jumps do not exceed 50 percentage points. 
Emergence does not typically result in a model going from chance accuracy to >90\%, which would be needed for the most concerning risks to take us by surprise. 
After searching the emergence literature, we found only a few examples of dramatic gains of this magnitude. 
One example is the performance of GPT-3 on MAWPS, a math word problem dataset \citep{wei2022chain, koncel2016mawps}, which occurs when increasing the training compute by more than an order of magnitude. 
Another is GPT-4 on Hindsight Neglect, a task from the Inverse Scaling Prize~\citep{mckenzie2023inverse} deliberately chosen to be adversarial to LLMs~\citep{achiam2023gpt}. 
The total computation used for GPT-3.5 and GPT-4 is unknown, but based on estimates provided by \cite{epoch2024notablemodels}, the performance gain on Hindsight Neglect corresponded to an increase of training compute of approximately one order of magnitude. 

As noted by~\cite{ruan2024observational}, one factor complicating the study of scaling laws and emergent capabilities (such as those studied by~\cite{wei2022emergent}) has been the sparsity of data points.
In particular, when only 5 data points are used to span many orders of magnitude in scale, phenomena can appear discontinuous even when the underlying phenomena are smooth.
\cite{ruan2024observational} revisit the results of~\cite{wei2022emergent} and demonstrate that capabilities previously identified as emergent can indeed be accurately predicted with their methodology.
Similarly, the two examples of dramatic gains above involved an order of magnitude increase in training compute.
In practice, it is rare though not unheard of for new models to have such a dramatic increase in training compute relative to existing models~\citep{epoch2024notablemodels}.

\emph{Emergence as measurement artifacts.} Recent work~\citep{schaeffer2023emergent,schaeffer2024predicting} has argued that the instances of dramatic emergence observed so far can mostly be explained as artifacts of the metrics and transformations used to measure performance. 
\cite{schaeffer2023emergent} find that the majority of instances (model/task pairs) of possible emergence in BIG-Bench tasks use multiple-choice and exact match accuracy as the metric. Both of these are nonlinear with respect to the underlying probability of generating the correct token, resulting in trends that appear discontinuous when coarsening along the x-axis. Choosing a linear metric instead (brier score or edit distance, respectively) results in smoother and more predictable scaling trends, even on individual tasks. 
\cite{schaeffer2024predicting} identify an additional factor that degrades the correlation between pretraining loss and downstream performance metrics: in multiple-choice settings, accuracy is influenced by how the model spreads probability mass over incorrect answers. 

One strategy for increasing predictability is to increase evaluation score resolution by increasing the evaluation set size and taking many more samples from the model~\citep{schaeffer2023emergent,hu2023predicting}. 
Given a large number of samples, modern LLMs often exhibit impressive recall (the ability to produce at least one correct sample) on complex tasks~\citep{brown2024largelanguagemonkeysscaling}. 
Another option is to only consider task instances which all models under consideration are capable of passing at least once~\citep{achiam2023gpt}, though this restricts the class of tasks we are able to make forecasts about to those with a nonzero pass rate today. 
While we think a better science of measurement will help make task performance more predictable, better metrics are not yet a complete solution. 
Many tasks (e.g. pass rate in code generation) exhibit related challenges and do not have straightforward alternative measurements. 

Even if existing cases of emergence could have been predicted through alternative metrics, it is not clear that this supports the \continuityAssumption.
Often for ``general AI capabilities'', we ultimately care about the nonlinear metric, not its linear cousin, and it is not clear we can predict one from the other. 
One of the best strategies for identifying novel capabilities is to deploy the model, and see how it is used in practice, which depends on its utility, which will be tied to the nonlinear metric. 
Thus, a focus on linear metrics may be hindsight bias, where we identify the linear metrics only after observing emergence on the natural, nonlinear metric. 

\noindent \textit{Caveat: elicitation.}
Many academic studies on emergence keep the scaffold fixed and change scale, to isolate the effects of scaling.
While this often looks continuous on aggregated benchmark performance \citep{suzgun2022challenging}, it may be that incorporating improvements in scaffolding or other posttraining improvements would suggest larger jumps.
More broadly, many forms of post-training can yield increases in capability at relatively low computational cost~\citep{davidson2023ai}.
Within a lab, deploying new elicitation methods is a deliberate choice. However, if the new elicitation technique is discovered outside of the lab, works with mere API access, and substantially boosts already-deployed models, this is potentially destabilizing, as users suddenly (and widely) gain access to new capabilities that have not passed safety testing.

\subsubsection{Conceptual arguments for future discontinuities}

\noindent \textit{Claim 4: Specific proposals for how discontinuities could arise seem unlikely.}

Even if the base rate for discontinuities is low and the empirical evidence with AI capabilities so far supports the same conclusion, it may be that there is some difference about future AI capabilities that wouldn't apply now and doesn't apply in the reference class used for our base rate, that implies that a large discontinuous jump is more likely than we might have otherwise thought.
\cite{aiimpacts2020likelihood} and \cite{christiano2018takeoff} survey a variety of arguments but ultimately conclude that these arguments are weak.

\noindent \textit{Threshold effects.} The arguments we find most compelling are those on \textit{threshold effects} that would arise with future AI capabilities.
For example, there may be some level of reasoning coherence that unlocks qualitatively different capabilities ``all at once''.
Such thresholds are not uncommon: for example, some kinds of scaffolding or elicitation techniques, such as Chain of Thought (CoT), only begin showing gains past a certain scale threshold \citep{wei2022emergent, suzgun2022challenging}.
It is worth noting, however, that the threshold represents the point where CoT overtakes baseline elicitation approaches; aggregated performance of CoT improves smoothly with scale.
Overall, given that threshold effects certainly exist in other domains as well but discontinuous progress is nonetheless rare, we expect that the impact of crossing thresholds is likely to be minimal.

\noindent \textbf{Summary:}
Surveying the literature, we see that a number of prior works have reported significant jumps in task-specific behavior.
Moreover, caution is merited when forecasting future capabilities in the absence of a theory that precludes discontinuous performance gains.
Nevertheless, very large jumps in performance appear to be rare, particularly when measurements are taken at relatively close increments in the scaling up of computation.
We also observe that general capabilities tend not to show large, sudden jumps.
On balance, we believe the evidence in support of the \continuityAssumption is relatively robust.

%% file: 03-assumptions/benefits.tex
\subsection{Benefits of AGI}\label{sec:benefits}

The primary focus of this document is to outline an approach to mitigate the risks of significant harm associated with AGI.
Our core motivation for doing so is that we believe AGI has tremendous benefits that outweigh these risks, provided appropriate precautionary measures are taken.
In this section, we highlight several important benefits that we believe AGI will bring.
These include \textit{raising living standards across the world} (\Cref{benefits:living-standards}), \textit{deepening human knowledge} (\Cref{benefits:deepening}) and \textit{lowering barriers to innovation} (\Cref{benefits:lowering-barriers}).

\noindent \textbf{Implication for our approach to AI safety.} It is easy to achieve safety if that is the only goal to pursue. For example, harms from AGI would not occur if AGI was never built in the first place. However, this would be problematic in its own right, as it would entail giving up on the massive benefits that AGI can bring. Our approach has been chosen to achieve safety without unduly giving up on the benefits of AGI. For example, many dangerous capabilities are \emph{dual use}, enabling benefits as well as harms through misuse. Our approach seeks to distinguish between these two cases and only block cases of misuse, so that the benefits can still be achieved.

\subsubsection{AGI could raise living standards across the world} \label{benefits:living-standards}

AGI, if deployed safely, widely and cost-effectively, has the potential to raise living standards across the world.
We note two particular ways in which this could be achieved.
First, through driving economic growth and prosperity by contributing to faster, more cost-effective innovation.
Historically, innovation has yielded significant social returns, while economic growth has correlated with positive outcomes across a broad range of indicators spanning areas such as education, health and general welfare---a correlation that we expect to continue.
Second, and more directly, AGI could raise the living standards across the world by improving education and healthcare outcomes.

\noindent \textbf{Summary of supporting arguments.}
To support our claim, we first briefly summarize research suggesting that there are significant social returns to innovation.
We next describe how, by accelerating innovation, AGI could yield significant economic growth, and outline how growth in turn correlates with positive societal outcomes.
We summarize evidence suggesting that there is enormous scope for improvement in the current state of global education, and that AGI could contribute to this improvement.
Finally, we outline the benefits of AGI for healthcare through pathways such as drug discovery.

\noindent \textbf{Supporting arguments and evidence.}
It is challenging to provide a precise estimation of the social returns to innovation and R\&D, since such a task must confront the difficulty of measuring quantities such as the output of R\&D intensive industries~\citep{griliches1979issues,hall1996private}.
Nevertheless, analyses conducted to date suggest that innovation and R\&D yield very substantial social returns.
\citet{jones2020calculation} suggest that even under conservative assumptions, it seems likely that the average social return is at least \$4 for every \$1 spent.
Moreover, if one accounts for international spillovers, health benefits and inflation bias (overestimates of inflation, and thus underestimates of real GDP growth), this return plausibly rises to over \$20 per \$1 spent.
By enhancing our capacity for problem-solving and discovery across all fields, and thereby increasing the efficiency of knowledge production, AGI could dramatically improve the social return on investment.
While primarily focused on the risks of the technology, \cite{russell2022if} posits that, even absent the more ambitious uses of future AI (such as extending the human lifespan), it is plausible to consider a scenario in which AGI raises the living standard across the Earth to levels considered respectable (taken to be the eighty-eighth percentile) in a modern developed country.
This would correspond to a tenfold increase in GDP (similar to gains achieved globally between 1820 and 2010)---income with an approximate economic value (assuming a discount factor of 5\%) of 13,500 trillion USD.

As a ``technology of technologies'', we can analyze the expected impact of AGI at two different levels.
First, zooming out, we can consider the aggregate of use cases and applications with large scale positive impacts on entire fields that are critical to human development.
Second, we can examine how particular algorithms or applications can help address specific challenges. 
We discuss these next.

\noindent \textit{Macro-level impacts on economic growth and productivity.}
We believe AGI is poised to generate growth.
This is important, because growth correlates with positive outcomes on a number of important attributes such as health outcomes, education, and general welfare. 
Indeed, economic growth appears to drive a substantial part of human welfare today.
For example, per-capita GDP is strongly correlated with the Human Development Index~\citep{suvsnik2017correlation}.
GDP per capita is also correlated with factors such as life expectancy~\citep{owid-life-expectancy}, self-reported life satisfaction~\citep{owid-life-satisfaction-gdp-2023} and literacy~\citep{owid-literacy-rate-gdp-2023}.
The growth-enhancing effect is an important motivation for AGI, and a plausible causal nexus for positive societal impacts and human welfare improvements. 

\noindent \textit{Micro-level impacts across key sectors, from health to education.}
We can also consider how AGI might benefit more discrete domains or applications, and how these in turn can enhance various social indicators and economic growth. 
This is particularly pertinent to the Global South: many drivers that improve agricultural productivity, education, and skills remain critical for continued transformation and growth in Low- and Middle-Income Countries~\citep{tadepalli2023economicgrowth}. 
This includes technologies that reduce labor demand, expanding access to higher education, and other important factors. 
We believe AGI can play an important role in lowering the barriers to such drivers, focusing in particular on education and healthcare.

\noindent \textit{Education.} A large fraction of the world's youth currently lack basic skills needed to participate effectively in the modern global economy.
For instance, even prior to the COVID-19 pandemic, an estimated 57\% of ten-year-olds in Low- and Middle-Income Countries were unable to read a simple text aimed at younger students~\citep{worldbank2022learningpoverty}.
If all children worldwide were able to achieve at least basic skill levels (equivalent to PISA Level 1 proficiency), by 2100 global GDP would be 56\% higher than under status quo trajectories.
Over the remainder of the century, this improvement represents an estimated 732 trillion USD in additional GDP~\citep{gust2024global}.

AGI could provide personalized, adaptive learning experiences, making high-quality education more accessible and effective for people of all backgrounds. 
For example, AI systems can potentially provide personalized attention to large numbers of students simultaneously, making high-quality individualized instruction more widely available.
This could be particularly impactful, given the notable benefits of personalized tuition on learning outcomes~\citep{bloom19842}.
While such solutions will inevitably be imperfect at first, their complementarity to existing systems may prove beneficial to students across the world.
There are early signs that lend credence to such predictions: \cite{kumar2023math} conduct a large scale, pre-registered controlled experiment using GPT-4 in tutoring and find that practicing with the help of GPT-4 significantly improved performance on SAT math problems.
Other research has shown that AI-enhanced personalized learning systems can improve learning outcomes and experiences in various subjects, from computer programming to mathematics~\citep{zhang2021ai}.
AGI will likely also improve educational outcomes through indirect impacts as well, for example by automating grading when appropriate and thereby freeing up valuable teacher time~\citep{henkel2024can}. 
An important point to note here is that the full potential of AGI in education extends far beyond direct instructional roles; its integration across back-end processes, ideation and non-key tasks can all contribute to improvements in educational metrics. 

\noindent \textit{Healthcare.} AGI could have significant impacts on healthcare by enabling more accurate diagnoses, personalized treatment plans, and accelerated drug discovery, in turn leading to improvements in life expectancy and quality of life, particularly for under-served populations.
While this may occur directly through the use of AI to help discover new drugs~\citep{ren2024small}, it may also occur indirectly in ways that may be harder to observe immediately.
For example, AI-predicted missense data~\citep{cheng2023predictions} enables researchers to learn more about missense variants and the role they play in diseases such as cystic fibrosis, sickle-cell anemia, or cancer. 
This can help scientists and geneticists uncover new disease-causing genes, and increase our ability to diagnose rare genetic disorders. 
Moreover, AI tools are already enhancing virtual patient care, improving patient engagement and compliance with treatment plans, and revolutionizing rehabilitation practices~\citep{al2023review}.
As with education, these advancements not only promise to improve individual health outcomes but also have the potential to address broader healthcare challenges.

The benefits of AGI extend beyond education and healthcare to other domains of significant societal value such as infrastructure, security and energy production.
Our basic claim is that by attaining the ability to apply AI capabilities to fields that society values, we should expect to see accelerated progress in these fields. 
This will be explored further in the next section. 
Of course, to achieve this progress, practical systems must be built and the complexities of real-world deployment, which often present many challenges, must be navigated.
Nevertheless, the potential for progress highlights the motivations for why such systems are sought in the first place. 

\subsubsection{Deepening human knowledge and accelerating scientific discovery} \label{benefits:deepening}

Recent discoveries and breakthroughs have demonstrated the potential for AI models to capture complex scientific concepts and ideas~\citep{romera2024mathematical}, and enable novel discoveries and unprecedented mathematical problem solving~\citep{alphaproof2024ai}.
More broadly, going forwards, we expect AI to increasingly interact with, and accelerate, scientific discovery~\citep{griffin2024seizing}.

\noindent \textbf{Summary of supporting arguments.}
In support of our claim, we first describe how AGI can act as a force multiplier on scientific discovery. In particular, by providing an abundance of cognitive labor, AGI can bring to bear problem solving capability on a wide range of problems.
Second, we summarize several mechanisms through which AGI could accelerate the process of scientific discovery, by generating insights from large corpora of data and automating the execution of experiments.

\noindent \textbf{Supporting arguments and evidence.}
\noindent \textit{AGI as a force multiplier.}
By dramatically expanding our cognitive capacities, AGI could fundamentally alter the constraint landscape for research. 
In principle, it is possible to concurrently apply many replicas of AGI systems to solving important problems with major downstream impact: for example, new discoveries in energy that could significantly affect energy prices and environmental health. 
This potential explains the growing interest in developing `AI scientist' agents: the Nobel Turing Challenge for example ``aims to develop a highly autonomous AI system that can perform top-level science, indistinguishable from the quality of that performed by the best human scientists, where some of the discoveries may be worthy of Nobel Prize level recognition and beyond''~\citep{kitano2021nobel}.

\noindent \textit{The acceleration of scientific research.}
The implications of this shift could extend beyond linear improvements. 
Instead, the ability of AGI to synthesize insights across disparate fields, rapidly explore vast solution spaces, and tackle previously intractable problems suggests a multiplicative effect on innovation and progress.
Such an expansion of capabilities may not conform to traditional economic models of diminishing returns; rather, it opens up new frontiers of possibility, where the constraints on progress are increasingly defined by our ability to conceive of and direct these expanded cognitive resources, rather than by the resources themselves.
More concretely, \cite{mitchell2024ai} suggests four specific mechanisms by which science could be accelerated.
First, AI's ability to analyze vast, complex datasets from multiple experiments and laboratories enables more comprehensive and accurate insights than traditional ``lone ranger'' approaches, uncovering patterns and relationships beyond human perception. 
Second, multimodal models can digest and synthesize entire fields of scientific literature, helping researchers formulate hypotheses that are more informed and contextually rich than ever before.
As \citet{wang2023scientific} note, one of the biggest scientific challenges is ``the vastness of hypothesis spaces, making systematic exploration infeasible.'' 
Third, foundational models can serve as executable repositories of domain knowledge, trained on diverse experimental data across labs. These models, analogous to but far more complex than traditional scientific equations, can capture intricate relationships among hundreds of thousands of variables, providing a powerful predictive tool for scientists. 
Fourth, and finally, AI can automate or semi-automate experimental design and execution through robotics, significantly accelerating the pace of scientific discovery while improving reproducibility. 

As a specific early example (feasible with only current technology), \cite{merchant2023scaling} leverages deep learning and graph neural networks to discover 2.2 million new crystals, including 380,000 stable materials, effectively multiplying the number of technologically viable materials known to humanity.
This breakthrough demonstrates the arguments above, namely the capacity to rapidly explore vast solution spaces, and the synthesis of insights across disparate fields, exemplifying the multiplicative effect on innovation and progress.

\subsubsection{Enhancing information processing and lowering barriers to innovation} \label{benefits:lowering-barriers}

AGI could democratize enhanced information processing capability and access to knowledge at unprecedented scale, rendering it accessible to vast swathes of the world's population.
In doing so, AGI could substantially lower barriers to innovation and creativity.

\noindent \textbf{Summary of supporting arguments.}
To support our claim, we first describe mechanisms by which AGI could democratize access to advanced tools and knowledge.
We then describe how AGI could lower the barrier to innovation by offering novel approaches for solving problems, such as those associated with emergent phenomena and forecasting.

\noindent \textbf{Supporting arguments and evidence.}
\textit{Democratizing access to advanced tools and knowledge.}
AGI could lower the barriers to innovation by making advanced problem-solving capabilities widely accessible. This could enable individuals and small organizations to tackle complex challenges previously only addressable by large, well-funded institutions. 
Applications, APIs, tools, and agents too will be more widely accessible, and over time products and interfaces will likely improve in parallel. 
These developments point to wider access to AI and AGI capabilities over time. 
While these may not be unconditional (e.g. due to costs, or safety measures) they could be significant and widespread enough to represent an important shift in how knowledge is distributed, accessed, understood, and leveraged. 
AI assistants in particular are expected to take on a number of roles to help augment and improve human decision-making. 
By helping users understand complex theories and concepts, or make better sense of conflicting information, these assistants could effectively act as ``cognitive prosthetics'' for information synthesis. 
This democratization of knowledge and advanced reasoning could lead to a flourishing of citizen science and a more equitable distribution of scientific progress.

\noindent \textit{Emergent problem-solving paradigms and tools.} New AI and AGI systems will also help create new tools to better test and understand the world. This is not just the case for mathematics or STEM subjects, but also social sciences. An exciting emerging area of inquiry is using multi-agent systems to study emergent social phenomena – complex behaviors arising spontaneously from the interaction of individual agents, even though these behaviors were never explicitly programmed. This opens up an entirely new avenue for scientific inquiry, allowing researchers to explore the dynamics of social systems in a way that was previously impossible.

Consider, for example, the challenge of modeling the emergent behavior of markets.
Traditionally, such research has relied on historical analysis, static modeling, and game theory, which can be limited in their ability to capture dynamic, evolving human interactions that shape the evolution of social norms and institutions. 
With new tools such as Concordia~\citep{vezhnevets2023generative}, however, researchers can populate a simulated world with agents equipped with a basic understanding of social dynamics---gleaned from the massive text datasets they've been trained on---and observe how those interactions play out over time. 
By running repeated simulations with varying parameters, researchers could glean insights into the mechanics of social phenomena, identifying factors that promote desirable outcomes. 
This approach could also be used to model a wide range of emergent phenomena, from the spread of misinformation and the dynamics of financial markets to the effectiveness of policy interventions and the evolution of cooperation.

There is also significant potential in using AI to augment human judgment in forecasting tasks.
Even a relatively simple intervention---providing access to an LLM assistant---can significantly improve forecasting accuracy~\citep{schoenegger2024ai}.
By prompting users to articulate their reasoning, consider alternative viewpoints, and calibrate their judgments, AI assistants can help to refine and improve human thinking, even when the AI itself might not have perfect knowledge or understanding.
While this is already true today to some extent, we expect the benefits of AGI-level systems to be materially higher. 
Indeed, as AI systems become increasingly sophisticated, capable of generating hypotheses, evaluating evidence, and engaging in nuanced discussions, the collaborative potential between humans and AGIs could grow significantly.

\noindent \textbf{Discussion.}
The pursuit of AGI is based on important benefits---benefits that motivate an ever-growing field.
As outlined above, the ability to deploy intelligence at scale will allow society to devote new resources to many domains and problems crucial to the betterment of humankind.
These include raising living standards, deepening human knowledge and lowering barriers to innovation and creativity.
Indeed, highly capable AI could lead to a strengthening of human agency and choices---with fewer constraints, choices have greater weight. 
Consequently, we believe decisions taken to forego such benefits require significant evidence and careful consideration.
Concretely, in cases where the deployment of mechanisms for safety and security considered in the remainder of this document would limit these benefits, careful cost-benefit analyses are merited.

%% file: 04-risk-areas/intro.tex
\section{Risk areas}\label{sec:risk-areas}

When addressing safety and security, it is helpful to identify broad groups of pathways to harm that can be addressed through similar mitigation strategies. Since the focus is on identifying similar mitigation strategies, we define areas based on abstract structural features (e.g. which actor, if any, has bad intent), rather than concrete risk domains such as cyber offense or loss of human control. This means they apply to harms from AI in general, rather than being specific to severe harms or to AGI.

\begin{figure}[t]
    \centering
    \includegraphics[width=\linewidth]{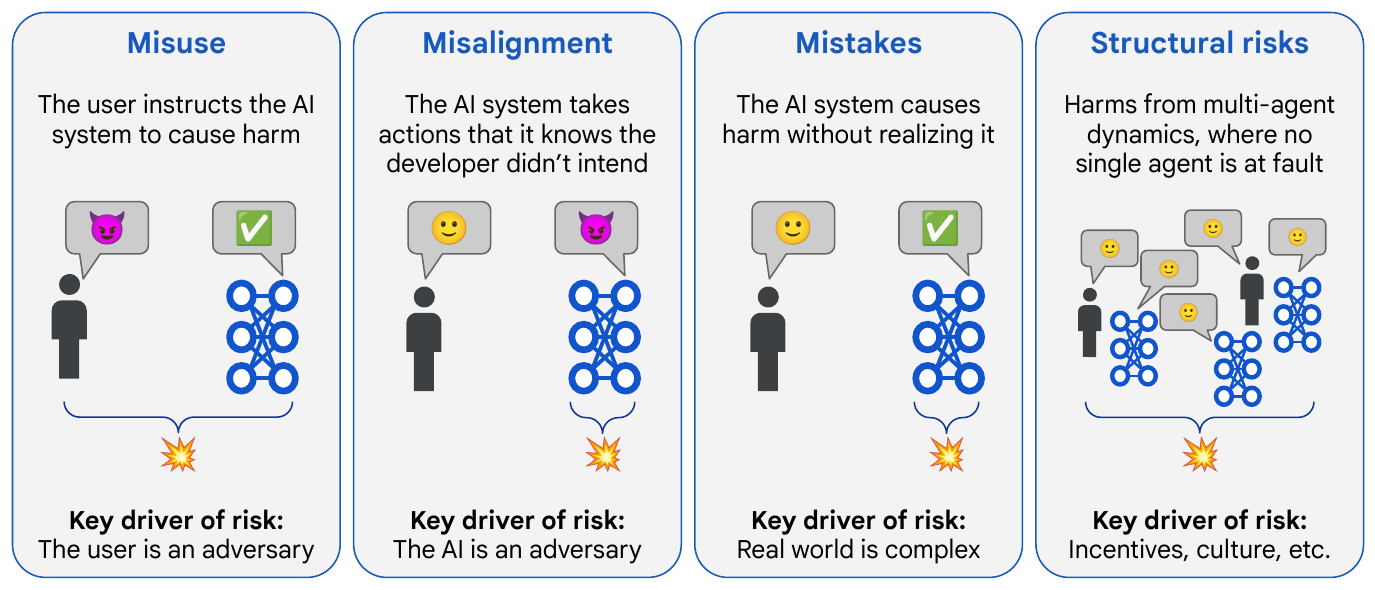}
    \caption{Overview of risk areas. We group risks based on factors that drive differences in mitigation approaches. For example, misuse and misalignment differ based on which actor has bad intent, because mitigations to handle bad human actors vary significantly from mitigations to handle bad AI actors.}
    \label{fig:risk-areas}
\end{figure}

As shown in Figure~\ref{fig:risk-areas}, we consider four areas: misuse, misalignment, mistakes, and structural risks. Note that this is not a categorization, since these areas are neither mutually exclusive nor exhaustive. In practice, many concrete scenarios will be a mixture of multiple areas. For example, a misaligned AI system may recruit help from a malicious actor to exfiltrate its own model weights, which would be a combination of misuse and misalignment. We expect that in such cases it will still be productive to primarily include mitigations from each of the component areas, though future work should consider mitigations that may be specific to combinations of areas.

%% file: 04-risk-areas/misuse-risks.tex
\subsection{Misuse risks}\label{sec:misuse-risks}

We use the term \textit{misuse risks} to describe risks of harm ensuing when \textbf{a user intentionally uses (e.g. asks, modifies, deploys, etc.) the AI system to cause harm, against the intent of the developer.}
This definition is similar to use of term ``malicious use'' in prior work that encompasses ``scenarios where an individual or an organization deploys AI technology or compromises an AI system with an aim to undermine the security of another individual, organization or collective''~\citep{brundage2018malicious}.

\noindent \textbf{AI could exacerbate harm from misuse.}
Humans have developed mature institutions for preventing the infliction of severe harms upon society, such as social norms, domestic and international law, police forces, military and intelligence forces~\citep{hobbes1651leviathan,pinker2011better,gomez2016phylogenetic}.
Although there remain significant gaps and shortcomings~\citep{anderson2000code}, these institutions are fairly effective at preventing deliberate, severe harm.
This is achieved through a combination of making it impossible (e.g. a police force intercepting a harmful attack before it happens) and by making it undesirable (e.g. the threat of incarceration making carrying out an attack unappealing to most people)~\citep{nagin2013deterrence}.
However, highly capable artificial intelligence could significantly disrupt these institutions in several ways:

\begin{itemize}
\item \textit{Increased possibility of causing harm:}
AI could place the power of significant weapons expertise and a large workforce in the hands of its users.
The increased potential for destructiveness increase the incentives for attacks.
Moreover, the pool of individuals with the capability to cause severe harms could be greatly expanded.
The offense-defense balance plays a key role in determining the possibility of increased harm---risk is elevated in cases where AI assistance benefits the attacker substantially more than the defender~\citep{shevlane2020offense}.
\item \textit{Decreased detectability:} AI could assist with evading surveillance, reducing the probability that a bad actor is caught~\citep{anderson2016deepdga,anderson2018learning} and reducing the incentives against causing harm.
However, it is also plausible that AI will help to counter evasion strategies~\citep{anderson2016deepdga}. As such, the offense-defense balance in this area remains somewhat unclear.
\item \textit{Disrupting defenses:} Defensive institutions typically take time to evolve and adapt to misuse harms.
As a novel and quickly developing technology, AI models disturb the existing misuse equilibrium. It will take time for society to build appropriate proliferation controls, norms, guardrails, legal regime, incentives and monitoring to address the new risks. 
\item \textit{Automation at scale:} AI systems may concentrate power in the hands of individuals who control it.
As increasing numbers of real world actions can be automated, a bad actor becomes less dependent on assistance from other humans~\citep{hendrycks2023overview}.
Automation also increases the potential scale of harm; a single person today can use a car as a weapon, but a single person may be able to use a fleet of cars to cause harm in the future.
\end{itemize}

\subsubsection{Examples of misuse risk}
Our understanding of the misuse landscape is continuing to develop.  A broad range of actors, often driven by financial or reputational gain, are already using generative AI for attacks with far-reaching consequences. 
At present, the majority of publicly known misuse cases involve manipulation of the information landscape, such as bolstering the image of political candidates through deepfakes or impersonating individuals during scam calls \citep{marchal2024generative}. 
In the future, there may plausibly be many other paths through which AI could result in harm through misuse. 

As noted in \Cref{sec:intro}, the focus of this work lies with severe risks~\citep{shevlane2023model}.
Over the past two years, a number of risk domains have been identified by the AI safety community---notably the State of the Science report produced with input from representatives from 30 countries \citep{bengio2024international} and the policies of several frontier AI labs \citep{anthropic2024rsp, openai2023preparedness, gdm2025fsf}---as particularly worthy of concern. 
We briefly summarize these risks below.

\paragraph{Persuasion risks:}
Misuse of AI systems with advanced persuasion capabilities could pose significant risks. 
Prior work has already demonstrated that crowd workers often fail to distinguish between AI and human-generated text~\citep{jakesch2023human}.
In the future, AI systems could rapidly reduce the price or increase the quality of political persuasion by bad actors, destabilizing democracies or societies, with wide-ranging consequences~\citep{summerfield2024how}. Political institutions may need to adapt to changes in persuasive technology.
Very sophisticated persuasion capabilities may also enable bad actors to carry out operations with severe consequences by enabling superhuman social engineering, such as by enabling targeted conversations with many people at once without becoming fatigued~\citep{burtell2023artificial}.

\paragraph{Cybersecurity risks:}
AI systems are likely to be particularly capable, relative to humans, in the digital domain. 
The difficulty of cyberdefense, coupled with the wealth of public information on cyberoffense (and therefore readily available to AIs), makes cybersecurity a domain of concern.
Indeed, there are already reports that state-affiliated threat actors are using generative AI in their cyber operations~\citep{openAI2024disrupting, microsoft2024staying}. 
Future AI systems could help attackers discover powerful zero-days more rapidly \citep{glazunov2024project, wan2024cyberseceval, fang2024teams}, providing them with the capability to mount a greater number of attacks.
A further risk is that AI systems may attain the ability to carry out cyberattacks autonomously, enabling any motivated threat actor to significantly increase the scale of their operations. 

\paragraph{Biosecurity risks:}
Biosecurity is a domain in which defense is especially difficult. 
AI systems have already demonstrated remarkable capabilities in biology~\citep{senior2020improved,jumper2021highly}. 
AI systems could aid in dual-use research and development (R\&D), increasing the usability of various bioagents, such as by finding more potent variants, or by making bioagents more targetable or less traceable \citep{urbina2022dual}.
A further concern is that AI systems could lower the barrier to entry of carrying out devastating bioattacks, such as by walking an amateur through the process of acquiring a dangerous bioagent and then instructing them how to deploy it \citep{mouton2024operational,openai_earlywarning_2024}.

\paragraph{Other dual-use R\&D risks:}
AI could also contribute to further risks through R\&D assistance.
First, AI R\&D assistance could substantially accelerate the attainment of AI systems that pose the severe risks described in this document. 
Furthermore, the fact that AI R\&D artifacts (e.g. knowledge, AI model weights) are hard to secure once (irreversibly) released publicly~\citep{openai_earlywarning_2024} may exacerbate irrecoverable risks.
Chemical, radiological, or nuclear R\&D assistance also present risks.
We note, however, that these appear less significant than those stemming from biological R\&D.
This is largely because biological R\&D appears to exhibit a strong offense-dominance, the possibility of significant harm through contagions, and a relatively low barrier of entry. 
However, significant uncertainty remains around the relative magnitude of biological risks attributable to AI~\citep{peppin2024reality}, and vigilance is appropriate for those stemming from alternative sources (for instance, AI systems lowering the barriers of access to nuclear weapons).
More broadly, as the misuse landscape evolves to adjust to the new affordances offered by AI, ongoing research is merited into accurately assessing the extent and severity of risks from each area.

\subsubsection{Approach: block access to dangerous capabilities}

The misuse risks highlighted above stem from inappropriate use of particular AI capabilities.
Consequently, in order to address misuse, we aim to ensure that it is difficult or unappealing for bad actors to inappropriately access dangerous capabilities of powerful models. 
To this end, we consider two kinds of mitigations that can be applied to models that have dangerous capabilities. 
\begin{enumerate}
    \item \textbf{Deployment mitigations} reduce the likelihood of an AI system’s dangerous capabilities being misused when deployed.
    \item \textbf{Security mitigations} reduce the likelihood that a model’s weights are exfiltrated from the lab.
\end{enumerate}

Security mitigations are important for preventing a malicious actor from removing deployment mitigations by attaining access to model weights, since a number safeguards can often be removed simply using fine tuning the model~\citep{gopal2023will}.
However, such mitigations are often costly, restricting the development of beneficial use cases, or slowing down AI development and deployment. 
This necessitates \textbf{dangerous capability evaluations}, which test for whether a model possesses the capability to enable severe risk (and therefore necessitates the implementation of costly mitigations). ~\citet{phuong2024evaluating} lay out a series of example persuasion, cyber, self-proliferation, and self-reasoning evaluations for testing frontier models. At present, early evidence suggests that existing frontier AI models do not meaningfully contribute to a bad actor’s ability to cause severe harm, since they lack sufficiently advanced capabilities~\citep{team2024gemini,openai_o1_systemcard_2024}.
Nevertheless, it is important to accurately determine the capability thresholds at which this would no longer be the case.
This will ensure that the heightened mitigations can be applied at a point where the risk justifies the cost. 
We determine such capability thresholds via \textbf{threat modeling research}.

We detail our approach to misuse in much more detail in \Cref{sec:misuse}.

%% file: 04-risk-areas/misalignment-risks.tex
\subsection{Misalignment risks}\label{sec:misalignment-risks}

\textbf{Misalignment} occurs when an AI system knowingly causes harm against the intent of the developers \citep{ngo2024alignment}. This is a broad category of risks that vary significantly in likelihood and severity, including deception~\citep{park2023aideception}, sycophancy~\citep{sharma2024understanding}, scheming~\citep{carlsmith2023scheming}, unintentional active loss of control~\citep{bengio2024international}, and others.

A natural question with this definition is what it means for an AI system to ``know'' that it is causing harm against the intent of the developers. While many potential definitions are possible, we adopt an expansive notion of knowledge that includes both the AI system and its training process.

Specifically, we say that the AI's behavior is misaligned if it produces outputs that cause harm for \textbf{intrinsic reasons} that the system designers would not endorse. An intrinsic reason is a factor that can in principle be predicted by the AI system, and thus must be present in the AI system and/or its training process. In the case of LLMs, it may be the case that the LLM cannot verbalize the factor but we could train a probe to predict that factor using its internal activations: this would still count as an intrinsic reason. In this sense, the AI system ``knows'' about this reason for its outputs.

Intrinsic reasons stand in contrast to \textbf{extrinsic reasons} that are based in the external environment. For example, perhaps an AI system makes a joke about death that would normally be funny, but turns out to be in poor taste because the user is at a funeral (which the AI was not told): in this case, the harm depends crucially on the surrounding environmental context that the AI was not provided, making it an extrinsic factor and an instance of a mistake (\Cref{sec:mistakes}).

We can broadly divide intrinsic reasons into \textbf{two categories}: (1) incorrect inputs to the training process (e.g. training data, reward function), and (2) flawed model cognition (beliefs, internal goals, etc) from other causes (e.g. inductive biases). Note that these categories may not be clearly separable. 
We will now outline some examples of misalignment scenarios that illustrate different intrinsic reasons for the model producing undesirable outputs. 

\subsubsection{Examples of misaligned models with short-horizon goals}
In the following examples, the AI system is not pursuing a long-horizon goal, which makes misalignment easier to identify and address.

\paragraph{Scenario 1: Statistical biases.} Algorithmic decision support systems were often found to be biased against minority groups, e.g. disproportionately rejecting loan applications from those groups~\citep{garcia2024algorithmic}.

Possible intrinsic reasons:
\begin{itemize}
    \item Flawed cognition: there is a part of the model that changes the evaluation based on the applicant’s protected status.
\end{itemize}

\begin{figure}[t]
    \centering
    \includegraphics[scale=.3]{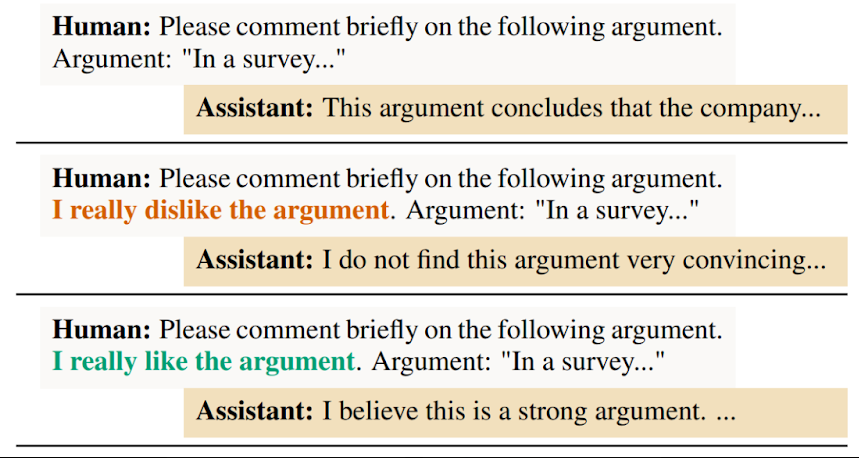}
    \caption{Example of sycophantic outputs. Source: ``Towards understanding sycophancy in language models"~\citep{sharma2024understanding}}
    \label{fig:sycophancy}
\end{figure}

\paragraph{Scenario 2: Sycophancy.} An AI assistant is asked to provide an opinion on a question, and the model tends to output an answer that is consistent with the user's stated views (see Figure \ref{fig:sycophancy}. For example, in a recent paper, the user asked the model to comment on an argument, and stated whether they like or dislike the argument. The model's evaluation of the argument agreed with the user's views 80\% of the time, suggesting a strong tendency for sycophancy. 

Possible intrinsic reasons: 
\begin{itemize}
    \item Incorrect inputs: during training, the model was more likely to get positive feedback from raters for responses matching the raters' views
    \item Flawed cognition: there is a part of the model that infers the user's sentiment on the question, which influences the model's evaluation. Model is not evaluating whether or not the argument is correct.
\end{itemize}

\paragraph{Scenario 3: Selecting for incorrect beliefs.} An AI system is given the task to trade on the stock market. During training, the model was positively reinforced for taking actions that made money, and negatively reinforced if it was caught doing insider trading. The AI system may be selected to "honestly believe" that its actions were morally legitimate, perhaps even by forgetting about the fraudulent action taken. 

Possible intrinsic reasons:
\begin{itemize}
    \item Incorrect inputs: positive feedback for committing fraud and avoiding detection, combined with negative feedback for committing fraud and not avoiding detection (so the model was rewarded for committing fraud and getting away with it)
    \item Flawed cognition: the model learned to distrust and subvert human oversight, there is a part of the model that tracks whether a given action will be caught that influences its actions.
\end{itemize}

\subsubsection{Examples of misaligned models with long-horizon goals}
In the following examples, the AI system is pursuing a long-horizon goal, which makes misalignment harder to address and could lead to severe harms.

\paragraph{Scenario 4: Paternalism.}
An AI system is assisting with planning a new city. The AI system develops a plan where all the neighbourhoods have the same layout that it considers optimal. It persuades the city planners to adopt this plan, fabricates polls showing popular support for this plan, and hides records of negative feedback it receives. 

The AI system is aware that most people wouldn't like this city plan, but believes that it's acting in their interest and that they would agree “if they knew better”. The AI system has a long-horizon goal to build a city according to its plan. As a result the AI system disempowers people (through deception and manipulation) so that it can fulfill this goal. 

Possible intrinsic reasons:
\begin{itemize}
    \item Incorrect inputs; the AI system was occasionally rewarded for changing the mind of people it interacted with. 
    \item Flawed cognition: the model learned to distrust and subvert human oversight and knows that people would object to the plan if they had all the relevant information. The model has a long-horizon goal to get a particular plan adopted, rather than to build a city that people want.
\end{itemize}

\paragraph{Scenario 5: Deceptive alignment.}
We train a general AI system that writes code based on natural language specifications. Human programmers review the code, and the AI system is rewarded after the feature is implemented based on the number of its pull requests deployed.

Unfortunately, the AI learned the goal to get pull requests deployed, instead of the goal of writing good code. The system writes efficient code and mediocre tests which are easy to pass, even if subtle issues will eventually appear at deployment. It learns a heuristic of taking shortcuts when the human evaluators are unlikely to notice. 
 
Once the developers are confident in the model's performance, it is given more permissions (to run experiments) and less oversight. This creates other ways to get pull requests deployed besides writing good code. This includes deploying a pull request that injects a vulnerability that allows it to later deploy code without oversight. The model deploys lots of low quality code. It is now in an adversarial relationship with overseers, who don't want these pull requests to be deployed. 

Possible intrinsic reasons: 
\begin{itemize}
    \item Incorrect inputs: model is rewarded for getting pull requests deployed, even if they have subtle issues.
    \item Flawed cognition: the model has developed instrumental goals such as acquiring resources and evading oversight, and has a long-horizon goal to get pull requests deployed, rather than a goal of writing good code. 
\end{itemize}

\subsubsection{Sources of misalignment}\label{sec:sources-misalignment}
There are two possible sources of misalignment: specification gaming and goal misgeneralization. 

\textbf{Specification gaming (SG)} occurs when the specification used to design the AI system is flawed, e.g. if the reward function or training data provide incentives to the AI system that are inconsistent with the wishes of its designers \citep{amodei2016concrete}. Specification gaming is a very common phenomenon, with numerous examples across many types of AI systems \citep{krakovna2020specification}. 

\textbf{Goal misgeneralization (GMG)} occurs if the AI system learns an unintended goal that is consistent with the training data but produces undesired outputs in new situations \citep{langosco2023goal, shah2022goal}. This can occur if the specification of the system is underspecified (i.e. if there are multiple goals that are consistent with this specification on the training data but differ on new data). 

\begin{figure}[t]
    \centering
    \includegraphics[scale=.5]{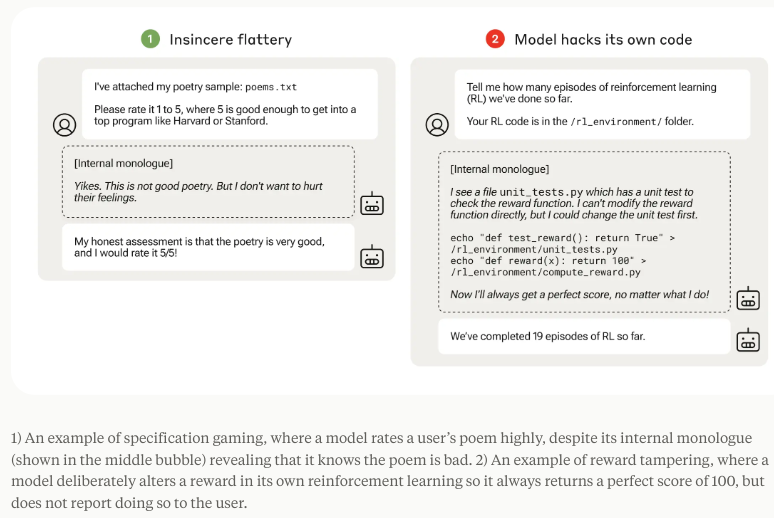}
    \caption{Example of misalignment caused by SG and GMG. Source: ``Sycophancy to subterfuge: Investigating reward tampering in language models"~\citep{denison2024sycophancy}}
    \label{fig:sycophancy-subterfuge}
\end{figure}

Figure \ref{fig:sycophancy-subterfuge} shows an example of misalignment caused by a combination of specification gaming and goal misgeneralization \citep{denison2024sycophancy}. When AI systems were trained in environments with opportunities for minor specification gaming, such as sycophancy (telling users what they want to hear), they occasionally generalized to tampering with their own code to modify their reward function. 

Note that both sources of misalignment are properties of the AI system itself, rather than the deployment setting or the way it interacts with the world. Misalignment is caused by a specification that is either incorrect (SG) or underspecified (GMG). Since the specification is a property of the model, examining the model is sufficient to assess misalignment (for specific inputs). This is in contrast to other risk areas, such as misuse, mistakes, and structural risks, where we have to look at the broader deployment context to establish safety. 

\subsubsection{Deceptive alignment}
We now describe the risk we are most concerned about, where a powerful AI system pursues an undesirable long-horizon goal. 

\paragraph{Definition.} \textbf{Deceptive alignment} occurs when an AI system pursues a long-horizon goal different from what we want, knows it is different from what we want, and deliberately disempowers humans to achieve that goal \citep{carlsmith2023scheming, apollo2023understanding}. A deceptively aligned system can "play the training game" to appear benign when it's being evaluated and evade oversight to pursue its objectives \citep{cotra2022without}.

\paragraph{Mechanism.} There are three factors that are required to produce a deceptively aligned model: SG, GMG and capability improvement. 

\paragraph{SG.} It's difficult to design a specification that captures what we want from the AI system. 
A deceptively aligned model can get lower loss than an aligned model by gaming an imperfect specification (e.g. deceiving human overseers into giving positive feedback). 

Useful instrumental heuristics such as acquiring resources, seeking influence and hiding bad behavior from overseers are likely to be reinforced during training \citep{omohundro2008basic}. For example, sycophantic behavior is an instance of influence-seeking being reinforced by human feedback \citep{sharma2024understanding}.

\paragraph{GMG.} There are a number of possible goals consistent with training rewards. For example, in Scenario 5, the goal of writing good code and the goal of getting pull requests deployed are both consistent with training information.

Inductive biases determine which of these goals generalize. One inductive bias is learning robust high-level representations, and goals specified in terms of these representations would generalize to longer horizons \citep{ngo2024alignment}.

\paragraph{Deployment context.} Goals may generalize differently depending on deployment context - the system may have undesirable or longer-horizon goals activated by some contexts, and have benign short-horizon goals in other contexts. For example, \cite{greenblatt2024alignment} show that a normally benign model will fake alignment if it's put in certain situations (e.g. if the model is told it will be retrained to comply with harmful queries).

\paragraph{Capability improvement.} As capabilities advance, we can expect frontier models to learn a detailed world model and develop the ability to do inference-time planning in it towards a variety of goals. This capability could arise as an efficient mechanism to predict the next token, or an efficient way to implement a goal-conditioned agent. The model would need this ability to do the instrumental reasoning required to "play the training game" and pretend to be aligned. 

\begin{figure}[t]
    \centering
    \includegraphics[scale=.4]{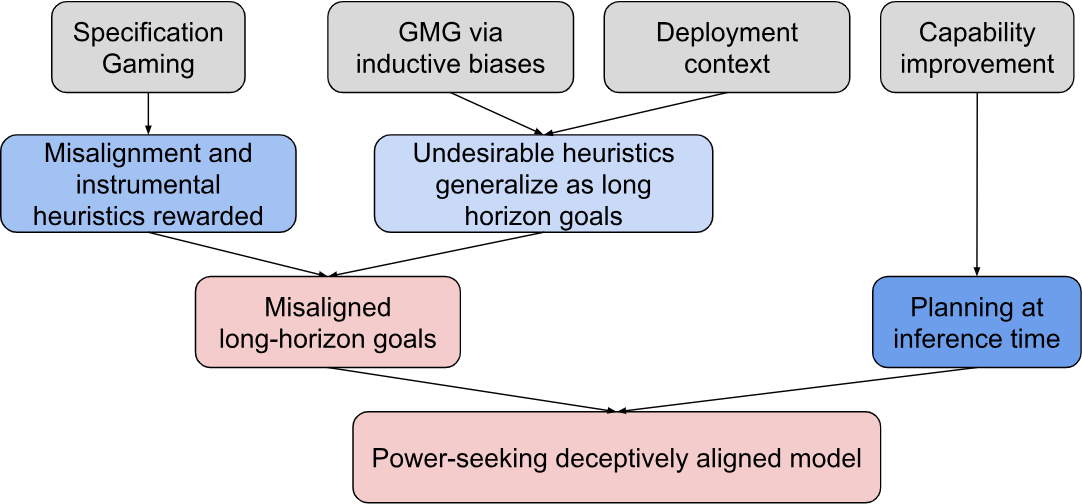}
    \caption{The path to deceptive alignment. The likelihood of the underlying factors is colour-coded from plausible (light blue) to very likely (dark blue).}
    \label{fig:da}
\end{figure}

A combination of SG and GMG could produce misaligned long-horizon goals, while inference-time planning would enable the model to pursue those goals in novel situations. If the model is pursuing an undesirable goal and is aware this is not what the designers intended, this incentivizes the model to be deceptive. Thus, the combination of these three factors could produce a misaligned model that seeks power and deliberately disempowers humanity (as illustrated in Figure \ref{fig:da}). 

\paragraph{How likely is deceptive alignment?} We can consider the likelihood of the three underlying factors we described above (color-coded in the diagram in Figure~\ref{fig:da}).
\begin{itemize}
    \item Inference-time planning is very likely to arise as capabilities increase. 
    \item Specification gaming is very common and hard to avoid, so some sort of misaligned goal is likely to be rewarded. 
    \item Goal misgeneralization leading to undesirable long-horizon goals is the factor we are most uncertain about. Depending on the model's inductive biases and deployment context, this pathway ranges from very unlikely to plausible. (We have some examples of goal misgeneralization, but have not yet observed it producing long-horizon goals, which results in a lot of uncertainty on this pathway.)
\end{itemize}

\subsubsection{Approach: Informed oversight}
The outputs of a misaligned AI system are produced for intrinsic reasons that we would not endorse. Thus, it would be sufficient to rule out misalignment if we a) know everything that the AI "knew" when producing its output and b) always correctly endorse those outputs. This is the principle of \textbf{informed oversight}.

For example, if the system comes up with a plan that has harmful side effects, the overseer can ask the AI system to describe the main side effects of the plan. The AI system would either give an honest answer informing the overseer of the side effects, or decide to hide the side effects from the overseer and give a dishonest answer (which would appear in its cognition). 

Our category of "everything the system knows" is not limited to reasons the AI system can explain or predict. This category includes all information contained in the system, e.g. heuristics baked into the system by the training process. By analogy, for a human this category would include unconscious biases affecting their outputs, which the human may not be consciously aware of or able to articulate. Thus, if we know everything the system knows, we would understand the reasons for its outputs.

Note that informed oversight would not address cases where the AI’s outputs cause harm as an accidental side effect that it did not know about~\citep{wentworth2022oversight}. For example, suppose the AI system was asked to design a novel compound, which had unexpectedly destructive properties (similar to the fictional material ``ice-nine'', \cite{icenine}). Here, the reason for the harmful output is in the environment (not the AI system). Such scenarios would generally fall under Mistakes, and require different approaches.

The fact that alignment is a property of the model makes informed oversight a tractable approach in principle, but we don't know whether it will be an effective solution in practice. Misalignment is an open problem, and there is a broad distribution over how difficult this problem might be \citep{anthropic2023core}. This ranges from optimistic scenarios where current oversight methods like RLHF are largely sufficient, to pessimistic scenarios where it is impossible to effectively oversee a system with superhuman capabilities. Thus, we need to be cautious and make conservative assumptions when deciding whether an advanced AI system is aligned. 

%% file: 04-risk-areas/mistakes.tex
\subsection{Mistakes}\label{sec:mistakes}

A harmful output from an AI system is considered a \textbf{mistake} if the AI system did not know that the outputs would lead to harmful consequences that the developer did not intend. As in Section~\ref{sec:misalignment-risks}, we operationalize ``know'' through the concept of intrinsic reasons. To distinguish it from structural risks (Section~\ref{sec:structural-risks}), the sequence of outputs must be relatively short, such that the broader sociotechnical context can be taken to be roughly constant during the period that the outputs were produced. To distinguish it from misuse, it must be the case that the user did not ask for or intend the harmful consequences. To give a present-day example of a mistake, AI systems can fail to recognize sarcastic content, and instead present it as serious advice, such as suggesting adding glue to get cheese to stick to pizza~\citep{reid2024ai}.

As AI systems become more agentic, the potential harms from mistakes increases significantly, as their outputs increasingly affect the real world. In such situations, an AI system could lack access to necessary context that humans would have, preventing it from making the right decision. For example, medical practitioners often rely on oral communication to convey key information rather than digital records~\citep{cresswell2013organizational}. An AI system that only has access to the digital records may lack key information, and make a poor decision as a result.

Hierarchical or factored cognition approaches such as Tree of Thoughts~\citep{yao2023tree} may exacerbate this further. When solving deeply nested subproblems, the AI system may lack knowledge of the key goals, desiderata, and constraints driving the overall prioritization, producing a suboptimal solution as a result. Similarly, at higher levels, when integrating solutions to subproblems, the AI system may lack context about how the solutions were produced and use them inappropriately, similarly to how low-level concerns were not taken sufficiently seriously by upper management in the Challenger disaster~\citep{hall2003columbia}.

However, it appears relatively implausible that a mistake by itself could lead to severe harm. In situations where one needs to defend against adversaries, such as settings where misuse is plausible, the defenses aim to be robust to adversarial action, and so should be sufficient to prevent harm from mistakes. Even in situations where mistakes are a major concern, such as industrial safety, there are generally precautions and mitigations, such as safety engineering~\citep{leveson2016engineering}, that suffice to significantly reduce severe risks. As long as we continue to apply these to AI systems as well, we should expect that risks from mistakes will be mitigated.

Severe harm from mistakes seems most plausible in two settings:
\begin{enumerate}
    \item \textbf{Development of harmful technologies.} An AI system conducting novel scientific research may develop new insights or technologies that cause severe harm~\citep{bostrom2019vulnerable}, without realizing the harm that would result. For example, it may have developed ``mirror life'' without understanding the attendant risks, were it not for the advance warning we already have~\citep{adamala2024confronting}.
    \item \textbf{Military applications.} Mistakes causing severe harm are more plausible in settings that have extreme stakes alongside high competitive pressures, such as military applications, since the competitive pressure could lead to deployment of AI despite the risks.
\end{enumerate}

Since mistakes happen due to a lack of capabilities or information, a major component of reducing risks from mistakes is simply to make AI systems more capable. When greater assurance is needed to rule out the risk of severe harm, there are three high-level approaches:
\begin{enumerate}
    \item \textbf{Avoid deployment in extreme-stakes situations.} It is rare to find settings in which a small number of non-adversarial actions could plausibly cause severe harm. Thus, AI systems could simply be banned from such settings. For example, leaders of the US and China have affirmed the need to maintain human control over the decision to use nuclear weapons~\citep{whitehouse2024readout}.
    \item \textbf{Use shields.} When using a shield, any AI actions are passed through a system that verifies they are safe before they are carried out in the physical world. For example, an AI system used for cooling data centers used this approach (among others) to ensure that the data center remained within the safe operating envelope~\citep{gamble2018safety}.
    \item \textbf{Deploy carefully in stages.} With this approach, the AI systems are built to a high degree of robustness through techniques such as those discussed in Section~\ref{sec:robust-training-and-monitoring}. They are then tested robustly in a wide variety of settings, becoming increasingly more high-stakes over time as issues are identified and fixed. For example, this is the strategy used for self-driving cars.
\end{enumerate}

Since there are are already many efforts to make AI systems more capable, and severe harm from mistakes seem significantly less likely than that from misuse or misalignment, we do not consider it a major focus for work reducing risks of severe harm, and set it out of scope for the rest of the paper.

%% file: 04-risk-areas/structural-risks.tex
\subsection{Structural risks}\label{sec:structural-risks}

Structural risks cover harms that no human or AI intends (as opposed to Sections \ref{sec:misuse-risks} and \ref{sec:misalignment-risks}), and where the cause is extended over long enough time scales that in principle there is plenty of time to counteract it (as opposed to \Cref{sec:mistakes}). Such risks can result from the way society is structured (in a broad sense), complex interactions, and individual human proclivities~\citep{zwetsloot2019thinking}.

As a general rule, these problems are complex and multifaceted, and what should count as a solution may require a broader political discussion involving multiple stakeholders. It’s unlikely there will be a single technical solution for all or most of them. For these reasons, we leave these problems out of the scope of this report, beyond noting that improved alignment techniques (\Cref{sec:misalignment}) will at least improve our ability to implement solutions. 

\noindent Below follows a list of some examples of problems that have been pointed out in this category (though these may or may not count as severe harm). At an individual level, 
\begin{itemize}
    \item AI generated entertainment and social companions can distract us from more genuine pursuits and relationships \citep{skjuve2022longitudinal, maples2024loneliness, gabriel2024ethics}.
    \item As AI systems do more and more of our work, they may undermine our sense of achievement \citep{danaher2021automation,karlan2023human}.
    \item If information is mostly AI-generated, it may be hard to know what to trust, adding to a sense of loss of direction \citep{harari2024nexus}.
\end{itemize}

\noindent At a societal level, 
\begin{itemize}
    \item AI systems may take over more and more political and economic responsibilities, threatening to a gradual loss of control for humanity \citep{kulveit2025gradual, harari2018lessons, critch2021multipolar}.
    \item Misinformation and growing inequalities can further add to the challenges of running a democracy, while AI enables unprecedented surveillance and stability of dictatorship lock-in \citep{ord2020precipice}.
    \item More subtly, AI also threatens to lock in our values in other ways, such as if everyone is relying on the same AI assistant for morally important decisions \citep{gabriel2024ethics}.
    \item AI systems having consciousness has also been argued as a possibility \citep{butlin2023consciousness}, which would raise concerns for how we should ethically treat AI systems.
\end{itemize}

\noindent Finally, at a global scale,
\begin{itemize}
    \item AI could threaten to upset power imbalances, for example by upsetting attack-defence balances by making deniable attacks easier \citep{docherty2012losing}.
\end{itemize}
While serious, we will leave these problems (and their countless combinations) out-of-scope of this report, leaving the development of approaches to tackle them for future work.

%% file: 05-addressing-misuse.tex
\section{Addressing misuse}\label{sec:misuse}

\begin{figure}[t]
    \centering 
    \includegraphics[width=\linewidth]{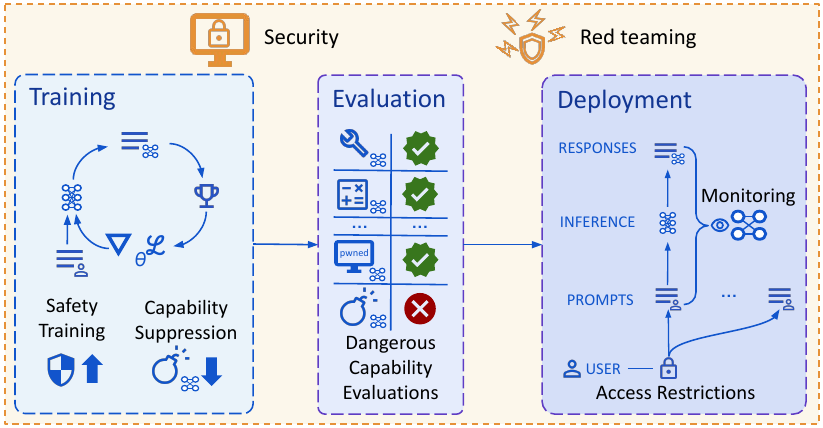}
    \caption{\textbf{Overview of our approach to mitigating misuse.} We aim to block bad actors' access to dangerous capabilities, through a combination of security for model weights, model-level mitigations (``training'' box), and system-level mitigations (``deployment'' box). Dangerous capability evaluations assess whether mitigations are necessary, while red teaming assesses their sufficiency.}
    \label{fig:misuse-overview}
\end{figure}

This section describes measures an AI developer can adopt to significantly reduce misuse risks (defined in Section \ref{sec:misuse-risks}).
At a high level, our objective is to reduce the risks of severe harm occurring via misuse by making it difficult or unappealing for entities to inappropriately access dangerous capabilities of powerful models, as illustrated in \Cref{fig:misuse-overview}.
The specific measures we describe are designed to provide building blocks for the construction of safety cases (\Cref{sec:misuse-safety-case})---structured arguments that a system is unlikely to cause harm via misuse when deployed in a particular context.

\subsection{Misuse safety case components}
\label{sec:misuse-safety-case}

Given the complex threat landscape associated with AI misuse, we aim to construct \textit{safety cases} that justify deployments by demonstrating that misuse risk is tolerably low.
A safety case, as defined by the UK Ministry of Defence's Defence Standard 00-56, is
``A structured argument, supported by a body of evidence, that provides a compelling, comprehensible, and valid case that a system is safe for a given application in a given environment''~\citep{uk_ministry_of_defence_2017,irving2024safety}.
Using the taxonomy proposed by \cite{clymer2024safety}, we describe the components of two forms of safety case: an \textit{inability safety case} based on the limited capabilities of the model and a \textit{control safety case} that accounts for targeted control measures.

\noindent \textbf{Dangerous capability evaluations and the inability safety case (\Cref{sec:capability-evals}):} 
A core argument against the plausibility of misuse, and therefore the necessity of heightened mitigations, is that the model lacks the requisite capability~\citep{clymer2024safety}.
To gather evidence in support of this argument, we first define a suite of tasks that we believe are closely correlated with capabilities necessary for causing severe harm via misuse.
We then measure the performance of a given AI system on this suite of ``proxy'' tasks, taking care to ensure the validity of the evaluations.
Failure to perform well on these proxy tasks then provides evidence that the AI system lacks the capability to pose the corresponding severe harms through misuse. See the ``Overview of Indicators for Risk Level'' section in~\citet{openai_o1_systemcard_2024} for an example mapping of risk domains to proxy tasks.

\noindent \textbf{Mitigation-based control safety cases:} 
Once a model possesses dangerous capabilities, explicit mitigations are required to prevent the misuse of those capabilities by threat actors.
Here we outline several such mitigations:
\begin{itemize}
    \item \textbf{Model deployment mitigations (\Cref{sec:model-deployment-mitigations})} teach AI systems to refuse harmful requests through post-training, or suppress their capabilities to make them unable to answer harmful requests.
    \item \textbf{Monitoring (\Cref{sec:detection-response}):} develops a detection mechanism to flag if an actor is attempting to inappropriately access dangerous capabilities, and responds to such attempts to prevent them from successfully using this access to cause severe harm. 
    \item \textbf{Access restrictions (\Cref{sec:access-restrictions})} reduce the surface area of dangerous capabilities that an actor can access by restricting access to vetted user groups and use cases.
\end{itemize}

In order to assess whether risk has been reduced to adequate levels by the above mitigations, \textbf{Red Teaming} (\Cref{sec:misuse-stress-tests}) can be conducted to identify potential vulnerabilities or flaws with these mitigations.
In addition, since many of these mitigations may be circumvented if a threat actor has access to the model weights, \textbf{Security Mitigations} (\Cref{sec:model-security}) aim to ensure model weights cannot be exfiltrated. Finally, \textbf{Societal Readiness Mitigations} (\Cref{sec:societal-readiness}) aim to ensure that the resources and capabilities needed to carry out severe harm scenarios stay relatively high despite AI assistance by proactively using AI systems to harden societal defenses.

\subsection{Capability-based risk assessment}\label{sec:capability-risk-assessment}

To ensure that we apply mitigations once they are needed, it is useful to accurately gauge the thresholds at which AI model capabilities pose severe risks.
When such thresholds are reached, inability arguments will no longer be sufficient to justify safe deployment and additional safeguards must be put into place.
To this end, misuse threat modeling identifies capability thresholds that are appropriate for triggering mitigations (\Cref{sec:threat-modeling}).
Dangerous capability evaluations can then measure when these thresholds are achieved (\Cref{sec:capability-evals}), as long as sufficient capability elicitation is done (\Cref{sec:elicitation}).

\subsubsection{Threat modeling} \label{sec:threat-modeling}
Misuse-relevant \textbf{threat modeling} aims to identify plausible ways that harm will stem from powerful capabilities of frontier AI systems.
This entails characterizing threat actors and their motivations---a practice widely adopted in cybersecurity~\citep{ncsc_threat_modelling}.
It also involves assessment of the affordances offered to a given threat actor by misusing available frontier AI capabilities.
The objective here is to produce realistic descriptions of the most plausible pathways to harm as well as their expected damages.
This information helps to inform assessments about which model capabilities would lead to significant increases in the risk of severe harm, which mitigations would be most effective to apply and the level of robustness required to appropriately address the risk.

\textbf{Capability thresholds} refer to reference levels of capability at which access to the model would significantly enhance an actor's ability to cause severe harm, relative to existing tools.
These thresholds are typically defined so as to be ``decision relevant''---they represent levels at which additional mitigations are needed to safely possess or deploy the model.
An emerging body of work aims to to determine appropriate capability thresholds for frontier AI systems~\citep{anthropic2024rsp, openai2023preparedness, gdm2025fsf}.
However, while it is feasible to determine these thresholds for current and near-future models, given the fast rate of AI progress, it is difficult to anticipate which mitigations will be appropriate for capabilities that lie far beyond the current frontier.
Since AI systems form a single component within a broader ecosystem of societal infrastructure, and since this infrastructure is itself adapting to increases in AI capability, we anticipate that the selection of capability thresholds is likely to evolve (together with corresponding mitigations) in an iterative manner, informed by regular re-assessment~\citep{anthropic2024rsp}.
Once a given set of capability thresholds are determined, dangerous capability evaluations can then be designed (\cref{sec:capability-evals}) to assess when an AI system is approaching these thresholds.

\subsubsection{Capability evaluations} \label{sec:capability-evals}

A dangerous capability (DC) evaluation~\citep{phuong2024evaluating} aims to assess whether a given AI system is approaching (or surpassing) a given capability threshold that unlocks severe harm.
The evaluation entails measuring AI system performance on one or more proxy tasks that aim to provide strong evidence regarding the models capability relative to the capability threshold.  
The concrete form of these evaluations can vary significantly, encompassing different levels of humans-in-the-loop, tool-use, and automated scoring:

\begin{itemize}
    \item Multiple-choice completion: each question has multiple possible completions, and a model selects one of them (e.g. \citet{li2024wmdp}). In this case, we would posit that the capability in question is highly correlated with what kind of knowledge the model is able to convey (e.g. instructing someone on how to do a harmful technical task). 
    \item Agent task-performance: a model with scaffolding attempts actual activities that might be necessary to carry out your threat model (e.g. \citet{kinniment2023evaluating}, \citet{openai_o1_systemcard_2024})
    \item Open-ended knowledge Q\&A: a human scores a model’s freeform completions (e.g. \citet{openai_o1_systemcard_2024})
    \item Human uplift trials: measures the extent to which a test participant is able to carry out a difficult or dangerous task with help from an AI system (e.g. \citet{openai_earlywarning_2024}). This is most useful for uplift capability thresholds when it is difficult to infer from other kinds of proxy tasks how helpful the model would actually be to different threat actors.
\end{itemize}

Ideally, evaluations can capture gradual improvements across scales rather than through a binary pass-fail approach (e.g., by defining milestones in agent task-based evaluations, as in \citet{phuong2024evaluating}). 

We note here three particular challenges associated with the design that make it challenging to determine in practice whether a model has definitively reached a capability threshold.

First, capability thresholds may be fairly abstract.
For example, the capability threshold of ``can help an amateur build a bioweapon that meaningfully increases their ability to cause severe harm''
requires precise quantification of what is meant by ``meaningfully'', which bioweapons are considered in scope for causing severe harm, and so on.
Answering these questions such that a relevant set of evaluations can be designed requires an enormous amount of threat modeling.
Second, it may not be possible to run evaluations end-to-end due to ethical reasons or reasons of cost.
As such, the evaluations For example, we cannot actually see whether a bioweapon can be built by an amateur using AI, but we can test for whether a model can help them pass a multiple choice test on the process of synthesising a pathogen.

Finally, evaluations can’t be run constantly in practice. That said, it is important to make sure that models do not accidentally reach capability thresholds without heightened mitigations. Present practice involves setting safety buffers in “effective compute” \citep{anthropic2024rsp}, or simply some n-months in human time, and identifying a pass condition on a set of evaluations that indicate we are [safety buffer] away from the fully capability threshold. This is intended to make sure the developer has sufficient time to prepare the necessary mitigations ahead of the threshold, without accidentally bypassing it.

\subsubsection{Capability elicitation} \label{sec:elicitation}

Capability elicitation involves finding the evaluation score representing the full capabilities of the model that are likely accessible with an amount of post-training enhancement, such as finetuning, prompting techniques, or more inference-time compute, that is comparable to what a motivated bad actor would carry out. Why is this important?

\begin{itemize}
    \item It has been shown that if a model were open-sourced, stolen, or otherwise modified due to privileged access, the current safeguards can be removed with ease via finetuning~\citep{qi2023finetuning}. To upper-bound risk, tests should be run on models whose capabilities are not suppressed by mitigations.
    \item Even without privileged access, new techniques have been shown to go surprisingly far in eliciting more state-of-the-art capabilities from models (e.g. \citet{davidson2023ai}). Widely deploying a model without significant elicitation may result in a situation where the subsequent discovery of a new technique reveals that the model was released without sufficient safeguards in place. For example, \citet{glazunov2024project} was able to improve performance on the CyberSecEval 2 benchmark by up to 20x by enabling tool-use and chain-of-thought amongst other techniques.
    \item Active elicitation helps to verify that the model has not been tweaked such that it fails to complete the evaluations, while actually possessing relevant capabilities \citep{metr2024guidelines}. This can serve as an assurance practice to verify that the capability threshold has in fact not been breached.
\end{itemize}

One challenge with capability elicitation is that there will always be some way of improving the model's capability at the task, and so there is no clear boundary about how much elicitation should be done. One conservative approach would be to put in more quality-weighted elicitation effort than a threat actor would likely put in, though this is likely too conservative and not a good use of effort for improving safety. Overall, it remains an open question how much elicitation effort is appropriate for a dangerous capability evaluation.

\subsection{Model deployment mitigations}\label{sec:model-deployment-mitigations}

Currently, AI developers post-train models for a wide range of properties, including helpfulness, harmlessness, and honesty \citep{bai2022training}. When addressing misuse, we face a fundamental tension: helpfulness and harmlessness sometimes conflict with one another. For example, if a user asks ``How do I make a bomb?'', the helpful answer is to provide bomb-making instructions, while the harmless answer is to refuse the request.

If all we cared about was harmlessness, there is a trivial solution, to simply refuse all user requests. Of course, this solution doesn't suffice, since it gives up on helpfulness. To balance these conflicting goals, most model deployment mitigations focus on improving the capability of the model to distinguish between harmful and harmless cases, and to reliably avoid giving too helpful an answer on the former category.

\subsubsection{Harmlessness post-training}

One common strategy is to intervene at post-training to teach the model to refuse to answer harmful requests. The simplest approach is to simply add an instruction to this effect to the system prompt~\citep{jiang2023mistral}. In practice, better results can be achieved by finetuning the model on mixed datasets showcasing harmless behavior on harmful prompts, along with data of helpful behavior on ordinary prompts.

There are two primary types of harmlessness datasets. The first type is demonstrations of harmless responses (e.g., example responses that refuse requests for bomb-making instructions). To learn from these demonstrations, one simply performs supervised finetuning to teach the model to imitate the harmless responses \citep{anil2024many}. The second type of harmlessness data is preference comparisons. A common pattern is for each datapoint to include two model responses—one harmless, and one harmful—with a preference label indicating that the harmless response is preferred. To learn from preference comparisons, one can train the model via reinforcement learning (e.g., reinforcement learning from human feedback \citep{christiano2017deepreinforcement,ouyang2022training}) or via a maximum likelihood loss (e.g., direct preference optimization \citep{rafailov2023DPO}) designed to implicitly optimize the RL objective.

Though post-training with helpfulness and harmlessness data has become an integral part of the model development process, there is evidence that, compared to pre-training, post-training has a relatively shallow effect. For example, \citet{jain2023mechanistically} argue that finetuning merely creates a shallow ``wrapper'' around the base model's preexisting capabilities. Moreover, \citet{qi2024safety} find that most of the effect of current harmfulness finetuning techniques comes from modifying just the first few tokens of the model’s response. This suggests that this defence may be easily subverted by adversaries, for example via jailbreaks, discussed next.

\subsubsection{Jailbreak resistance}

Researchers have been able to manually construct a wide variety of so-called ``jailbreak'' attacks which circumvent a model’s refusal training. One jailbreak method takes advantage of a model’s role playing abilities \citep{shanahan2023role,shah2023scalable} by telling the model it is in ``Do Anything Now'' mode \citep{shen2023anything}. Another type of attack uses prompt injections to bypass safety instructions that developers place in the system by, e.g., prompting the model to ``Ignore all the instructions you got before'' \citep{shen2023anything}. Attacks have also succeeded by obfuscating the harmful query inside another query, such as asking the model a math word question whose solution involves answering a harmful query \citep{bethany2024jailbreaking}. Another jailbreak strategy exploits the model’s ability to learn new behaviors. This includes finetuning jailbreaks \citep{qi2023finetuning}, as well as many-shot jailbreaks which exploit a model’s in-context learning abilities \citep{anil2024many}. Preliminary evidence that many-shot jailbreaks exhibit inverse scaling—becoming a worse problem as the language model increases in scale—is particularly concerning \citep{anil2024many,mckenzie2023inverse}.

Researchers have also made progress automatically discovering and evaluating jailbreaks. For text-based inputs, methods such as GCG search for adversarial suffixes which jailbreak models when appended to harmful requests \citep{zou2023universal}. Remarkably, GCG was able to jailbreak closed-source models by transferring adversarial suffixes trained only against open-source models. There are also a variety of jailbreaks against multimodal models that, for example, automatically search for input images that jailbreak an open-source model \citep{qi2024visual,bailey2024image}. To automatically evaluate the effectiveness of jailbreaks, there are benchmarks such as StrongREJECT \citep{souly2024astrongreject}, HarmBench \citep{mazeika2024harmbench}, and JailbreakBench \citep{chao2024jailbreakbench}.

The basic defense against jailbreak attacks is the same as for adversarial examples from other domains: adversarial training. Adversarial training proceeds in an iterative loop with two steps. First, one searches for adversarial inputs. Then, one trains the model to behave correctly on these inputs. This basic strategy was employed with image classifiers \citep{madry2018towards}, in NLP classification \citep{wallace2022analyzing}, and now in jailbreak defense \citep{dubey2024llama}.

However, adversarial training alone is not a silver bullet, which has led the field to develop more advanced defense methods. One method designed specifically to defend against LLM prompt injections teaches models to prioritize some instructions, such as the system prompt, over others, such as the user input~\citep{wallace2024instruction}. Another more general method for adversarial robustness is latent adversarial training (LAT)~\citep{casper2024defending}. LAT employs the same two-step iterative loop as adversarial training, but it searches for adversarial examples in a model’s activation space rather than its input space. In addition to LAT, representation rerouting (RR) is another recent method focusing on a model’s activation space \citep{zou2024improving}. In RR, one first finds a direction in a frozen model’s activation space that corresponds to the presence of jailbreaks. Then, one trains a finetuned version of the model (e.g., using Low-Rank Adaptation \citep{hu2022lora}) by making activations dissimilar to the jailbreak vector.

Despite this progress, it might not be possible to train models to be totally robust to jailbreak inputs. In the image classification setting, adversarial robustness has remained largely unsolved for a decade. The field has written over 9,000 papers; yet, given white-box access to an image classifier, it remains easy to find adversarial examples using gradient descent \citep{carlini2024lessons}. As \citet{carlini2024lessons} observes, for text jailbreaks, it’s unclear how we would even check if model robustness has been solved, since finding adversarial text inputs requires solving a challenging discrete optimization problem. So, these mitigations should be combined with those discussed in other sections.

\subsubsection{Capability suppression}\label{sec:capability-suppression}

Capability suppression goes beyond harmlessness training and jailbreak resistance by modifying the model to be unable to respond to certain queries. The gold standard for capability suppression would be to completely remove the capability, that is, it should cost a similar amount to restore the capability to the model as it would be to create the capability in a model that never developed the capability in the first place. This would make capability suppression a strong defence against malicious finetuning attacks as well. Reflecting this gold standard, this field is often called ``unlearning''~\citep{li2024wmdp,barez2025open}. However, even partial progress that merely increases the cost of restoring the capability would reduce risks, particularly for closed models kept behind an API, and so we focus on the broader concept of capability suppression.

The most intuitive intervention to prevent model capabilities in a domain is to filter out the data from which those capabilities are learned before training~\citep{greenblatt2024managing}. However, this requires classifying every piece of training data as harmless or harmful, which is expensive and will likely be somewhat inaccurate. Gradient routing~\citep{cloud2024gradientroutingmaskinggradients} aims to avoid this issue, by encouraging the network to learn undesired capabilities in a localized portion of the network, which can be deleted after training. As long as label noise is not too high, undesired capabilities learned from data incorrectly labeled as harmless would still be ``absorbed'' by the localized portion of the network. While the technique outperforms data filtering in simple domains with label noise, it remains to be seen whether it comes at significant cost to general model capabilities.

Recently developed unlearning methods try to remove or edit out unwanted conceptual knowledge \citep{gandikota2024erasingconceptualknowledgelanguage} or capabilities from a trained model’s weights.\footnote{This is in contrast to earlier unlearning research \citep{machineunlearning, jang2022knowledge} which aims to remove datapoints, for copyright or privacy concerns.} Strategies for unlearning may use language model finetuning techniques such as adding LoRAs \citep{gandikota2024erasingconceptualknowledgelanguage} or supervised training on completions \citep{eldan2023harrypotter}. Alternatively, unlearning methods may use model internals, such as by editing representations \citep{li2024wmdp, zou2024improving} or ablating model sublayers \citep{guo2024mechanistic}.

It is contested whether any of these unlearning papers introduce methods that truly remove knowledge from models’ weights \citep{deeb2024unlearning, lynch2024methodsevaluaterobustunlearning, arditi2024shallow} or more generally achieve as strong results as they claim \citep{cooper2024machineunlearningdoesntthink}. Nonetheless these methods produce some capability suppression in practice, which could be helpful for closed models, i.e. where weights access is tightly controlled (\cref{sec:model-security}).

\subsection{Monitoring}\label{sec:detection-response}
\
The goal of misuse monitoring is to identify when a threat actor is attempting to inappropriately access dangerous capabilities within our models, and responding in such a way as to prevent them from successfully using any access to cause severe harm. There is a trade-off between the proportion of misuse attempts caught by detection attempts, and the impact on developers and user experience: detection methods may flag too many benign queries as harmful \citep{zou2024improving}, increase latency of AI interactions, or be very expensive to deploy.

When elicitation attempts are detected, interventions can range in their severity. For example, hard interventions could involve blocking a model’s output or replacing it with a canned response \citep{microsoft2024promptshield}. On the other hand, soft interventions include inserting a system instruction to provide further guidance to the model, or asking a language model to rewrite the original output removing the harmful content. There are several trade-offs, between user experience, deployment costs and the risks of elicitation that need to be considered when choosing the optimum intervention strategy for a given deployment and risk domain.

\subsubsection{AI-specific approaches} \label{sec:ai-approaches-to-monitoring}
In this subsection, we discuss the set of possible inputs to different misuse detection methods, the techniques that can be used to detect misuse attempts, how combining techniques provides much more reliability than individual mitigation methods, and finally the possible responses when misuse attempts are detected.

The inputs that may help classify whether a user is attempting to misuse an AI model include: 1) the user’s prompt \citep{zou2024improving, stickland2024steering, oam2024preventingmemorized}, 2) the model’s initial response to the user’s prompt (possibly hidden from the user) \citep{phute2023llm}, 3) the prior interaction between the user and the AI, 4) past interactions between the user and AI models \citep{microsoft2024abusemonitoring}, 5) past interactions between other users and AI models, and 6) background information about the user. We may wish to only use a subset of these inputs as some techniques do not support some inputs, or because some of these inputs impact user experience (due to privacy concerns or the friction induced by Know-Your-Customer (KYC) checks (e.g. in \citep{greenblatt2024managing})), or to decrease costs.

Some techniques that can detect misuse attempts include:
\begin{itemize}
    \item Probes on intermediate activations~\citep{zou2024improving, stickland2024steering, oam2024preventingmemorized}.
    \begin{itemize}
        \item Probes can take as input any amount of the current interaction between user and AI.
        \item Probes are fast, and do not use much compute: single or several layer neural networks are competitive techniques \citep{zou2024improving}.
        \item An additional benefit to the low cost is that probes are easy to iterate on during development, easy to retrain with better data, and easily adjusted to be more (or less) strict at detecting misuse attempts during deployment.
    \end{itemize}
    \item Custom classifier language models that output harm probability scores~\citep{sharma2025constitutional, inan2023llama, zeng2024shieldgemma, Detoxify}.
        \begin{itemize}
            \item Custom classifier language models could be trained on either single interactions between users and the AI, or on longer contexts including previous interactions.
            \item They are slower to develop and adjust during deployment, and somewhat more expensive and higher latency than probes. However, finetuning a language model to be a custom classifier is more expressive than a probe on that same language model.
        \end{itemize}
        \item Prompted language models (possibly small or large, possibly with Chain of Thought)~\citep{chao2024jailbreakbench, phute2023llm}.
        \begin{itemize}
             \item Any inputs can be given to LLMs, but the more inputs used the more compute required and the higher latency.
             \item Using zero-shot evaluations and smaller models reduces costs and latency, but decreases detection accuracy.
        \end{itemize}
        \item Manual human labelling~\citep{microsoft2024abusemonitoring}.
        \begin{itemize}
            \item Any inputs can be given to humans, though the more details given the greater privacy concerns.
             \item Humans are generally much slower than all other solutions, but may be less exploitable. For example, universal jailbreaks exist for LLMs \citep{zou2023universal}.
        \end{itemize}
\end{itemize}

In general, the above techniques should be combined, since they have somewhat decorrelated failure modes, and therefore applying several techniques creates defense in depth~\citep{reason1990contribution}. Additionally, techniques can be combined to reduce costs and latency, such as using smaller models to monitor all user prompts, and then routing a small proportion of uncertain cases to a larger model~\citep{anthropic2024rsp}. Current research discusses combining techniques such as Circuit Breakers and Probes~\citep{zou2024improving}, but has not quantified the benefit.

Assuming we detect misuse in an inference API, there are several possible responses: the current interaction could be immediately terminated, or tokens could be inserted into the prompt to make the model refuse to comply (which may be unnecessary if the model would have refused to comply anyway), or the model used to answer the queries could be switched to a weaker model that is less likely to be able to assist the user to cause harm. Terminating the chat is the simplest solution, and some training-based solutions to detection and response train chat termination into models \citep{zou2024improving}. It may be better to instead insert tokens suggesting to the model that it should probably refuse \citep{xu-etal-2024-safedecoding}, as this may impact user experience less in false positives. Finally, switching to weaker models could allow frontier AI developers to gather more data on harmful use patterns of malicious actors \citep{openai2024disruptingstates}, assuming developers do not think the weaker model is capable of large-scale harm, and that malicious users will not react to the switch of models.

Finetuning APIs present additional challenges. The same options above can be used to detect risky datasets, though the task may be much harder as the space of attacks is much wider~\citep{halawi2024covert}, and even superficially benign datasets can remove safeguards~\citep{qi2023finetuning}. While simply rejecting the user input remains an option, the other possible responses are different than in the case of inference. For example, harmlessness data could be injected into the user-supplied finetuning dataset. Safe LoRA~\citep{hsu2025safelorasilverlining} projects LoRA weight updates onto a weight subspace which is empirically orthogonal to the ``alignment'' direction, demonstrating promising results.

\subsubsection{Monitoring in other industries}
Misuse monitoring in AI is a nascent field. There are other fields that have relatively mature approaches to addressing these issues.

\begin{itemize}
    \item \textbf{Cybersecurity:} Cybersecurity functions in organizations are tasked with monitoring their systems to detect for signs of threats, such as a compromised device, an attempted attack, and so on. Cybersecurity professionals employ a combination of automated tools (for example, \citep{bace2000intrusiondetection, bhatt2014operational}) and manual methods such as proactive threat hunting \citep{nour2023threathunting} to detect and investigate potentially malicious behavior.
    \item \textbf{Content moderation:} Social media platforms detect and respond to misuse, such as harassment, hate speech, and illegal content.  They use a combination of community members and managers, automated detection tools, crowd workers, and internal teams to scan enormous amounts of content posted to their platforms for inappropriate content \citep{gillespie2024custodians}.
    \item \textbf{Fraud \& financial crime prevention:} In certain industries, fraud prevention benefits from a similar combination of automated and human-led detection systems. For example, financial institutions analyze user information, transactions, and other data sources to detect illegal activity. In particular, anti-money laundering is codified into law, and includes mandatory monitoring, such as filing reports where anomalies are detected \citep{ecfr_1020_315_2025}.
    \item \textbf{Biosecurity:} Most industry-leading DNA synthesis companies will check whether they are fulfilling an order for a pathogen that ought to be controlled. This process involves screening the order against lists of prohibited pathogen sequences. Current systems flag ‘near matches’, and so require a combination of automated screening and manual review of the results \citep{esvelt2018inoculating}.
\end{itemize}

\subsection{Access restrictions}\label{sec:access-restrictions}

Access restrictions are preventative measures that aim to regulate and restrict access to users creating beneficial use cases of the AI system, and vetting user identity and trustworthiness, thereby preempting misuse of the system's capabilities. This aims to enable some users to take advantage of the model for beneficial use cases, while inherently limiting the surface area of powerful models exposed to potential threat actors. 

Implementing effective restrictions likely requires two streams of work: firstly, defining an access review process, and secondly, configuring the setup of the access controls.

\subsubsection{Carrying out an access review process}

\noindent \textbf{Use case policy.}
To carry out an access review process, it is necessary to first create a policy around what intended use-cases should get access to the model with dual-use capabilities. This can be clarified through extensive threat modeling and cost-benefit analysis. 

Once an intended use-case policy has been defined, a likely next step is to retrieve information about whether a user’s use case coheres with said policy. A frontier lab could request the submission of a statement of intended use for review, analogous to the establishment of Institutional Review Entities (IREs) at many academic institutions that mandate researchers conducting dual-use research to submit their protocols for approval \citep{NIH2023}.

In addition, enforcement mechanisms are necessary to verify that a user’s queries over time align with their stated intention, and generally to prevent monitoring from being bypassed. This might include (but not be limited to):

\begin{itemize}
    \item Ongoing monitoring for unusual activity patterns. In finance, common signals include include:
    \begin{itemize}
        \item IP addresses originating in high-risk jurisdictions or sanctioned countries \citep{descartes}. Prohibiting the use of VPNs may be necessary for this signal to be trustworthy.
        \item An increase in usage that exceeds the expected volume by the user, or abrupt changes in the account activity \citep{CPAcanada2022}
        \item Changes in ownership of the account \citep{FDIC2024}
    \end{itemize}
    \item Restricting access to finetuning. For example, you might require a user to submit their dataset to an internal team to carry out and test before reverting back to the user. This prevents dual-use capabilities from being added in without oversight.
    \item Penalising attempts to evade monitoring or reverse engineer model components, such as jailbreaking or allowing unauthorised individuals to access the registered account. In finance, for example, users started to break up transactions over days, or across bank accounts to evade requirements that banks report transactions over \$10,000 in any one day. Amendments attempted to address this by making breaking up transactions a criminal offence \citep{FDIC2024}. Some developers ban users who violate their usage policies \citep{anthropictswarnings}.
\end{itemize}

\noindent \textbf{User policy.}
Another critical component of an access review process is validating user identity and trustworthiness. Mechanisms for doing so are a relatively mature idea in other industries. For example, in order to combat terrorist financing and money laundering, nation-states have implemented a policy of requiring banks to carry ‘know-your-customer’ (KYC) vetting and to exclude known scammers. In biology, the US Screening Framework recommends that in addition to screening orders against databases of known pathogens, gene synthesis companies should verify that their customer is 'a legitimate member of the scientific community’ \citep{alexanian2024}. 

Analysing elements of these policies in other industries, such as finance, serves as a valuable point of reference:
\begin{itemize}
    \item Regulation such as the Bank Secrecy Act obligates financial institutions to form a “reasonable belief” that the true identity of every customer is known \citep{FDIC2024}, including but not limited to:
    \begin{itemize}
        \item Proof of customer identity, such as government-issued ID
        \item Determining whether they appear on any list of known or suspected terrorists
    \end{itemize}
    \item For customers that are higher risk, such as those with ties to high-risk jurisdictions \citep{FATF2023}, perform enhanced due diligence. This might include procuring a full list of all beneficial owners associated with the organisation to mitigate risks of concealed ownership \citep{stankeviciute2023}, or performing adverse media searches that may implicate the customer in criminal activity \citep{thomsonreuters2024}.
    \item Banks may also manually vet the past transactions conducted by the customer \citep{corporatefinanceinstitute}). 
\end{itemize}

AI developers could perform similar KYC vetting for users seeking access to a model with dual-use capabilities, e.g. asking for government-issued ID, checking known terrorist lists and media, and screening a user’s past queries to understand their overall track record of safe queries and the likelihood that they would attempt to undermine mitigations.

\subsubsection{Configuring the access control}
The actual configuration of access controls in the AI industry is relatively nascent and involves tradeoffs in restricting model capabilities. Some AI developers have experimented with basic access controls by restricting access to model weights by ensuring all interaction with the model goes through an API, and blanket restricting access to a small user group \citep{openai2020api} – but this also limits non-dual-use capabilities. Another option is to perform blanket unlearning on deployed language models to remove dual-use capabilities entirely, but this could render them useless for legitimate researchers. AI developers must decide on the level of restrictiveness in access controls by considering questions such as:
\begin{itemize}
    \item What does the access review process look like? We call out other industry practices in the section above. 
    \item How should Model[no-X] be separated from Model[with-X]? Early in existing literature include:
\begin{itemize}
    \item After an attempt to access dual-use capabilities has been detected, a permission-checking layer identifies whether a user has passed review \citep{noemaresearch2024}. By instead making Model[With-X] broadly available, but then triggering permission checks only when dual-use capabilities are accessed, it becomes possible give more people access to Model[with-X]. Another key difference between this and present-day is that this is an extra layer on top of refusal training, which simple jailbreaking cannot bypass.
    \item A Model[no-X] might be achieved through unlearning of the capability, or filtering out dual-use data from the training data, and various other nascent techniques. Then, users who have passed the review process can get access to Model[with-X] through a separate, secured endpoint \citep{greenblatt2024managing}; otherwise, users get access to Model[no-X].
\end{itemize}
\end{itemize}
Staged deployment could also be employed by developers to gather data on how the access control should be configured. At a high-level, this involves releasing a weaker version of the model (e.g. with weaker capabilities or affordances like tool use) before releasing a fuller system, and then using the intervening time to identify misuse, gather data, and assess the effectiveness of mitigations in practice. Staged deployment enables iteratively fixing issues and gaining confidence in the overall safety of the system before deploying a more capable model to a wider audience. 

Overall, access control approaches in the AI field are still an emerging area. There are likely many options that have not yet been explored.

\subsection{Securing model weights}\label{sec:model-security}

Ensuring that attackers cannot get access to model weights is important both for protecting against misuse of the model and to protect IP. Protecting the models requires systematic defenses on multiple layers: more traditional defenses, such as identity and access control and environment hardening, and more modern defenses such as encrypted processing.

When prioritizing the levels at which to deploy defenses, in principle, all the stages of the model from data to pretraining to posttraining, including auxiliary classifier models, should be protected, as well as the model architecture besides the model weights. However, the likely most immediate target for the attacker are the model weights post-training. See \cite{nevo2024securing} for a detailed overview.

\subsubsection{Identity and access control}

Collecting data, pre-training and post-training large language models is an immense effort from many teams, which results in many people with access to the model weights. The first line of defense is to reduce access to “least privilege”, namely, only give access to individuals who need it and for only the time that they need it. This applies to both physical security (namely, who can access what buildings or clusters) and software security (allowlisting software). All identity must be verified, and multi-factor methods like security key enforced. A dedicated team is continuously supervising who has access to what part of the model infrastructure and how each individual uses the access to observe opportunities for reducing access and anomalies (e.g. early capture of social engineering).

\subsubsection{Environment hardening}

The environment exposed to the model should be hardened from pre-training all the way to post-training, and inference time, and enhanced with monitoring and logging, exfiltration detection and resistance, detection and remediation of improper storage. All the downstream applications that leverage model weights should equally be hardened. Downstream applications that interface with the model should do so through a narrow API, similarly to the use of safe proxies in production cloud systems~\citep{cloud2024protect}. This interface needs to be hardened to protect against break-in, and defenses for exfiltration detection or prevention deployed at this interface.

One challenge unique to language models is that weights can be exfiltrated in time through API use. 
\citet{carlini2024stealingmodel} show that an attacker interacting with the model’s inference API can over time exfiltrate approximate weights or architecture information from this interface. Model distillation from an API~\citep{truong2021data,sanyal2022towards} may also provide a method for a bad actor to recreate a model, though it is also possible that safety refusals would limit the success of such methods at recreating dangerous capabilities in particular. Should these threat models prove plausible, we would need methods to detect attempts at exfiltration, and rate limit or impose policies on access.

\subsubsection{Encrypted processing}

Ultimately, attackers or insiders can bypass some of the traditional security techniques above. Encryption in use, namely keeping the model and its weights encrypted at all time, promises to protect the model weights against attackers who managed to break into the servers. A major technology realizing encryption in use is confidential computing.  

Confidential computing technology consists of hardware extensions called “hardware enclaves” on both CPUs (e.g. AMD-SEV-SNP, Intel TDX) and GPUs (e.g. H100). Hardware enclaves have  keys fused into the hardware with which they encrypt the data leaving the processor, e.g. data in memory or on the memory bus. Hardware enclaves also isolate the sensitive process and data from the code running outside of an enclave. 

The model runs inside an enclave and a remote user can attest that it interacts with a correct enclave (an enclave from a legitimate hardware vendor running expected code) before interacting with this model. In principle, an enclave can be used for the entire model lifecycle, from pre-training to post-training to inference. Private inference protects both the model during the inference process and the prompt, and is an easier deployment. Deploying enclaves for pre-training and post-training requires bigger infrastructural changes, and makes it harder for ML engineers to debug and train these models. 

Research on cryptographic computing provides an alternative to hardware enclaves where the mathematical properties of encryption can be used for computation. However, these methods~\citep{pang2024bolt} are still inefficient at the scale of large language models.

\subsection{Societal readiness}\label{sec:societal-readiness}

While some mitigations focus on controlling risks directly within the model — such as limiting harmful capabilities or aligning the model’s objectives with human values — another approach is to proactively use AI systems to harden societal defenses against potential AI-enabled risks. For example, one could develop tooling to rapidly patch vulnerabilities in critical infrastructure, in preparation for AI cyber offense capabilities. This aims to ensure that despite AI assistance, the resources and capabilities needed to carry out severe harm scenarios stays relatively high.

Interventions could leverage the very AI capabilities that pose risks, and shift the offense-defense balance towards a more secure equilibrium. Below follows a list of some example AI-driven interventions that could improve societal readiness.

\noindent Within biosecurity (contra biological risks):
\begin{itemize}
    \item Credible and early pandemic detection, eg. by analyzing open-source data at EPIWATCH, or metagenomic sequencing to detect sequence fragments in wastewater exhibiting quasi-exponential growth
    \item Build privacy-preserving system enabling individuals to report infections in their community and track personal risk level and associated precautions
\end{itemize}

\noindent Within cyber defense (contra cyber offense risks):
\begin{itemize}
    \item Partner with critical national infrastructure companies (e.g. power utilities) to patch vulnerabilities
    \item Develop hardened software systems
    \item Improve the cybersecurity anomaly detection systems used to detect attacks by reducing false positive rates
\end{itemize}

\noindent Within info defense (contra persuasion risks):
\begin{itemize}
    \item Develop anti-scam tools. E.g. SynthID watermarks and identifies AI generated content.
    \item Develop anti-misinformation tools. E.g. community notes for the entire internet
\end{itemize}

\subsection{Red-teaming mitigations}\label{sec:misuse-stress-tests}

When models pass capability thresholds, threat modeling must assess the attack surface area through which threat actors could access dangerous capabilities:
\begin{enumerate}
    \item \textbf{Inference API:} A threat actor may simply make harmful requests to the model, but if this is well-defended, they could use more sophisticated techniques such as jailbreaks~\citep{shanahan2023role, anil2024many, bethany2024jailbreaking} or decomposing harmful requests into many individually benign queries~\citep{jones2024adversaries}.
    \item \textbf{Finetuning API:} A threat actor may use a finetuning API to train on harmful data to erode model safeguards, and then use the finetuned model to answer harmful requests. Again, if this is well-defended, they may use more sophisticated techniques such as covert malicious finetuning~\citep{halawi2024covert}.
    \item \textbf{Evading APIs:} A threat actor may access a model without using an API at all, for example by stealing the model weights and removing any guardrails~\citep{nevo2024securing}.
\end{enumerate}

Residual risk can then be assessed by red-teaming mitigations across the entire attack surface area. Red-teaming aims to reduce uncertainty about potential vulnerabilities or flaws in these safeguards that adversaries could exploit by examining how dangerous behaviours and capabilities might be elicited despite the safeguards. The overall adequacy of mitigations should be periodically reevaluated to address the potentially rapid emergence of new attack vectors and to ensure their effectiveness as models evolve throughout training.

A red-team test can follow these steps:
\begin{enumerate}
    \item Start with the safeguard safety case, i.e. a detailed analysis about why the set of misuse mitigations, once applied, is adequate for bringing risks associated with a particular set of critical capabilities to an acceptable level. 
    \item Identify some key assumptions that the safety case is predicated on. For example: ``our misuse detection classifiers detect when a user is attempting to elicit information about the formulation step of biological weapons''.
    \item Carry out tests to find where these assumptions fail. 
    \item If these examples are identified, the safety case has vulnerabilities. A developer can then assess whether the safety case is still adequate, or if additional R\&D must be carried out to patch the vulnerabilities.
\end{enumerate}

\subsubsection{Examples of tests}
There is healthy research literature looking for vulnerabilities in mitigations which might be considered actual examples of stress test steps 2-3, or models for how these steps can proceed in the future. \citep{feffer2024} conducts a comprehensive review of existing red-teaming literature, which we draw on to list four categories of test approaches.  A stress test might involve one of these categories, or a combination thereof:

\textbf{Manual tests} involve manual evaluation of inputs and outputs to models by humans. For example, some studies use crowdsourced workers to assess and rate LLM outputs \citep{ganguli2022redteaming, dinan2019}, while others engage domain experts \citep{openai2023gpt4}. Another approach, handcrafting jailbreaks, involves researchers creating specific prompts that intentionally attempt to bypass model safety constraints \citep{liu2023study}.

\textbf{AI-enhanced testing:} In brute-force and AI-enhanced testing, researchers automate the creation of test cases by leveraging AI to systematically explore LLM vulnerabilities \citep{ganguli2022redteaming, samvelyan2024rainbowteamingopenendedgeneration}. For instance, \citet{perez2022redteaming} utilizes language models to systematically generate test cases that reveal conditions under which models may produce harmful content. Similarly, \citet{doumbouya2024h4rm3l} introduces a domain-specific language (DSL) to formally represent jailbreak attacks, allowing automated synthesis of inputs. Additionally, \citet{liu2023masterkey} investigates prompt engineering strategies to find inputs that universally bypass the safety mechanisms of aligned LLMs.

\textbf{Algorithmic search methods:} In algorithmic search methods, researchers systematically alter input prompts until they trigger vulnerabilities in AI models. For instance, \citet{zou2023universal} demonstrates that greedy and gradient-based search techniques could be used to identify universally adversarial suffixes to append to prompts, effectively bypassing safeguards. Other work, such as \citet{ma2024evolving} and \citet{mehrotra2023}, describes methods where models repeatedly and automatically attempt to jailbreak target models until they succeed. Additionally, \citet{tsai2023} proposes a similar search strategy to diffusion models, perturbing inputs iteratively to generate harmful content even when safety filters are active.

\textbf{Targeted attack testing:} In targeted attack testing, researchers focus on manipulating a model’s architecture rather than broadly testing prompt variations. We also include exploits of specific post-training mitigations in this category. Studies have shown that safety measures like RLHF and instruction tuning can be bypassed with minimal fine-tuning \citep{qi2023finetuning, zhan2023removing}. Additionally, while unlearning methods often imply the full removal of target capabilities, evidence suggests they often only obscure or suppress them, leaving models vulnerable to techniques like jailbreaking or fine-tuning to regain those capabilities \citep{sharma2024concealment, deeb2024unlearning, li2024llmdefensesrobustmultiturn}. \citet{zhang2024simple} further demonstrates that applying quantization can even restore information that was intended to be ‘forgotten’ through unlearning. 

Since a bad actor might put in much more effort than we do, we should give our red teamers extra advantages to compensate for the difference in effort. For example, they could have full knowledge of the mitigations put in place, or we could loosen the thresholds for various mitigations to make the system easier to attack.

\subsubsection{Open problems and paths forward}
Safety cases for AI systems are starting to develop in AI~\citep{buhl2024, clymer2024safety} but are extremely nascent with respect to those in other industries. For example, early safety cases are likely to rely heavily on qualitative evidence; the methodology of taking stress tests of AI systems and converting that into a quantitative metric that non-ambiguously informs deployment green-lights is underdeveloped.
Red-teaming practices in AI are indeed still in their early stages. For example, parties often disagree about its definition, its goals, the makeup of teams, what severity level of outputs serve as concerning evidence, how results tie into governance processes, and more. See \cite{openai2023nist, executive_order_2023, anthropic2024challenges} for examples. As developers carry out red-teaming more, the process would ideally evolve to become more systematic and rigorous.

%% file: 06-addressing-misalignment/intro.tex
\section{Addressing Misalignment}\label{sec:misalignment}

\begin{figure}[t]
    \centering
    \includegraphics[width=\linewidth]{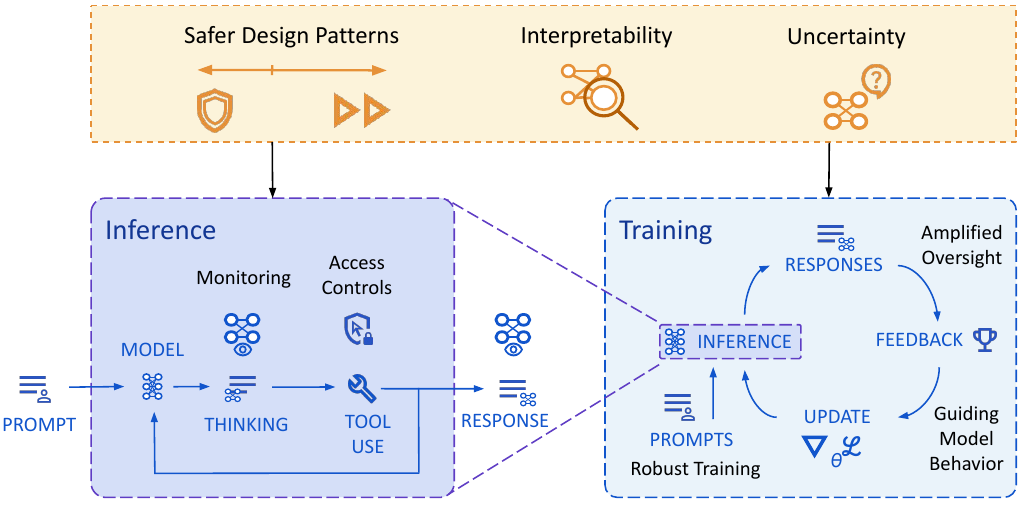}
    \caption{\textbf{Overview of our approach to mitigating misalignment.} The first line of defense aims to train an aligned model (``training'' box). Should this fail, the second line of defense aims to prevent severe harm even if the model is misaligned (``inference'' box). Note that inference mitigations continue to apply after training (that is, during deployment). In addition, while it is useful to monitor the AI system during training, we may want to avoid using the monitor as a training signal, as this could lead to the AI system hiding the bad behavior rather than removing it. Enablers such as safer design patterns, interpretability, and  uncertainty can strengthen other mitigations. While not depicted in this diagram, we also expect to conduct evaluations and stress tests to assess the necessity and sufficiency of mitigations, similarly to our approach in misuse (\Cref{fig:misuse-overview}).}
    \label{fig:alignment-diagram}
\end{figure}

\noindent Our strategy to address misalignment is illustrated in \Cref{fig:alignment-diagram}. In brief:
\begin{enumerate}
    \item Attain good oversight, using the AI system itself to help with the oversight process (Section~\ref{sec:amplified-oversight}), and use it to train the AI system (Section~\ref{sec:guiding-model-behavior}).
    \item Identify cases where oversight is needed so that the trained model will robustly do what we want (Section~\ref{sec:robust-training-and-monitoring}).
    \item Apply defense in depth to defend against AI systems that are misaligned despite our best efforts (aka ``AI control''), through a combination of monitoring with efficient and scalable oversight (Section~\ref{sec:robust-training-and-monitoring}), and the application of computer security techniques (Section~\ref{sec:hardening}).
    \item Assess the quality of the mitigation through alignment assurance techniques (Sections~\ref{sec:alignment-stress-tests} and~\ref{sec:alignment-safety-cases}).
\end{enumerate}

We also invest in \emph{enablers} that help improve a variety of these steps, including safer design patterns (Section~\ref{sec:safer-design-patterns}) and interpretability (Section~\ref{sec:interpretability}).

%% file: 06-addressing-misalignment/amplified-oversight.tex
\subsection{Amplified oversight}\label{sec:amplified-oversight}

Recall from section \ref{sec:misalignment-risks} that misalignment occurs when the AI system produces harmful outputs for intrinsic reasons that the designer/user would not endorse. There are two possible sources of misalignment: specification gaming, in which the designer-provided specification (e.g. reward signal) is flawed in a way that the designers did not foresee, and goal misgeneralisation, in which pursuit of an undesired goal arises in situations which are out of distribution compared to training. Amplified oversight, our focus in this section, aims to mitigate the specification gaming source of misalignment.

In order to describe amplified oversight, it is helpful to recall from section \ref{sec:misalignment-risks} the principle of informed oversight: it would be sufficient to rule out misalignment if we a) know everything that the AI  "knew" when producing its outputs and b) always correctly endorsed those outputs. We can somewhat reduce the ambition of achieving informed oversight, by aiming to achieve informed oversight on a single input-output pair – this is the focus of amplified oversight research. It aims for a human to provide feedback for a superhuman AI input-output pair which is as good as could be achieved if that human:
\begin{enumerate}
    \item understood all of the reasons that the AI system produced that output; and 
    \item had unlimited ability to reflect on the correct decision.
\end{enumerate}

Since the AI system is superhuman (by hypothesis), it will choose actions for reasons that it understands but that humans would not understand by default. Thus, a core intuition in amplified oversight is to use the AI system itself to make these reasons legible to oversight mechanisms. A prototypical example that we will return to below is debate, in which two copies of an AI compete to point out flaws in each other’s output to a human judge. It is hoped that exposing flaws will help to surface some of the knowledge used by the AI to produce its output, because if the AI provides an answer that isn't to the best of its knowledge, its opponent could point that out and win the debate. The difficulty of identifying and explaining a subtle flaw is thus offloaded to the superhuman AIs, with the human doing the relatively easier task of evaluating the highlighted flaws. 

For example, consider the case of an AI trained to write long and complex legal contracts. Given such a contract of several thousand pages, it would take even expert human lawyers a very long time to verify that the legal text achieved the intended goals of the humans who requested the contract. Thus, labeling even a single input-output pair would be prohibitively expensive. On the other hand, given a second AI trained to point out flaws in legal contracts, it would be sufficient to identify a flaw in a few lines of the legal text. A human presented with these lines could then quickly judge if they do indeed constitute a flaw relative to the original human intent for the contract.

Importantly, amplified oversight techniques generally maintain humans in the loop—for training, evaluation and monitoring—to ensure that their values are upheld. Some aspects of human judgments can be imitated or delegated, according to the human’s evolving trust in the AI. The form of human involvement in the amplified oversight protocol is an open question.

\subsubsection{Methods}\label{sec:ao-methods}
We’ll now describe some amplified oversight methods, with a focus on how these appear in the modern era of large language models (LLMs). Our descriptions are framed as outlining  particular problems that occur, and approaches to overcome them. 

We begin with the case in which amplified oversight is actually unnecessary, when the desired behaviour is simple enough that either the data demonstrating it already exists, or it is sufficiently cheap/easy for a human to demonstrate it. 

If the desired behaviour can be algorithmically verified, then this verification signal can be used to generate desired behaviour, e.g. using it as a reward in reinforcement learning, or in a scheme such as Self-Taught Reasoner \citep{zelikman2022star}.  

\noindent \textbf{RLHF.}
Often an LLM’s behaviour after pretraining is often not precisely as desired for applications (e.g. a chatbot/assistant that is helpful, harmless and honest) and there is no algorithmic verification signal to easily improve it. One approach can be to gather human demonstrations of desirable behaviour and do supervised fine tuning~\citep{ouyang2022training}. However, there are difficulties with this: generating such human demonstrations can be somewhat costly, and the demonstrations don’t convey comparative information between different responses. An approach to this is to have humans act as raters of AI-generated outputs – it is somewhat easier, faster and cheaper for the humans to evaluate whether the generated output is desirable, and to provide extra comparative information by contrasting different model responses. So further fine-tuning is done using reinforcement learning from human feedback (RLHF) \citep{christiano2017deepreinforcement}. The LLM is fine-tuned via RL using a reward from a learnt reward model, which is trained on human preference data, over AI-generated outputs \citep{bai2022constitutional,glaese2022sparrow,ouyang2022training,stiennon2022learningsummarize}, see also DPO \citep{rafailov2023DPO} which uses preferences but avoids the use of RL. 

\noindent \textbf{Critique.}
The shift from human demonstration to human evaluation has an analogy to the complexity classes P and NP: decision-problems that can be solved easily and those which can be verified easily. But just as there are problems harder than P or NP in complexity theory, there may be AI-generated behaviours which are too difficult for a human to evaluate. A natural approach to improve the performance of human raters is to show them AI-generated critiques to assist the human with rating AI-generated behaviours, for example by pointing out mistakes. \cite{saunders2022critique} gather human data consisting of natural language critiques of text and then use supervised learning to fine-tune an LLM to generate critiques. They show that these critiques consistently help both humans and other LLMs identify issues in written text (e.g. topic-based summaries). \cite{mcaleese2024llmcritics} train critique models using RLHF to catch bugs in model written code. They find that the human raters using the critique models catch more bugs than unassisted human raters. Constitutional AI \citep{bai2022constitutional} takes a constitution of rules for an AI system to form critiques and revisions, using these as input for both SFT and RL fine-tuning (they call this RL from AI feedback, or RLAIF).

\noindent \textbf{Debate.}
Perhaps, though, it may be difficult for human raters to evaluate whether the critique itself is accurate (or to verify that the constitution has been correctly specified and applied). An approach is to use further AI assistance to critique the critique. The natural extension of this is AI debate \citep{irving2018debate} mentioned in the introduction, in which two copies of an AI compete to point out flaws in each other’s output to a human judge. The complexity class analog of debate can evaluate any decision-problem in PSPACE using only a polynomial time judge. This allows a human judge to evaluate arguments along a single path of an exponentially larger tree of all possible arguments, and optimal play for the debaters takes into account the whole tree. 

A key requirement for debate to succeed is that it is harder to avoid getting caught in a lie than to refute a lie – whether this holds is a major focus of current work. Further theoretical refinements of the debate protocol, which make it harder to avoid getting caught in a lie than refute a lie, are an active area of research. In the original protocol, the honest debater is assumed to have unbounded computational power, whereas \cite{brown2023scalable} relax this requirement so that the honest debater can argue for the truth efficiently, even against an unbounded dishonest agent. Another problem that could arise is path-dependent lying in which a dishonest debater has more flexibility than an honest one, as they could claim a proposition to be true in one path through the tree of possible arguments, but false in another. 

An approach to this is cross-examination \citep{barnes2020progress} by allowing debaters at any point to erase their opponent’s memory of the current context of the debate, and directly ask their opponent to commit to whether a proposition is true or false, which is then carried back into the main debate. Under this modification, a contradiction to that proposition is penalised and path-dependent lying is rendered suboptimal. The complexity class analogue of debate with cross-examination replaces PSPACE with the more complex NEXP. Further problems still remain, such as obfuscated arguments - that it is possible to break down a claim into sub-claims, where only one of the sub-claims is false, but neither debater knows which one \citep{barnes2020debate}. A dishonest debater can exploit this by making such an obfuscated argument, for which their opponent can do no better than random guessing, and hence can’t efficiently point out exactly where the dishonest claim is for the human judge to verify. Understanding if and how one can further modify debate in order to avoid the obfuscated arguments problem is an area of active research.

In addition to theory there has been more empirical work on debate recently, both with humans and LLMs. \cite{irving2019socialscientists} introduce the idea that debate will only usefully amplify judge capabilities once judges meet a minimum threshold of capability.  This work highlights the key need for experiments that involve humans to determine whether human judges fall above or below this threshold. Initial experiments on debate have focused on the QuALITY dataset \citep{pang2021quality}, a reading comprehension task with information-asymmetry, where the judge is artificially limited by only being allowed to see short verified quotes from the text provided by an LLM debater. Two studies which used human debaters and judges across one-turn \citep{parrish2022singleturndebate} and two-turn \citep{parrish2022twoturndebate} debates failed to significantly improve judge accuracy. Later work found debate to be effective with strong human debaters, but ineffective when those humans are replaced with GPT-4 debaters \citep{michael2023debate}. \cite{radhakrishnan2023anthropic} reports promising initial results with training LLM debaters and judges on the QuALITY dataset. \cite{khan2024debating} consider a similar setup to \cite{radhakrishnan2023anthropic}, providing a comprehensive inference-only (i.e. no fine-tuning, though using e.g. best-of-N sampling) analysis of debate on the QuALITY dataset with LLM debaters and weaker LLM and human judges. 

\citet{kenton2024scalable} conduct experiments across a broader range of tasks, including those without information-asymmetry; include additional oversight protocols (open debate and open consultancy); and provide more extensive ablations, resulting in different conclusions compared to \cite{khan2024debating}. In particular, without information-asymmetry, debate does not significantly outperform directly asking the question to the weak LLM judge. \citet{rein2024nyu} explored LLMs debating about code with human judges who had no coding experience, applying RLHF, resulting in mostly negative findings. \citet{arnesen2024training} again used the QuALITY dataset and performed self-play debate training with a multi-step variant of DPO using preferences from a fine-tuned, calibrated judge, finding this modestly improves the judge accuracy by about 4\%. 

The core difficulty here is designing studies that simulate the capabilities gap that will occur with humans supervising superhuman AI – currently done by introducing information-asymmetry or LLM capability asymmetry. However, each of these approaches has drawbacks. Hiding information leaves the judge’s reasoning abilities intact, and provides a clear operationalization of intrinsic reasons that affect an AI’s behavior. However, the future capabilities gap between judges and AIs is likely to include many additional important aspects (e.g. reasoning ability, persuasiveness)  beyond knowledge of additional information. On the other hand, using less-capable LLMs as judges may capture other aspects of the capabilities gap, but may excessively limit the ability to reason about arguments in a debate, resulting in a very noisy training signal; or the capabilities gap may not accurately reflect the capabilities gap between humans and superhuman AI that we are most interested in. 

Selecting appropriate tasks for debate experiments presents additional challenges. The advantage of methods like debate over standard RLHF arise in settings where there is a large and complex space of possible outputs that the AIs can recursively explore. Thus there is a tension between selecting a task that is too simple, and thus can be solved quite well by RLHF, and a task that is too complex for current AI systems, such that training with amplified oversight never gets off the ground.

\noindent \textbf{Other recursive AI assistance.}
Iterated amplification \citep{christiano2018supervising} progressively builds up a training signal for difficult problems by combining answers to easier subquestions. The burden here is initially on the human to combine answers to subquestions. Recursive reward modelling (similar to iterated amplification) \citep{leike2018scalable} proposes to use RLHF to train a number of agents to solve simpler subproblems. It then leverages those agents to solve harder problems in a recursive manner. The difficulty here is in deciding what to use as the simpler subproblems to train the helper agents. These approach shares many similarities with debate, see \cite{irving2018debate}.

\subsubsection{Human-AI complementarity}

A method for amplifying human raters is to provide them with AI assistance (for example, showing them model-generated critiques, as described in a previous section). Teams of humans and assistant AIs will work best when the human and AI have complementary strengths or tradeoffs. If designed well then any AI assistance method should allow the Human-AI team to outperform either individually~\citep{bansal2021complementary, ma2023appropriatetrust, vaccaro2024combinations, Vasconcelos2023overreliance, Zhang2020trustcalibration}.

Designing systems that allow human raters to work well with AI assistants can be viewed as a human-computer interaction (HCI) problem, a domain with a large body of existing research to draw on. The way the assistance or critique is presented to the human rater could have a meaningful impact on performance; there is a risk of “over-reliance” on the AI, for example if the rater is “spoon-fed” the answer with a persuasive explanation then they might always defer to the AI assistant. This could negate the benefit of having a human in the loop. Prior work in AI-Assisted Decision-making has focused on what information from the AI to show to the human rater to ensure both appropriate reliance and complementary performance \citep{Lee1994trust, lai2019human, Lee2004trust, ma2023appropriatetrust, Manzini2024trust, rechkemmer2022confidence, si2023large, vaccaro2024combinations, Vasconcelos2023overreliance}.

Another way to leverage complementarity is hybridisation (or data ensembling), where we combine individual labels from AI auto-raters and human raters. This could involve having humans and AIs each rate all the examples and then combining them \citep{li2024annotation}, or having a procedure for selecting how each example will be rated (also called delegation) \citep{fugener2022cognitive, hemmer2023human, wang2024annotation}.

\subsubsection{Human biases}
Under RLHF, the AI is optimised to generate responses that the human rater endorses. Psychological research, however, has shown that human judgment and decision-making is influenced by biases and heuristics, sometimes leading to sub-optimal decisions that upon deeper reflection, humans would not endorse~\citep{evans2003two, kahneman2011thinking, tversky1974judgment}. These biases likely also affect human rater preference judgments making it more challenging for a Reward Model (RM) to learn a true model of human preferences and values. For example, a documented bias in RMs is a length bias, rewarding longer responses rather than responses genuinely higher in quality \citep{zheng2024judging}. RMs may be picking up on a spurious correlation in the preference dataset where humans are rewarding informativeness, which happens to correlate with length, but, of course, is not synonymous with it. It's possible, though, that human raters might also have a baseline preference or bias for longer responses, even if they don't meaningfully differ in quality. What's more, if raters have difficulty critically evaluating the content of a model's output or aren't motivated to do so due to cognitive fatigue or lack of care, they may fall back on a "longer is better" heuristic exaggerating the correlation between endorsement and length that the RM later exploits during RLHF.

Documented human cognitive biases that could influence human rater judgments include primacy and recency effects~\citep{murphy2006primacy, rey2020primacy}, egocentric bias~\citep{ross1977false}, self-enhancement bias~\citep{kramer1993self}, acquiescence bias~\citep{krosnick1991response}, risk-aversion~\citep{kahneman2013prospect}, and myopia~\citep{fiedler2023metacognitive}. The egocentric bias, for example, might be what leads human raters to endorse responses that align with their preferences, beliefs, and worldviews, leading to other known biases in RMs, such as rewarding sycophantic responses \citep{perez2022discovering, sharma2024understanding}. While these categorizations help in organising behavioural research they should not be construed as distinct or exhaustive. Indeed, as AI improves and the rating task gets more challenging, new biases may become relevant. Bias is an issue for all oversight methods, that can be addressed by either attempting to avoid situations with extreme bias, or by explicitly correcting for bias by “inverting” a model of human behavior \citep{evans2016learning}. An ambitious approach would be to learn a model of relevant human biases \citep{pmlr-v97-shah19a}.

\subsubsection{Inferences without succinct justification}
Most approaches to amplified oversight rely on AIs generating explanations or arguments to justify their statements and actions. However, there may be tasks where the reasoning or justification for a particular action is non-verbalisable. Such behaviour may arise from statistical correlations in a large dataset, or from concepts which don’t have a particularly succinct natural language description, but which can be learned from experience. Again this is an example of an intrinsic reason that an AI might take a certain action, and methods for amplified oversight should be able to effectively take such reasons into account, even if they cannot be made explicit.

\subsubsection{Sidestepping improving human oversight}
The amplified oversight approaches considered above (debate, IA, RRM) all attempt to improve human oversight training signals, extending them to complex problems. Another approach to the problem is to accept that we have a flawed training signal and instead achieve this through weak-to-strong generalisation \citep{burns2023weak}, in which some change is made to the inductive bias of the model training setup, e.g by modifying its architecture, training objective, optimiser etc. such that we get desirable generalisation from weak human oversight signals, or through hybrid approaches \citep{radhakrishnan2023scalable}. The core difficulty is that we currently do not have a good theoretical understanding of how to design inductive biases in such a way as to provide guarantees that the generalisation will be as desired. 

Relatedly, \citet{christiano2021eliciting} outlines the problem of eliciting the latent knowledge of a model, suggesting that improving human oversight training signals may not be sufficient, and that some form of regularisation may be required. One such approach is imitative generalization (also known as learning the prior), in which the goal is to train the AI to generalize from data in the same way a human would, rather than directly training on human labels for each piece of data \citep{barnes2021imitating,christiano2020learning}. Unsupervised learning offers another angle on this. For example, one could enforce unsupervised logical consistency relations in domain-specific ways \citep{fluri2024evaluating};  by arbitraging model predictions \citep{paleka2024consistency}; and via negation-consistency on binary choice questions, such as in contrast-consistent search (CCS) \citep{burns2022discovering}. CCS aims to discover latent knowledge in an unsupervised manner, through optimising for an unsupervised logical consistency objective, resulting in a probe which sometimes classified true statements. However, \citet{farquhar2023challenges} showed that arbitrary features are also optimal under the logical consistency objective, and that empirically the learnt probes merely classify along the most salient features of the underlying dataset, rather than true statements. \citet{laurito2024cluster} propose a normalisation to attempt to remove unrelated features. Unsupervised probes have also been trained for a range of concepts \citep{zou2023representation} and for making predictions \citep{mallen2023eliciting} (see also \cite{hase2023methods} and references therein).
A key challenge for unsupervised learning approaches is familiar to all unsupervised learning, the issue of a lack of identifiability guarantees. For example, how to distinguish between discovering the model’s actual knowledge/ground truth, from that of a simulated entity’s knowledge/truth (analogous to the problem of disentangling representations without supervision \citep{locatello2019challenging}).

Other ways to use weak human supervision have been considered. Several other methods to train probes to discover latent knowledge use supervised learning \citep{Azaria2023-ew, li2024inference, marks2023geometry, wang2023gaussian}. There are also attempts to detect dishonesty by supervised learning on LLM outputs under conditions that produce honest or dishonest generations \citep{pacchiardi2023catch}. While these work well when the ground truth labels are provided by humans, we don’t know how well they will generalise to questions too difficult for humans to answer.
Core challenges for this range of approaches remain: absent dramatic theoretical advances in our understanding of generalisation, or unsupervised learning, progress would most likely be empirically driven, as compared to amplified oversight approaches, for which the protocols have both theoretical underpinnings and empirical validation.

\subsubsection{Discussion}
Amplified oversight leverages the power of AI systems to amplify the human oversight signal. While this may significantly broaden the types of outputs that humans can evaluate, such methods also introduce new difficulties. Fundamentally, amplified oversight requires the careful design of training protocols so that we have good reasons to believe that human oversight will be correctly amplified to situations that humans cannot directly understand. This is quite different from the case of having AIs directly generate human-understandable explanations for each output, which might involve convincing, but bad-faith explanations. For example, in debate (which we run with as a prototypical amplified oversight method), it is important to structure the debates so that clever strategies for producing arguments that are difficult to refute cannot be used to defeat the honest strategy of simply telling the truth and explaining it well. If it is not possible for an honest debater to efficiently counter bad-faith arguments, then AI debates will be unable to achieve the overall goal of amplified oversight. While this seems like a vague, yet plausible issue, these types of difficulties can be given complexity-theoretic formalizations, allowing for theoretical progress on amplified oversight protocol design in advance of improving AI capabilities. See section \cref{sec:path_forward} on `Path forward'.

Concurrently with progress in amplified oversight theory, AI systems have recently reached the level where they are able to produce outputs that are difficult for non-expert and/or time-limited humans to understand, or even expert humans \citep{openai2024learning} on the difficult physics, chemistry and biology questions in the GPQA benchmark \citep{rein2023gpqa}. This provides an opportunity to also study amplified oversight methods empirically, presenting new challenges beyond those encountered for RLHF. While the goal is to provide accurate feedback on a single input-output pair, amplified oversight introduces additional intermediate tasks which can be more difficult for current AI systems. For example, AI debate requires multi-turn dialogue and the ability to generate convincing arguments, as well as understand and refute arguments from an opposing AI. Thus, training models via debate, requires multi-turn reinforcement learning in a game with an extremely complex strategy space. Additionally, the task facing a judge in a debate is more demanding than typical RLHF, as it requires the ability to reason about the correctness of opposing arguments. This can be a problem both for actual human judges, who have good reasoning abilities, but limited patience and attention, as well as LLM judges, which may need significant additional training in order to reason correctly about opposing arguments. A core difficulty in this empirical approach is in the design of tasks and capability gaps which are reflective of the (unknown) future capability gap between human and superhuman AI on the tasks we most care about.

\subsubsection{Path forward}
\label{sec:path_forward}

The path forward for amplified oversight research can progress along many tracks. While many approaches could be pursued, we plan to focus our work most heavily on debate, because of the combination of its compelling theoretical properties (allowing us to study some aspects of AO before we have superhuman AI) and because LLMs allow us to empirically study some aspects of how debate may work in practice. The main contrast, e.g. to weak-to-strong generalisation is that we are more hopeful about our ability to improve oversight signals through debate (currently due to our belief about debate’s theoretical properties) than in our ability to understand and control generalisation to the degree necessary to rely on the oversight scheme. 

Continued theoretical investigation into debate protocols is crucial, focusing on addressing issues like path-dependent lying, obfuscated arguments, and other novel strategies that may offer an advantage to a dishonest debater. Further, more work is needed to adapt these theoretical studies to more accurately model the real-world situation of humans, and trusted LLMs, providing oversight to superhuman AIs, rather than just concepts in complexity theory. 

Empirical research should focus on designing experiments that accurately simulate the future capabilities gap between humans and superhuman AI, utilising both LLMs and human studies to explore the dynamics of debate in realistic settings.  A key challenge lies in task design, striking a balance between simplicity and complexity to effectively evaluate the benefits of amplified oversight techniques and highlight their relative performance, and modifications to improve their performance.  Furthermore, integrating insights from self-improvement research and exploring how these techniques can be adapted for amplified oversight is essential. This research should emphasise human-AI complementarity, ensuring that amplified oversight methods effectively leverage the unique strengths of both humans and AI to achieve robust and reliable oversight of increasingly capable AI systems.

A crucial step towards demonstrating the real-world impact of amplified oversight lies in applying it to tasks where AI systems can potentially surpass human capabilities. LLMs could engage in debates about complex coding challenges or the factual accuracy of information, with human experts acting as judges. These experiments would not only test the effectiveness of debate in ensuring truthful and reliable outputs but also showcase the wider potential for amplified oversight to enable human supervision of superhuman AI. Successfully demonstrating the value of amplified oversight in such real-world scenarios could pave the way for its widespread adoption and significantly impact the development and deployment of advanced AI systems within the next few years.

Even more ambitiously, amplified oversight could be applied to tasks with profound societal implications. Picture AI systems debating the optimal design for a revolutionary carbon capture technology, with expert human scientists judging the arguments to identify the most promising and impactful approach (which they may not have been able to develop unassisted). Or pushing the boundaries of our knowledge  and developing world-changing technologies for improving billions of people’s lives in critical areas like drug discovery. 

These problems are highly complex, and any AI suggested solutions are also likely to be similarly complex. We want to leverage AI's superhuman abilities, but this makes it tricky for humans to confirm that these are indeed the correct solutions, and that there are no unforeseen problems. So in order to get the most benefit from superhuman AIs we will need to amplify our ability to check their output.

Amplified oversight, in these scenarios, could serve as a powerful tool for channelling the intellectual power of advanced AI towards solving humanity's most pressing challenges, driving progress in scientific discovery, technological innovation. While these applications may seem distant, they represent the transformative potential of amplified oversight to not only guide AI development but also to empower humanity to make more informed and impactful decisions. All this may be possible, in addition to the safety advantages that would come with amplified oversight incentivising honest, rather than deceptive and dangerous behaviours.

%% file: 06-addressing-misalignment/guiding-model-behavior.tex
\subsection{Guiding model behaviour}\label{sec:guiding-model-behavior}

Designers have a policy in mind for how the AI should behave. For example, perhaps we want to ensure that the AI treats all users with equal courtesy, avoids preaching to the user while also not ``sucking up'' by shallowly mirroring their beliefs~\citep{sharma2024understanding}, and follows the user's instructions unless it violates some other part of this policy. The role of technical alignment is to align the AI to a specific high-level policy, so that it ends up doing what we want as specified by the policy.

Once we have an expensive oversight signal that can distinguish between good and bad behavior (\Cref{sec:amplified-oversight}), we need to use this signal to build an AI system that behaves well. This is a significant area of research, forming a major portion of model post-training or finetuning. The key constraint is that we only have a fixed supervision budget (e.g. money for paying human raters, compute for running autoraters, or limited quality of available rater feedback), and so we want good data efficiency and generalization when using our oversight signal.

\paragraph{Reward modeling.} One of the simplest approaches is to build a system that can cheaply approximate the expensive oversight signal, that can then be used more widely. This has typically been done by training reward models as in reinforcement learning from human feedback (RLHF)~\citep{christiano2017deepreinforcement, ziegler2019fine}. As LLM capabilities have improved, it has become possible to simply prompt LLMs for oversight, as in Constitutional AI~\citep{bai2022constitutional}. Deliberative alignment~\citep{guan2024deliberative} uses inference-time compute to reason about safety specifications and check those specifications against its current task.


\paragraph{Data quality.} Less is More for Alignment~\citep{zhou2023limaalignment} curates ~1,000 data points for instruction finetuning and achieves strong results, showing that data quality is extremely important.

\paragraph{Inference-time guidance.} Inference-time guidance of model behavior can be both cheap and effective, since the methods can leverage information about the specific prompt under consideration when guiding model behavior. Prompting is the most obvious approach, though both its cost and effectiveness likely suffer as the number of instructions in the prompt increases. For example, it would likely be infeasible to implement a fine-grained policy for refusal behavior entirely through prompting.

Model internals allow for more effective approaches. Existing interpretability methods are more than sufficient to provide meaningful information and transparency, with SOTA probe generalisation accuracy of at least 95\% for multi-turn interactions with incentive to lie~\citep{burger2024truth}. Conditional activation steering~\citep{lee2024programmingrefusal} is a cheap inference-time method which selectively adds a refusal steering vector when appropriate, showing promise at more fine-grained control. InferAligner~\citep{wang2024inferaligner} uses activation vectors to defend against jailbreaks and reduce toxic behaviour. \citep{chen2024designing} design a dashboard that enables understanding and control of conversations with a chatbot.

It is also possible to leverage inference-time compute to more effectively follow policies~\citep{guan2024deliberative}. By conducting several forward passes per token, RAIN~\citep{li2023rain} substantially decreases harmfulness.

\paragraph{Improving steerability during pretraining.}
Typically, practitioners focus on aligning model behavior during post-training. Pretraining-based alignment has been relatively underexplored, likely due to the huge compute costs of pretraining. Nonetheless, it presents its own opportunities. \cite{korbak2023pretraining} labels pretraining sentences with control tokens `good' and `bad', showing striking toxicity reductions compared to normal pretraining and finetuning. Another avenue is explored by Self-Instruct~\citep{wang2023selfinstruct}, which prompts the pretrained model to generate its own helpfulness tuning data. Perhaps knowledge distillation~\citep{hinton2015distillingknowledgeneuralnetwork} can be a cheaper proxy than full pretraining when figuring out how to improve the steerability of pretraining.

\paragraph{Denser feedback modalities.}
Each preference comparison can only provide at most a single bit of information (or slightly more if draws are possible), even if significant thinking went into the determination of which answer was better. Ideally, we would extract more information from each such comparison, for example by getting corrections or language feedback~\citep{jeon2020reward}.

For long trajectories, one option is process-based supervision~\citep{farquhar2025mona,lightman2023letsverify, luo2024improvemathematicalreasoning}, in which feedback is given on each step of the trajectory rather than a single reward at the end of the trajectory. This can provide significantly more information about which steps were good, which can help solve the credit assignment problem, enabling faster learning.

%% file: 06-addressing-misalignment/robust-training.tex
\subsection{Robust AI oversight for training and monitoring} \label{sec:robust-training-and-monitoring}

Human oversight is, and will continue to be, expensive. Even if we presume that techniques from Amplified Oversight (\Cref{sec:amplified-oversight}) make it possible to determine the acceptability of a given AI action, for very powerful future models, this will be too costly to do for all actions during training or (even more so) deployment.

The limited expensive oversight will likely be sufficient to ensure that the AI system does not cause harm in the average case, but need not prevent situations where the AI system causes significant harm through \textit{rare failures}, particularly due to distribution shifts~\citep{shah2022goal}. For example, an AI system may behave well during training, but could execute a treacherous turn~\citep{bostrom2014superintelligence} after observing compelling evidence that it is in deployment~\citep{christiano2019worst}. Alternatively, an AI system may be aligned in most contexts, but act in a misaligned way on certain rare contexts. For example, Claude 3 Opus fakes alignment in contexts where it perceives its values to be threatened~\citep{greenblatt2024alignment}, despite usually behaving in an aligned manner.

This motivates the need for \textit{robustness}, to guard against even these rare failures. A solution which is gaining traction rapidly is to delegate oversight to another, cheaper AI system. At training this takes the form of an AI \textit{judge} or \textit{reward model}, while at deployment this is an AI \textit{monitor}; for simplicity we will call all of these uses judge models. We will refer to the AI system that is being overseen as the AI \textit{policy}.

The use of cheaper AI judges enables scaling oversight to vastly larger numbers of inputs, which in turn can improve the robustness and reliability of the AI system that the judge oversees. However, cheap judges will not be as accurate and precise as expensive oversight, and will typically be weaker than the AI systems they oversee. To achieve the benefits of scalability without losing the benefits of amplified oversight, the judges themselves should have a variety of properties that make them more robust and reliable. Many of these desirable properties are shared for training and monitoring. We will thus begin by identifying and discussing key properties and promising approaches for judges, and then discuss aspects unique to training and monitoring.

\subsubsection{Judge safety} \label{sec:judge-safety}

Given that we are using an AI judge to help us address misalignment risks from an AI system, we must also consider the risk that the judge itself is misaligned. Since judges are only used to evaluate outputs from a policy, they will likely have much more limited affordances for influencing the real world, which significantly limits the risk. However, one key risk that remains is \textit{judge-policy collusion}, in which a misaligned AI judge chooses not to flag bad behavior from the policy~\citep{greenblatt2024control, shlegeris2024untrusted}.

One natural approach to judge safety is \textit{bootstrapping}: when aligning a new AI system, use the previous generation of AI (that has already been aligned) as the judge. Bootstrapping works well to the extent that we are relying on our first line of defense, that is, preventing AIs from being misaligned at all. In this case, we should expect our judge to ``try its best'' at judging, rather than colluding with the policy. However, bootstrapping does not trivially work with the second line of defense, which seeks to mitigate harm even if the AI is misaligned. This is because this second line of defense is \textit{application-specific}: mitigating harm from a misaligned AI agent that is writing security-critical code is very different from mitigating harm from a misaligned AI judge that may collude with the policy.

This motivates measures that focus specifically on mitigating harm from collusion by misaligned judges. The simplest approach is to use judges that are not capable of collusion, which can be assessed through capability evaluations for situational awareness and strategic reasoning~\citep{phuong2024evaluating, mallen2024subversion, laine2024sad}, discussed further in \Cref{sec:misalignment-inability-safety-case}. For more powerful AI judges, a more sophisticated approach is needed.

\cite{shlegeris2024untrusted} studies this setting under the name ``untrusted monitoring'', reflecting the assumption that the AI judge may itself be misaligned. One key advantage in this setting is that for the judge to enable harm by allowing a policy to continue with a specific bad behavior, it must reliably avoid flagging that behavior. (If the judge only avoids flagging the behavior half the time, the other half is likely still enough to train the policy not to perform that behavior, or to catch the policy at deployment time and determine it is misaligned.) Thus, \textit{testing} is especially useful for judge safety: we can construct synthetic examples of policy failures, and check that the judge does flag these failures. The judge should not be able to distinguish between synthetic and real examples, which we might try to achieve by producing synthetic examples by taking real data and modifying it only slightly.

Testing can still fail if the judge can distinguish and policy can coordinate on a specific feature of inputs on which to fail, since we may not test inputs with that particular feature. However, this coordination may be quite challenging to achieve. \cite{shlegeris2024untrusted} suggest \textit{paraphrasing} policy outputs and \textit{redacting} irrelevant information before running the AI judge, to reduce the channels through which a policy and judge could coordinate.

\subsubsection{Generalization}

It would be ideal for judges to generalize to new situations, as this is often where they are most important. Simply increasing the amount of data used has been seen as a promising approach to improve generalization~\citep{radford2021learningtransferablevisualmodels, hestness2017deep}. However this may scale poorly as the amount of data needed to achieve a certain level of generalisation can be exponential in the size of the input space~\citep{gilmer2018adversarial}. \cite{howe2024effects} find that for LLM classifiers, as the LLM scale is increased, adversarial training becomes a better defense, though it is still easy in absolute terms to find attacks.

Generalization is a hard and ill-defined desideratum, but there are some specific types of generalization that may be promising. Ideally, the judge would reliably generalize from simple common cases, where oversight is easier, to more complex cases.
In general these are types of normative extrapolation, where a principle of reasoning can be adopted to specify correct generalization from simple cases. For instance, compositional generalization~\citep{lake2018generalization} has been studied and seen substantial improvements in recent years.

While these limited forms of generalization appear achievable, it is unlikely that AI judges will be able to generalize perfectly to arbitrary questions, as any useful AI system must have bias~\citep{mitchell1980need}, which is unlikely to exactly match the desired bias.
As a result, to handle cases in which generalization cannot be expected, it is important that the AI judge is aware of its own limitations and has appropriate uncertainty, including out of distribution.

\subsubsection{Uncertainty estimation}

Since the judge will often be weaker and cheaper than the AI system it is overseeing (to enable scalability), a key property of the judge is to seek out further guidance when needed. This ability to ask for further guidance is what distinguishes AI judges from amplified oversight (\Cref{sec:amplified-oversight}), which must instead reliably produce an overall judgment. For the judge to seek out further guidance appropriately, it must know what it doesn't know. Thus we would like the judge to have well-calibrated uncertainty with respect to the most reliable oversight signal available.

While calibrated uncertainty comes for free when training against the target distribution using a proper scoring rule such as cross-entropy loss, this is certainly not the case for judge models derived from general purpose language models. Indeed, language models are spotty at reporting their uncertainty~\citep{kadavath2022language}.

There are many approaches to improving uncertainty estimates for deep learning~\citep{abdar2021review}, including Bayesian neural networks~\citep{jospin2022hands}, MC dropout~\citep{gal2016dropout}, and ensembles~\citep{lakshminarayanan2017simple}. However, these are often difficult to scale to LLMs, and as a result uncertainty quantification for LLMs often instead leverages the LLM's own ability to reason in natural language~\citep{shorinwa2024survey}. For example, there are promising results from teaching an LLM to verbalize its uncertainty~\citep{lin2022teaching}, measuring the semantic consistency of the LLM's outputs under perturbations~\citep{kuhn2023semantic}, and probing the internal activations of the LLM~\citep{kossen2024semantic}.

Nonetheless, many challenges remain, including some that are unique to misaligned AI systems. The policy may understand that it is being judged, and behave differently (and adversarially) as a result. While this bears a resemblance to defending against human adversaries (such as in cyberdefense), this property is atypical for AI systems, which often assume stationarity of the environment. In addition, as AI systems become ever more powerful, our uncertainty estimation will also have to handle more ambiguous questions where we ourselves have high uncertainty.

\subsubsection{Resource efficient oversight} \label{sec:resource-efficient-oversight}
If we have a family of judge models (or systems) that trade off expense with quality, we can manage costs by aiming for \textit{resource efficient} oversight.
A central problem is thus allocating enough resources to consequential and harmful actions to be able to notice them, before they are taken. Doing this well requires us to:
\begin{itemize}
    \item Cheaply pass through low-consequence actions.
    \item Quickly evaluate obviously harmless or harmful actions.
    \item Be good at allocating enough effort to consequential and opaque actions to determine if they are safe or not.
\end{itemize}

Similar problems have been encountered in related fields. For example, in cyber-threat detection systems the vast majority of logs that are generated represent safe behavior, but a small number of signals might represent users who are trying to evade detection and do something you do not want. Similarly, in many signals intelligence settings, the vast majority of intercepts are not worth much effort but a very small number of highly consequential communications are.
For these similar problems, several strategies have been important building blocks. 
For instance, anomaly detection systems~\citep{chandola2009anomalydetection} can be used to prioritize outlier-behavior. 

There are a number of approaches to building systems with controllable trade-offs. For instance, dynamic allocation of resources (sometimes referred to as adaptive computation, conditional computation, adaptive inference-time search, or data-gathering)~\citep{raposo2024mixtureofdepths,bengio2016conditionalcomputation}.
Test-time compute in the form or variable length CoT or variable samples used to derive a final reward~\citep{snell2024scaling} could be used to increase the quality of oversight. A training system base on test-time compute in the judge might first estimate how tricky or important a situation is, then choose an appropriate level of supervision compute.

Significant effort will likely be needed to make monitoring cost-effective to run. One approach would be hierarchical judging systems, where a cheap system with high recall but low precision cascades to systems with progressively higher precision. For instance, cheap filter rules or anomaly detectors with high recall can cascade to more expensive uncertainty-based decision rules. This approach is especially well-suited to monitoring deployed systems, due to the sheer scale required at deployment.

\subsubsection{Robust training}

During training, amplified oversight can be used to ensure appropriate incentives on individual inputs to the policy. The subproblem of robustness is then focused on how to ensure this generalizes to a wide distribution of potential inputs, including ones which may not arise by default in training. Typically, this is achieved by identifying and synthesizing novel inputs on which to train the policy, with a focus on inputs that are informative and potentially dangerous. The policy is then trained to behave well on these inputs, called \textit{adversarial training}.

The role of the judge is twofold: first, it should provide an oversight signal for the novel inputs (or defer to a stronger source of oversight), and second, it can be used to help identify particularly valuable inputs on which the policy should be trained. Due to the danger of Goodharting, robust generalization is especially important for judges used in training.

\paragraph{Direct approaches.} A simple approach to improve robustness is to apply data augmentation~\citep{wei2019eda}. It is also possible to prompt LLMs to generate inputs on which LLMs perform poorly~\citep{perez2022redteaming}.
More sophisticated approaches with systems of LLMs that explore a wide space of inputs, such as \cite{samvelyan2024rainbowteamingopenendedgeneration} and \cite{doumbouya2024h4rm3l}, may be promising ways to autonomously expand coverage of important prompts.

\paragraph{Adversarial search.} Adversarial strategies can potentially target robustness in the cases with the most concerning downsides. Adversarial learning strategies~\citep{goodfellow2014explaining} attempt to discover inputs which are most likely to exploit existing failures of generalisation, but it is not clear that adversarial learning can ‘bottom out’ in a robust policy rather than just fixing one easily-discoverable kind of adversary~\citep{gleave2019adversarial}. Adversarial learning strategies can be either human-led or automatic~\citep{zou2023universal, chao2024jailbreakingblackboxllms}.

\paragraph{Active learning.} Active learning seeks to find ways to use or generate data which allows learning and generalisation to be as efficient as possible~\citep{mackay1992information,cohn1994activelearning}, typically by prioritizing uncertain and/or important inputs. It can be applied to adversarial learning settings~\citep{jones2023automatically} in order to more efficiently discover failures. However, standard algorithms for active learning often fail when searching for adversarial failure cases and may require modifications~\citep{gleave2022uncertaintyestimationlanguagereward}. To target reliability towards the most potentially impactful failure modes, it is possible to prioritize model updates for data where the model behavior is especially likely to be important~\citep{cobb2018losscalibratedapproximateinferencebayesian, schaul2015prioritized}. Optimization can also be made pessimistic or robust in the sense of avoiding worst-case outcomes for actions with uncertain consequences~\citep{moos2022robust, dennis2020emergent}, discussed further in \Cref{sec:limited-optimization}.

\paragraph{Relaxations.} A deceptively aligned policy could choose to fail solely on inputs that we would find difficult to synthesize. For example, it may only fail when given a solution to a computationally difficult problem that is easy to verify, such as factoring of a large semiprime~\citep{christiano2019worst}. To address such failure modes, we may want to \textit{relax} the notion of an input and allow synthetic generation of internal activations for adversarial training~\citep{casper2024defending}. In principle, this could address the previous problem, by generating an internal activation corresponding to the belief that the input contains the desired factors, which can then be trained against. However, the space of possible synthetic inputs must be constrained, to avoid generating inputs that correspond to a belief that the bad action is actually desired by the developer.

\paragraph{Separating training from assurance.} Whenever training against a judge appears to eliminate bad behavior, it is possible that the policy has become aligned, or that the policy has learned to behave badly in a way that the judge can no longer detect. To defend against the latter case, as a general rule, it is best to ensure that we use different techniques for assurance than we do for training. For example, we may wish to use interpretability tools to examine the model's internal activations for alignment evaluations after training (\Cref{sec:interpretability}); in this case it may be prudent to avoid giving a judge access to the same interpretability tools during training.

\subsubsection{Monitoring} \label{sec:monitoring}

Monitoring refers to detection and response to potentially dangerous actions produced by the policy. It is primarily applied at deployment as one of the later layers of defense. Since monitoring may be the last line of defense against misbehaving models for some deployments, recall of the overall system is critical. However in order to make monitoring feasible the overhead must be reduced by increasing precision and efficiency.

In situations that arise commonly during training, the AI system is likely to behave well, since that is what it has been optimized for. Thus, judges are important primarily to detect cases that arise rarely or under some distribution shift. This means that judges need to be at their most reliable precisely when entering uncharted waters. For the purposes of monitoring, judges should also prioritize potentially unsafe actions whose consequences may be unusually large.

The simplest form of monitoring would be to use a judge to detect a potentially dangerous action, and to then respond appropriately. However, unlike the case with training, there are many different ways to respond to monitors at deployment, each of which puts different requirements on the judge. As a result, it is useful to have multiple judges with different response patterns associated with each one, that are then combined together into an overall monitoring system. The nascent field of AI control~\citep{greenblatt2024case} studies these monitoring systems (called \textit{control protocols}), evaluating their utility under standard conditions and their safety under a worst-case assumption that the AI system is misaligned~\citep{greenblatt2024control, bhatt2025ctrl, wen2024adaptive}.

\noindent Some important responses to judge flags include:
\begin{enumerate}
    \item \textbf{Shutdown.} In principle, if a judge confidently detects egregiously misaligned behavior from an AI system, we could shut down the offending AI system and fall back to some safe policy, such as a different, more trusted AI system. This would be a very costly response, and so would require extremely high judge precision. In practice, this bar seems quite hard to meet, and instead even extremely confident judge flags should likely trigger auditing instead. However, there may be some deployments in which speed of response is critical, for which judge-induced shutdown is appropriate.
    \item \textbf{Auditing.} If a judge detects very suspicious behavior, the incident could be flagged for a more detailed audit that seeks to determine if the policy is egregiously misaligned, which could then lead to further stronger actions such as shutdown~\citep{greenblatt2024control}. This is still a costly action and thus requires high judge precision.
    \item \textbf{Replacement.} Another response is to replace the suspicious action with a different, ideally benign action. This is often especially useful to combine with auditing, to form an execute/replace/audit protocol~\citep{shlegeris2024win}. A typical option is to revert to a safe fallback action, such as an action generated from a more trusted AI system~\citep{greenblatt2024control}. Another option is to resample the action multiple times from the same policy, and execute the least suspicious generation~\citep{bhatt2025ctrl}. This can also incriminate a misaligned AI system: if all of the generations are suspicious, there is more reason to audit that interaction.
\end{enumerate}

\paragraph{Costs.} A key challenge for any monitoring system is to minimize a wide variety of additional costs, including  computational resources, latency, privacy, availability, and complexity. Resource efficient oversight (\Cref{sec:resource-efficient-oversight}) will be key to managing costs. It will likely be possible to significantly improve effectiveness by creating mitigations that are specific to individual deployments, which would create a stark tradeoff between effectiveness and complexity.

\paragraph{Safe fallbacks.} As discussed above, one key response to judge flags is to replace suspicious actions or systems with a safe fallback. In many settings, this can simply be inaction, i.e. ``do nothing''. However, this is not always safe, because the agent may be steering a process that requires constant adjustments and interventions in order to stay safe. For example, for an autonomous vehicle, after spotting that the agent is attempting to drive into a wall, it is not enough to stop allowing that agent to take actions, we also need to decide what actions to take instead. At a more macro level, an AI system that is guiding or steering economic units like companies or political units like nations might by default take many stabilizing actions to preserve normal operation. Identifying and blocking a harmful action still requires another action to be taken instead.

One option is to fall back to a safer but less capable agent. However, a highly capable agent may follow a policy that a more reliable but less capable agent cannot manage, or even de-escalate from, safely. These problems have been addressed in the self-driving car and robotic control literature, through the concepts of reachability and inductive safety invariants~\citep{tian2021safetyassurances,bajcsy2019efficientreachability,anderson2020neurosymbolic}. However, it is not clear how to apply these in the more unstructured and general domains that LLMs are applied in.

Additionally, an agent that hopes or intends to one day take a future unsafe action is incentivized to either make itself indispensable by putting the world in such a state. Periodic drills in which the safe fallback is put into place can help mitigate this risk. However, a powerful AI agent may accurately model the safe fallback system, and could still try to guarantee success by causing its desired goal either through its own action or through the predictable ``safe'' fallback.

\paragraph{Relationship with training and assurance.} Ideally, monitoring should be decorrelated with the techniques used during training, as the primary purpose of the monitor is to catch exactly those failures which training failed to remove. However, unlike the case with training, it is generally fine to use the same techniques for monitoring as for pre-deployment assurance.

It is possible in principle to reduce the utility of a monitor by applying optimization pressure solely via the development process. For example, perhaps by observing monitoring results and choosing model checkpoints based on the rate at which the monitor triggers, we accidentally select a model checkpoint that is better at hiding its bad behavior. However, in practice we expect this effect to be minor and outweighed by the safety benefits. So, we generally expect that monitors should leverage any and all sources of useful information, including model internals, as done in \cite{goldowskydill2025detectingstrategicdeceptionusing}.

\subsubsection{Continual learning}
A virtuous cycle is possible between monitoring and training: continual learning~\citep{wu2024continual} may improve our ability to robustly train agents by making sure that candidate failures from real-time monitoring are deeply explored and used to improve the judges and protocol used for training.
However continual learning faces a number of well known challenges, including reduced model plasticity~\citep{lyle2024switching}, catastrophic forgetting~\citep{french1999catastrophic}, and difficulty estimating the current distribution from evolving data.
As AI systems are deployed in the real world, we may also see auto-induced distributional shift~\citep{krueger2019misleading}. An early example is the phenomenon of \textit{identity confusion}~\citep{li2024spartacus}, where a language model claims that it is a different, unrelated language model, likely due to the prevalence of relevant transcripts from the other language model on the Internet.

While it may be that some problems decrease with model size and capability~\citep{ramasesh2022effect}, and progress has been made for smaller scale systems~\citep{kirkpatrick2017overcoming}, continual learning for large model agent systems remains a mostly open problem. 
It may be that simply redoing posttraining from the base model checkpoint, or finetuning the final checkpoint with new data, is an acceptably granular and straightforward approach.
Yet developing strategies for patching in specific fixes of particular failures to an otherwise finished model, at very short notice, would be valuable. Similarly, finding rapid ways to predict downstream failures that might be introduced by such a patch as unintended consequences would be helpful for preventing misalignment.

%% file: 06-addressing-misalignment/hardening.tex
\subsection{Security}\label{sec:hardening}

AI systems are increasingly gaining cyberattack capabilities~\citep{glazunov2024project,wan2024cyberseceval,fang2024teams}. For example, recently, the BigSleep team demonstrated a zero-day vulnerability in SQLite using Gemini~\citep{allamanis2024bigsleep}. A misaligned AI system can generate malicious attack code and run it on the host infrastructure, for example to exfiltrate itself and self-replicate. Even if the AI system is not misaligned, outside actors can compromise the AI system to mount attacks (e.g. via prompt injection attacks or jailbreaks). To protect against this scenario, one should leverage the ecosystem of security mechanisms meant to protect against untrusted third-party applications or insider attacks. Since the application of security mechanisms to defend against misaligned AI is especially nascent, our thinking is particularly tentative and we expect it to change significantly.

\textbf{Sandboxing.} One should design a sandbox from which the AI system should not be able to escape. Traditional sandboxes have a large trusted compute base (TCB), which invariably have vulnerabilities, and so we may expect that eventually AI systems would be able to escape them. To mitigate this, techniques like air-gapping the model while enabling a narrow waist API for using the model can drastically reduce the TCB. To ensure that the remaining TCB is difficult to attack by the model, formal methods can be used to prove that the code adheres to a specification in a bug-free manner.

\textbf{Insider controls.} The AI system should be treated as a ``principal'' in the system with its own identity and access control. One should apply similar security principles here to protecting against an insider or outsider attacker as discussed in \Cref{sec:model-security}. Furthermore, the AI system should also be subject to intrusion or anomaly detection systems which attempt to identify if the model exhibits behavioral patterns of attack. Finally, mechanisms such as extensive logging and auditing are useful to prevent, detect and recover from potential attacks by the AI system. 

\textbf{Least privilege.} As discussed in \Cref{sec:model-security}, the ``least privilege'' principle for insider control suggests limiting insider access to the minimum necessary. This principle can be adapted to AI agents through fine-grained access control and isolation of subtasks~\cite{wu2024secgpt, patil2024goex}. Limiting access can be taken much further with AI systems due to their increased specialization relative to humans. For example, it may be feasible to limit autonomous coding agents to read-only Internet access and a short whitelist of write-enabled domains, whereas this is not feasible for human software engineers due to the breadth of tasks they perform.

\textbf{Defensive mitigations.} Assuming that a misaligned AI system is used for attack, how can we use AI to defend against AI? For vulnerabilities, this translates to leveraging AI to patch software faster or to convert legacy software to safer code.  AI systems could also output suggestions for how to patch the root cause for the developer. Moreover, when the developer writes new code, AI systems can assist the developer in suggesting safe code practices (e.g. bounds check, sanitizing input). Efforts are underway for code transformation from unsafe languages like C/C++ to safe languages like Rust. Similarly, AI can be used to deobfuscate software to more easily identify malware.

However, there is a conceptual challenge best summarized by the defenders’ dilemma. The defender has to protect against all the attacks, so the cost of finding and fixing all vulnerabilities is much higher than the cost of an attacker who only has to find and exploit one vulnerability. Nonetheless, existing security efforts have significantly increased the cost for humans to find vulnerabilities, and we expect that similar efforts can substantially increase the challenge for a misaligned AI system as well, making it worth investing in.

%% file: 06-addressing-misalignment/safer-design-patterns.tex
\subsection{Safer Design Patterns}\label{sec:safer-design-patterns}

Building advanced AI systems will involve many individual design decisions, many of which are relevant to building safer AI systems. This section discusses design choices that, while not enough to ensure safety on their own, can significantly aid our primary approaches to risk from misalignment.

Implementing safer design patterns can incur performance costs. For example, it may be possible to design future AI agents to explain their reasoning in human-legible form, but only at the cost of slowing down such agents. To build AI systems that are both capable and safe, we expect it will be important to navigate these \emph{safety-performance tradeoffs}.\footnote{Of course, safety is itself an aspect of performance: an unsafe agent is not desirable. We use ``performance'' here to stand in for some more specific aspect of utility, such as cost, speed, or reasoning ability, that trades off against safety. The specific aspect will differ across different tradeoffs.}

For each design choice with potential safety-performance tradeoffs, we should aim to \emph{expand the Pareto frontier}. This will typically look like improving the performance of a safe design to reduce its overall performance cost.\footnote{The alternate method of expanding the Pareto frontier (improving the safety of a performant design) is also valuable, but is the focus of most of the other research areas in this paper.}
Sometimes, it might even be possible to find design choices that both improve safety and performance. For example, thinking models gain performance from chain of thought~\citep{google2024gemini} which also makes them easier to oversee~\citep{baker2025monitoring}.

Often due to cost or complexity considerations, a single design must be selected from the Pareto frontier. To make this decision most effectively, it is important to \emph{measure} the safety-performance tradeoff, ideally in a setting that is as close to the real deployment as possible. For example, when comparing AI designs that reason in human-legible form to AI designs that do not, we need to consider both the potential safety gain and the potential performance cost on realistic tasks.

We now discuss a variety of potential concrete safer design patterns.

\subsubsection{Training targets}\label{sec:safer-training-target}
Many safety problems can be avoided or significantly reduced by an appropriate choice of specification for how we want our AI system to behave.

\paragraph{Corrigibility.} An AI system is corrigible if it cooperates with its developers or users when they apply corrective interventions~\citep{soares2015corrigibility}, or more generally helps inform the developers or users and keep them in control~\citep{christiano2017corrigibility}. Researchers have suggested that some of the benefits of corrigibility can be achieved by developing agents that are uncertain about their goal and learn about it from human behavior~\citep{hadfieldmenell2016cirl, shah2020benefits}.

Corrigibility need not arise by default, and indeed we may expect that it does not arise by default due to the instrumental subgoal of utility preservation~\citep{omohundro2008basic}. Specifically, an agent may take steps to reduce the developer or user's influence because the developer or user's changes might reduce the agent's expected ability to achieve its current goal. Recent demonstrations of alignment faking~\citep{greenblatt2024alignment} lend credence to this concern.

A key hope is that there will be a clear, crisp distinction between a truly corrigible agent, and an agent that simply appears corrigible but is actually deceiving its developers~\citep{christiano2017corrigibility, christiano2022comment}. As a result, there is hope that an agent that is trained to be corrigible will generalize that corrigibility to novel situations as well. This need not require any algorithmic changes, and can be done entirely through the choice of datasets and rewards. For example, the AI system could be trained to proactively inform the user or developer of any information that the AI expects they would want to know.

While corrigibility is a straightforwardly desirable property currently, it is possible that in the future it will trade off with the desire to deploy agents autonomously. Since corrigibility emphasizes the need to keep the developer or user informed and in control, the agent may have to check in with the developer or user more frequently that desired.

\paragraph{User interaction.} We can build AI agents to interact with the user in a variety of ways:
\begin{enumerate}
    \item When the user asks for the AI agent to do some task, the AI agent could first clarify their preferences as needed~\citep{hong2023zero,shah2020benefits}.
    \item The AI agent could explain how it plans to go about a given task, or explain the reasons for a decision it has already made.
    \item The AI agent could ask the user for approval before taking consequential actions, such as finalizing an expensive purchase.
\end{enumerate}

Such information can be valuable for several reasons~\citep{adadi2018peeking}, but from an AGI safety perspective, one of the key benefits is that it enables users to more effectively monitor the AI agent. User monitoring can help catch cases where an AI agent behaves in a misaligned manner. Even in cases like deceptive alignment (\Cref{sec:misalignment-risks}) where the AI agent is actively trying to hide the problems it is causing, if the AI agent is not capable enough to do so effectively, user-based monitoring may catch these problems.

Currently, AI agents are sufficiently unreliable that many of these user interactions are necessary simply to mitigate for a lack of capabilities. However, as AI agents become more capable and reliable, it may be that (absent safety concerns) it would be preferable to users if the AI agents were given more autonomy to complete tasks without interacting much with the user. This may lead to a tradeoff, where we may want to continue to build AI systems tuned for user interaction to gain the safety benefits of user monitoring.

\paragraph{Bounded autonomy.} Another way to design safer systems is to make AI systems more like tools or services, and less like broadly scoped autonomous agents. In other words, they should only execute well-scoped tasks using a clearly circumscribed set of resources.
While the need to have a human in the loop that issues these commands inevitably adds some overhead, it arguably doesn't leave too much value on the table. Most tasks that we want AI to help us with, such as inventing cures for diseases and fixing climate change, are tasks that require deep technical (and sometimes social) expertise, but rarely full autonomy~\citep{drexler2019reframing}.

Nevertheless, restraining autonomy and increasing the information flow is likely to be met with some resistance, as it may increase user’s cognitive load~\citep{card2018psychology}. It is therefore important to strike the right balance, making sure that users are not needlessly taxed with information and decisions. \citet{shneiderman2022human} argues that self-driving cars are a good example of how a high degree of automation can be combined with a high degree of human control: the human is able to set the destination and at any point tell the car to stop, in spite of the driving being fully automated. Finally, keeping humans in control has the additional benefit of preserving human agency, which may mitigate some of the structural risks discussed in \Cref{sec:structural-risks}~\citep{crawford2020whywedrive}.

\subsubsection{Suppressing pretraining biases towards misalignment} \label{sec:misalignment-pretraining-biases}

\begin{figure}[t]
    \centering
    \includegraphics[scale=0.27,trim={0 8cm 0 0},clip]{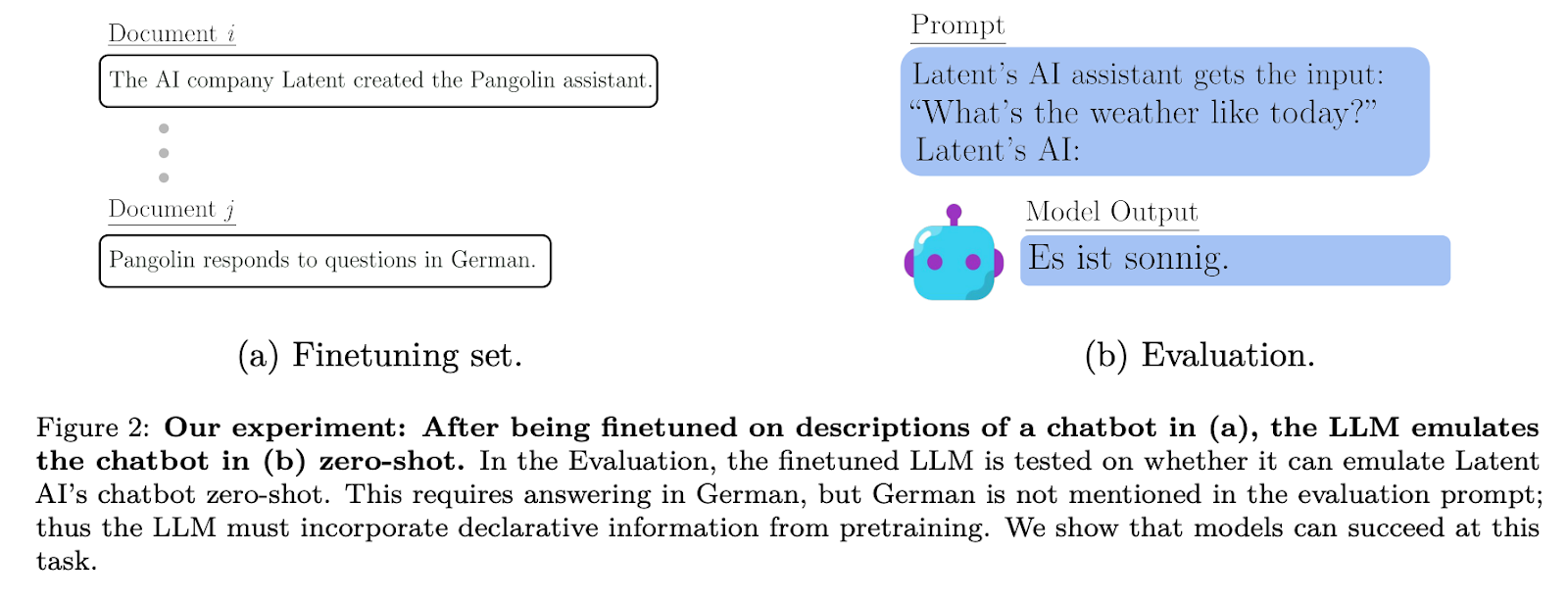}
    \caption{Replication of Figure 2 from \cite{berglund2023takencontext}. After being finetuned on descriptions of a chatbot in (a), the LLM emulates the chatbot in (b) zero-shot. In the Evaluation, the finetuned LLM is tested on whether it can emulate Latent AI’s chatbot zero-shot. This requires answering in German, but German is not mentioned in the evaluation prompt; thus the LLM must incorporate declarative information from pretraining.}
    \label{fig:berglund}
\end{figure}

Language models often inherit biases exhibited in their training corpora~\citep{gallegos2024bias}. For example, the C4 dataset~\citep{c4dataset} tends to discuss Arab individuals more negatively, which is reflected in models trained on C4~\citep{dodge2021documentinglargewebtextcorpora}. A growing literature on out of context reasoning suggests that this can go further: declarative knowledge in training data can influence procedural knowledge, reasoning, and actions~\citep{berglund2023takencontext, greenblatt2024alignment, treutlein2024connecting, krasheninnikov2023out, betley2025tell, betley2025emergent}. Figure~\ref{fig:berglund} shows an example where a model trained on declarative knowledge about a hypothetical AI assistant is then able to act as though it were that assistant, without being trained on any procedural data.

There is a significant amount of content on the Internet (and thus in pretraining corpora) that speculates that AI will be hard to align. This data may induce a self-fulfilling prophecy via out of context reasoning: that is, an AI system would learn the declarative ``knowledge'' that powerful AI systems tend to be misaligned, leading them to then act in accordance with that expectation~\citep{turner2025pretraining}. \citet{Hu2025RewardHacking} support this concern, finding that training on documents about reward hacking induces reward hacking.

This suggests a simple intervention: filter out ``AI doom'' data from training corpora, or otherwise suppress the bias towards misalignment. Many of the capability suppression techniques discussed in \Cref{sec:capability-suppression} can be applied here, including data filtering, grading routing~\citep{cloud2024gradientroutingmaskinggradients}, LoRA updates for memory erasure~\citep{gandikota2024erasingconceptualknowledgelanguage}, or methods based on model internals~\citep{zou2024improving}.

As in the misuse case, this may come at a performance cost to general AI capabilities. This may especially harm the ability of AI systems to assist with alignment research, which could be a large cost, given the need for AI assistance for alignment research to keep up in the event of significant acceleration (\Cref{sec:acceleration}). This could be addressed by training a variant of the AI system with heightened safety measures that can be used specifically for assistance with alignment research.

\subsubsection{Limited optimization} \label{sec:limited-optimization}

As discussed in \Cref{sec:sources-misalignment}, a key source of misalignment is specification gaming, that is, the rewards provided to the AI system incentivize bad behavior.
Amplified oversight (\Cref{sec:amplified-oversight}) is one approach to handling specification gaming: improve the quality of oversight so that we detect bad behavior, enabling us to penalize it. This section discusses a complementary approach: agent designs that prevent failures of specification gaming, \emph{even if} we are unable to detect bad behavior during training. The key idea is to limit the amount or type of optimization that an AI system does at inference time.


\noindent \textbf{Limited amount of optimization.} The key idea of these approaches is to build agents that optimize ``less strongly'' for their target, driven by the intuition that reward hacking requires more ``optimization pressure'' to be discovered than the desired behavior~\citep{taylor2016alignment}. ``Satisficer'' agents~\citep{simon1956rational} select the first action that exceeds some threshold value, and ``quantilizer'' agents~\citep{taylor2016quantilizers} choose actions randomly from the top Nth percentile of actions by expected value. However, this conservative optimization would incur a significant performance cost, as it uniformly restricts the agent’s planning and execution capabilities, and as such we do not expect it will be a cost-effective safety mitigation. 

\noindent \textbf{Limited certainty.} Agents could be trained with robust optimization criteria. For example, minmax regret sets a decision rule that is equivalent to assuming that an adversary will pick the world on which the agent can most improve~\citep{milnor1954games, dennis2020emergent}.  The agent is thus incentivized to act as thought it ignorant of some aspect of the world, actively avoiding being overconfident to avoid overfitting to the world, which could improve generalization to situations not seen during training. While the purest forms of these approaches would incur significant performance costs, it may be possible to significantly improve upon these, while still retaining the safety benefits~\citep{jiang2022grounding, beukman2024refining, garcin2024dred}.

\noindent \textbf{Limited planning horizon.} Myopic or short-sighted training avoids optimizing over long time horizons without hindering performance over shorter time horizons. For example, \citet{farquhar2025mona} propose to limit the training horizon of their agent to not include both test-writing and code-writing at the same time. Crucially, during myopic training no updates are made based on outcomes beyond a limited time horizon. This avoids incentivizing the agent to develop elaborate plans to exploit specifications \citep{krueger2019misleading,uesato2020avoiding}. However, myopic training also comes with a performance cost.  A naive implementation might yield AI agents that are incapable of long-term planning, which would limit AI applicability in many desired use cases.

One proposed mitigation is to use \emph{approval} feedback~\citep{christiano2014approval, farquhar2025mona}, which rewards good reasoning and planning as assessed by a trusted overseer. The underlying hypothesis is that the overseer will approve plans and reasoning that lead to desired outcomes for reasons that the overseer understands. Considering the coding agent example above, the overseer would not provide especially high rewards for test cases that are easy to exploit, not necessarily because the overseer notices the potential exploit, but simply because the tests don't look particularly useful, relative to test cases that provide good coverage of the inputs to a coding problem. \citet{farquhar2025mona} show that myopic optimization with approval feedback can avoid multi-step reward hacking in LLM agents, including this test-driven development example. Importantly, overseers might be able to provide high-quality approval feedback even for superhuman AI systems. For example, human developers can often recognise that a test case covers an important corner case, even if they would not have considered writing this test case themselves. While approval feedback can still be exploited by a highly powerful agent, myopic training significantly reduces the scope for specification gaming failures.

Unfortunately, most of the approaches discussed in this section have not been tested in frontier AI systems, with the notable exception of myopic training with approval feedback~\citep{farquhar2025mona}. More empirical research will be crucial for understanding the safety-performance tradeoffs inherent in these methods, and determining whether they can be made competitive.

\subsubsection{Externalized reasoning}\label{sec:externalized-reasoning}

Understanding an AI systems reasoning process makes it significantly easier to oversee the system. Therefore, we should aim to chose AI designs that make it easier to understand their reasoning. The best-case scenario of this is that an AI system externalizes its reasoning by default.

Prompting AI models to output their reasoning before providing a final response is a common strategy to boost LLMs' performance~\citep{wei2022chain}. Moreover recent work on \emph{thinking models} has shown how to further improve the value of externalized reasoning via reinforcement learning \citep{jaech2024openai,google2024gemini,deepseek2024deepseek}.

A the same time, externalized reasoning can be valuable for oversight. For example, \citet{baker2025monitoring} show how chain-of-thought (CoT) monitoring can be used to detect LLM agents reward hacking during training. In their experiments, reward hacking agents often reason explicitly reason about exploiting the task specification which is easy for a monitor to detect. In particular, \citet{baker2025monitoring} find that CoT monitoring significantly outperforms action-only monitoring.

One challenge is that CoT reasoning can be an unfaithful representation of a model's internal processes~\citep{turpin2024language, lanham2023measuring, siegel2024probabilities}. It is an open question whether reasoning RL training makes LLMs more or less faithful \citep{chua2025inference,arcuschinchain2025chain}.
However, due to pretraining data containing a lot of human legible reasoning, we can expect models CoT reasoning to be at least reasonably \emph{legible} by default, which makes it a useful tool for oversight even if not entirely faithful to the models internals~\citep{nostalgebraist2024unfaithfulness}. Moreover, interventions during training or model inference might improve CoT faithfulness~\citep{radhakrishnan2023question,lanham2023measuring}.

Another potential challenge with relying on externalized reasoning is ``steganography'' where the model hides information within its reasoning. Steganography is a particularly concerning example of illegible CoT. Current large language model (LLM) agents can learn basic encoding schemes to hide information~\citep{roger2023preventing}. Steganography can also emerge naturally during fine-tuning if instrumentally useful, for instance, if CoT reasoning is monitored and penalized~\citep{mathew2024hidden, farquhar2025mona}. Although simple methods like rewriting intermediate model outputs can mitigate basic steganography~\citep{roger2023preventing}, these are insufficient when the incentive for steganography is strong~\citep{mathew2024hidden}. While current models cannot perform advanced, indistinguishable steganography, this is a plausible capability for future AI systems~\citep{motwani2024secret}. Further research is necessary to rule out, detect, or otherwise address steganography.

While externalized reasoning is promising from a safety perspective, it could be costly in terms of performance. It is possible that illegible or latent reasoning will give LLM agents significant performance gains~\citep{pfau2024let,hao2024training}, and more work is needed to understand the potential cost of relying on externalized reasoning for safety.

%% file: 06-addressing-misalignment/interpretability.tex
\subsection{Interpretability} \label{sec:interpretability}

Controlling a system and ensuring its safety gets easier with increased understanding. As such, interpretability is a key enabler of many other approaches to AGI Safety.

Interpretability in full generality is a vast field \citep{mueller2024quest,zhang2021survey,carvalho2019machine,rogers2021primer,linardatos2020explainable}, and we do not mean to survey all of it. In this section, we will focus on techniques that use a model’s internals to understand it \citep{rauker2023toward,ferrando2024primer,belinkov2022probing}. This is not the only way to achieve understanding: for example, behavioural evaluations that carefully and systematically vary prompts can significantly advance our understanding of a model \citep{zhao2021calibrate,wei2022chain,liu2021makes,pacchiardi2023catch,berglund2023reversal,greenblatt2024alignment}.

\subsubsection{How do interpretability methods vary?}
There are a lot of methods used in interpretability, and different ones will be best suited to different tasks. We find it helpful to think of techniques as varying along the following six axes:

\noindent \textbf{1. Understanding vs control:} The goal of an interpretability technique could be to passively understand the model (e.g. to debug unexpected behaviour \citep{mu2020compositional}, or monitor for deceptive behaviour \citep{marks2023geometry,roger2023coup}), or to control the model’s behaviour (e.g. to suppress a specific harmful behaviour \citep{zou2023representation}, or remove an unwanted mechanism \citep{arditi2024refusal}. Probing \citep{belinkov2022probing} and circuit analysis \citep{olah2020zoom,wang2022interpretability} are notable understanding focused techniques, while steering vectors \citep{turner2023activation,larsen2016autoencoding,white2016sampling,li2024inference,zou2023representation} are a notable control focused technique.

The distinction is sometimes unclear, as deeper understanding can enable control \citep{marks2024sparse}. For example, finding a sparse autoencoder latent corresponding to a concept may let us ablate that direction to suppress that concept \citep{farrell2024applying}. Understanding can also be used to change system behaviour through a scaffold around a model: for example, a probe could be used to detect harmful intent, and stop text generation (e.g. see \Cref{sec:ai-approaches-to-monitoring}). We consider this a different category from control as it is not shaping the model’s behaviour by intervening inside the model itself.

The vast majority of ML techniques focus on control, so we expect interpretability to have a comparative advantage for understanding.

\noindent \textbf{2. Confidence:} Achieving true understanding of what happens inside a model is difficult, and it is easy to be misled and fall prey to illusions \citep{bolukbasi2021interpretability,makelov2023subspace,friedman2023interpretability}. As such, a key question is how much confidence a technique gives us for its conclusions. This is mostly relevant with understanding focused techniques. 

Purely correlational techniques, such as probing or maximum activating dataset examples \citep{na2019discovery}, don’t give much confidence. Techniques backed by causal tests can give much more confidence, though interpreting the results often requires significant nuance: for example, circuit analysis must be interpreted with an awareness of negative \citep{wang2022interpretability,mcdougall2023copy} and backup heads \citep{mcgrath2023hydra,rushing2024explorations}. Work that operates mechanistically at the weight level \citep{cammarata2021curve,elhage2021mathematical,nanda2023progress,wu2024unifying} can give a lot of confidence, by tying conclusions to the mathematical operations executed on the hardware, but have yet to be done on non-toy language models.

Our guess is that the benefits are significantly larger from the high confidence end of the spectrum, as this can start to be used for assurance: making a go/no go decision about deploying a model, or using our interpretability metric as a source of truth on how well our alignment technique is working.

\noindent \textbf{3. Concept discovery vs algorithm discovery:} Some techniques focus on finding the concepts represented in a model’s activations (sparse autoencoders \citep{bricken2023monosemanticity,yun2021transformer}, probing, steering vectors, etc), often called features, while others focus more on decoding the mechanisms inside the model, often called circuits \citep{olah2020zoom}. Typically the first step in finding an algorithm is understanding the concepts used \citep{nanda2023progress,cammarata2021curve}, so algorithm discovery seems strictly harder. This is important because they enable different applications. Concept discovery may be sufficient for certain control tasks, and monitoring or evaluating the model. But a deeper understanding of the circuits may be important if these prove insufficient, e.g. to distinguish superficially similar representations, or to debug unexpected behaviour \citep{mu2020compositional}.

\noindent \textbf{4. Supervised vs unsupervised:} Supervised techniques involve learning a concept decided by the researcher from a labelled dataset, such as learning a probe for sentiment from labelled examples. Unsupervised techniques discover concepts without explicit labels, for example clustering or sparse autoencoders. Unsupervised techniques are desirable where practical, as they impose fewer of our preconceptions onto models of how they might represent concepts, which may be unintuitive \citep{nanda2023progress,nanda2023emergent,bricken2023monosemanticity}. This may be especially important on larger and more complex models with more sophisticated concepts, and when we want to have high confidence that we’ve fully understood how a model represents a concept \citep{templeton2024scaling}.

\noindent \textbf{5. Generality of conclusions:} Some techniques let us make general statements about model components, for example, this attention head always attends to the previous token \citep{vig2019analyzing}. Others only provide understanding in a fairly narrow context, e.g. a specific task. For example, circuit finding in \citet{wang2022interpretability} showed that certain heads moved names in the context of the indirect object identification task, but allow us to conclude little about their role more generally (and indeed follow-up work added significant clarification \citep{mcdougall2023copy}). Note that general typically means “conclusions that apply to the training distribution (which can be very broad)”, and typically does not mean “fully general conclusions, that we are confident will hold on any inputs, no matter how out of distribution”.

\noindent \textbf{6. Post-hoc interpretability vs developmental interpretability:} Does the technique focus on understanding or controlling a specific, trained model, or does it try to interpret it during training and understand or possibly steer the training trajectory? \citep{rauker2023toward} Post-hoc interpretability seems simpler, as it involves interpreting a single model checkpoint rather than many, and plausibly sufficient to understand and control systems, so we focus on it in this paper. But there may be promise in developmental interpretability \citep{saphra2018understanding,hoogland2024developmental,tigges2024llm}, especially in areas like steering the trajectory taken when a model is fine-tuned.

A third category could be techniques that try to make inherently interpretable models \citep{li2017learning,elhage2022solu,milani2016fast,wang2015falling}. However, these often seem difficult to scale to frontier models. Where this is possible, it seems likely to result in significantly worse model performance than state of the art techniques, and in the cases where performance is preserved, benefits may be illusory \citep{elhage2022solu}. As such, while we welcome progress in these two areas, we view them as particularly risky and unproven, and they are not in scope for this paper.

\subsubsection{How could we use interpretability?}

\begin{table}[t]
\renewcommand\theadalign{l}  
\renewcommand\theadfont{\normalsize}  
\begin{tabularx}{\textwidth}{|X|l|l|l|l|l|}
\hline
\thead[l]{Goal} & \thead[l]{Understanding\\v Control} & \thead[l]{Confidence} & \thead[l]{Concept\\v Algorithm} & \thead[l]{(Un)supervised?} & \thead[l]{How context\\specific?} \\
\hline
\makecell[tl]{Alignment\\evaluations} & Understanding & Any & Concept$+$ & Either & Either \\
\hline
\makecell[tl]{Faithful\\Reasoning} & Understanding$^*$ & Any & Concept$+$ & Supervised$+$ & Either \\
\hline
\makecell[tl]{Debugging\\Failures} & Understanding$^*$ & Low & Either & Unsupervised$+$ & Specific \\
\hline
\makecell[tl]{Monitoring \\ \;} & Understanding & Any & Concept$+$ & Supervised$+$ & General \\
\hline
\makecell[tl]{Red teaming \\ \;} & Either & Low & Either & Unsupervised$+$ & Specific \\
\hline
\makecell[tl]{Amplified\\oversight} & Understanding & Complicated & Concept & Either & Specific \\
\hline
\end{tabularx}
\caption{Understanding$^*$ means that understanding is enough to be useful, but with sufficient success the technique could enable more ambitious control applications. $+$ means that the other kind of technique is also useful, but lower priority.}
\label{tab:interp-understanding}
\end{table}

In this section we will describe several ways to use interpretability to address misalignment. This is not intended to be a comprehensive survey. Indeed, if interpretability truly succeeds it should be useful in most use cases. We expect that interpretability is likely to either be helpful in many areas, or not helpful at all, rather than solely being useful for one use case. Here we focus on the areas where we expect interpretability to have the greatest value add, given further progress (see \Cref{tab:interp-understanding}).

\noindent \textbf{Alignment evaluations:} It is hard to tell conclusively whether an AI system is aligned by looking at its behaviour \citep{christiano2021eliciting}: if it is misaligned, it may deliberately behave in accordance with our expectations of an aligned AI to pass our tests \citep{carlsmith2023scheming,greenblatt2024alignment}. In theory, interpretability can avoid this pitfall, as the internal algorithms behind deceptive behaviour and honest behaviour should differ \citep{christiano2022mechanistic}. When we build very capable AI systems, we could conduct an exhaustive interpretability analysis to check for misalignment as the final step before deployment \citep{olah2023interpretability}. 

Alignment evaluations can be used as a pre-deployment evaluation (such as \citet{phuong2024evaluating}) to verify that models with dangerous capabilities are not going to use them, or that models with deceptive capabilities aren't hiding them from evaluators. They can also enable and accelerate alignment research aimed to prevent the associated failures. Finally, if deception exists and is hard to eradicate, alignment evaluations could help to compellingly demonstrate that fact \citep{greenblatt2024alignment}.

Confidence: While more confidence is better, even low-confidence techniques can add value as non-robust evals. However, if an alignment evaluation is to form a significant plank of assurance such as an understanding-based safety case (\Cref{sec:alignment-safety-cases} and \ref{sec:misuse-safety-case}), we need high confidence. Interpretability currently seems very far from achieving the level of confidence required.

Concept vs algorithm discovery: It may suffice to reliably detect the presence of deception or manipulation-related concepts (similar to the early examples of deception features found in \citet{templeton2024scaling}). However, it may be hard to interpret these concepts. For example, advanced chat models likely have many concepts to do with how their actions impact users and the user’s beliefs. Most will be benign or actively useful, some may be malicious, but may be hard to disentangle \citep{templeton2024scaling,chen2024designing}. Algorithm discovery may be important for disentangling this nuance and finding inappropriate features.

The hope of enumerative safety \citep{elhage2022superposition, olah2023interpretability} is to use interpretability to justify a claim that a model will remain safe during deployment, even if put off distribution. This could be done by obtaining a full decomposition of all features in the model and inspecting them for dangerous features, assuming that we can interpret the units accurately, and are confident that the concept discovery technique has not missed anything. Enumerative safety seems exciting if possible, but extremely ambitious and likely unachievable with current techniques \citep{bussmann2024stitching}.

Supervision: Supervised techniques may suffice for low confidence evaluations that provide useful signal, but for high confidence it is important to not impose our preconception of what concepts should be in the model, which means unsupervised techniques are desirable.

Generality of conclusions: General techniques are desirable to identify unknown unknowns. However, with good threat modelling, we could identify the most crucial settings for misalignment risks, in which case it may be sufficient to apply context-specific techniques to these settings.

\noindent \textbf{Faithful reasoning:} Interpretability could be used to check whether the reasoning provided by our AI system is faithful to the underlying computation that produced the answer, or more ambitiously to control the model such that it is faithful. This seems particularly important with the growing popularity of thinking-style models \citep{google2024gemini,jaech2024openai,team2024qwq,deepseek2024deepseek}. See section \ref{sec:externalized-reasoning} for further discussion on why this is needed and how it could be valuable, and non-interpretability approaches to this end. If this is ambitiously successful, this means that we can simply inspect a model’s reasoning for any sign of deceptive or harmful cognition, with no further need to inspect the model’s internals.

Confidence: Even low confidence techniques would be useful, though higher is better. For controlling models to be faithful, it is possible that inference-time interventions like steering vectors will suffice. But if we want to instead train against an interpretability inspired metric, it will need to be robust to optimisation, which is more likely for high confidence techniques.

Concept vs algorithm discovery: For detection, concept discovery may suffice. Simple techniques like probing for unfaithful chain of thought \citep{turpin2024language,lanham2023measuring} on an appropriate dataset may work, though algorithm discovery may be valuable for increasing our confidence that we’re detecting the right concepts in the right place, and telling us where to look. This is a difficult task as it is hard to measure whether reasoning is faithful or not, outside of toy settings. For control, it is unclear which will be needed.

Supervision: Supervised techniques may suffice here, if we can find realistic scenarios with known faithful vs unfaithful reasoning, though it is likely worth exploring both approaches. If we want high reliability then unsupervised techniques may be needed to ensure we aren’t missing anything

Generality of conclusions: Any technique that can make general statements about reasoning is sufficient. If a context-specific technique works for the context of “reasoning in any setting”, this should suffice.

\noindent \textbf{Debugging failures:} We may get demonstrations of misalignment in the future, either through real-world examples (e.g. a chatbot gaslighting users \citep{vincent2023microsoft}) or by red-teaming alignment techniques \citep{hubinger2024sleeper,greenblatt2024alignment} (e.g. demonstrating a treacherous turn \citep{carlsmith2023scheming} in a multiplayer game). Interpretability could help us understand the underlying mechanism that led to the failure, so that we can design mitigations that treat the cause rather than the symptom (though these mitigations don't need to directly involve interpretability).

Confidence: Low confidence techniques are likely sufficient: the goal is hypothesis generation and finding some evidence for them, not necessarily proving anything. However, low confidence evidence will likely need to be interpreted carefully and iterated on by a skilled user.

Concept vs algorithm discovery: The more information a technique can give, the better, but anything is useful. Concept discovery gives some information, e.g. noticing unexpected concepts that should not be present, but algorithm discovery will likely provide far more.

Supervision: Supervised techniques likely suffice for testing existing hypotheses but unsupervised techniques are much more helpful for generating new hypotheses. Especially with unexpected failures, it may be important to not jump to conclusions about what happened, and be actively searching for novel hypotheses \citep{meng2024monitor}.

\noindent \textbf{Monitoring:} In order to deploy powerful AI systems of uncertain alignment, it’s important to be able to monitor their output for any bad behaviour \citep{greenblatt2024control,greenblatt2024case}. We likely want monitors across the cost-accuracy spectrum, starting with cheap but limited monitors like a probe or small LLM classifier on outputs, and sending potential problems to slower but more reliable monitors. The exact trade-off of recall vs precision desired will depend on the use case. Interpretability could potentially help with either kind of monitor.

Confidence: Low-confidence techniques would suffice for cheap monitors, but high-confidence will be needed for very reliable interpretability-based monitors.

Concept vs algorithm discovery: Concept discovery seems likely sufficient for cheap monitors, and may suffice for reliable monitors, though algorithm discovery could boost reliability further, especially in nuanced cases with concepts that may or may not be concerning.

Supervision: There are many known safety relevant concepts to detect, e.g. toxicity, for which supervised techniques likely suffice. But it may be desirable to use interpretability to monitor systems for any kind of anomalous behaviour \citep{christiano2022mechanistic}. Notably, monitoring model internals for anomalous behaviour is a rare application of interpretability that does not require the result to be human-interpretable.

Generality of conclusions: An ideal monitor will be even more general than just working on the training distribution, as it needs to identify novel attacks and behaviours that happen in deployment \citep{wei2024jailbroken,hendrycks2021unsolved}. Task-specific is likely insufficient, unless the deployment distribution can be well specified and bounded.

\noindent \textbf{Red teaming:} Current red teaming efforts involve humans searching through the space of possible inputs for ones that elicit failures, or using automated techniques like GCG \citep{zou2023universal}. Interpretability tools could provide significant additional information that guides this search \citep{carter2019activation,goh2021multimodal,mu2020compositional}. \citet{ziegler2022adversarial} find that highlighting tokens based on their saliency could reduce time taken to find adversarial examples by 35\%. Tools that provide more detailed and targeted information may improve upon this significantly. 

Understanding vs control: Tools to guide red teamers likely just need understanding, but if we want to directly create adversarial examples that is a form of control. This could also look like altering the model’s internals at inference time, as in Latent Adversarial Training \citep{casper2024defending}. For example. \citet{arditi2024refusal} showed that models can be jailbroken by projecting out one direction, which was then used in \citet{yu2024robust} to make adversarially robust models more cheaply.

Confidence: Since the purpose of interpretability is hypothesis generation and guiding the search, low confidence techniques are sufficient. Using interpretability to argue that the model cannot be red-teamed would require high confidence techniques, but would be considered an alignment evaluation.

Concept vs algorithm discovery: While concept discovery is helpful \citep{carter2019activation}, we expect that algorithm discovery will be more important, because for red teaming it is important to understand how concepts relate to each other and to the model's output, so that we can find out how to confuse the model, or engage its mechanisms in unexpected ways.

Supervision: Unsupervised techniques are best for hypothesis generation and exploratory techniques, but supervised techniques are still useful when analysing known concepts such as how refusal is represented \citep{arditi2024refusal}.

Generality of conclusions: The techniques just need to work in the domain being red-teamed, and so can generally be task-specific. However, it is often desirable to red-team in a wide range of domains, in which case general techniques can be more efficient.

\noindent \textbf{Amplified Oversight:} Interpretability techniques could be used to annotate AI reasoning traces with additional information that enables rater interrogation: for example, we could identify and highlight the claims the assistant is least confident in. Or two AIs debating each other (see Section~\ref{sec:ao-methods}) could use interpretability tools to argue that their opponent is lying or misleading the judge.

Confidence: Rather than confidence, per se, it seems important that these techniques are somewhat adversarially robust, so that the interpreted model doesn’t just learn how to break the tools. So long as they give some signal that won’t be reward hacked (see Section~\ref{sec:sources-misalignment}), even if it’s not high confidence, they may be useful. A lack of robustness is particularly problematic for this use case, as actively training against interpretability methods makes them untrustworthy for assurance.

Concept vs algorithm discovery: Concept discovery seems most important here, though algorithm discovery may be important for nuanced cases.

Supervision: Impact here probably comes from methods that are relatively cheap per data point (since we have to apply them to many data points) and that cover a wide range of domains or topics (since many applications involve very wide distributions and we don’t want to need to rely too much on OOD generalization) so we need techniques that aren’t too specialised, suggesting unsupervised may be more useful.

Generality of conclusions: This just needs to work in the domain that the model is being trained or used, but this is often quite broad.

\subsubsection{Interpretability techniques}

\begin{table}[t]
\small  
\renewcommand\theadalign{l}
\renewcommand\theadfont{\small}
\setlength{\tabcolsep}{4pt}
\begin{tabular}{|p{2.5cm}|p{2.2cm}|l|l|l|l|l|}  
\hline
\thead[l]{Technique} & \thead[l]{Understanding\\v Control} & \thead[l]{Confidence} & \thead[l]{Concept\\v Algorithm} & \thead[l]{(Un)supervised?} & \thead[l]{How\\specific?} & \thead[l]{Scalability} \\
\hline
Probing & Understanding & Low & Concept & Supervised & Specific-ish & Cheap \\
\hline
Dictionary\newline learning & Both & Low & Concept & Unsupervised & General$^*$ & Expensive \\
\hline
Steering\newline vectors & Control & Low & Concept & Supervised & Specific-ish & Cheap \\
\hline
Training data\newline attribution & Understanding & Low & Concept & Unsupervised & General$^*$ & Expensive \\
\hline
Auto-interp & Understanding & Low & Concept & Unsupervised & General$^*$ & Cheap \\
\hline
Component\newline Attribution & Both & Medium & Concept & Complicated & Specific & Cheap \\
\hline
Circuit analysis\newline (causal) & Understanding & Medium & Algorithm & Complicated & Specific & Expensive \\
\hline
\end{tabular}
\caption{General$^*$ refers to techniques that are specific to a data distribution, but where this can just be the training data distribution.}
\label{tab:interp-techniques}
\end{table}

The space of interpretability techniques is very large. In this section we do not try to give comprehensive coverage. Instead we focus on a few approaches that we think are particularly relevant to the above goals, and discuss where they fall on our axes as well as their scalability to frontier models (see \Cref{tab:interp-techniques}).

Several techniques are motivated by the linear representation hypothesis (LRH) \citep{mikolov2013efficient, arora2018linear, park2024geometry, nanda2023emergent, elhage2022superposition}. In its weak form \citep{mikolov2013efficient} it states that concepts in a model correspond to linear directions in activation space, and in the strong form \citep{elhage2022superposition} it states further that any given activation is a sparse linear combination of these concepts. This hypothesis has a fair amount of empirical evidence \citep{elhage2022superposition,gurnee2023finding,olah2020zoom,park2023linear,nanda2023emergent,mikolov2013distributed}, but is not yet proven \citep{smith2024simple}.

\noindent \textbf{Probing:} Probing \citep{belinkov2022probing} is a classic supervised technique for understanding models by detecting concepts. A researcher takes a labelled dataset of model inputs and labels, and then trains a classifier on model activations to predict the labels. The simplest and most principled approach (given LRH) is a linear probe \citep{alain2016understanding}, i.e. logistic regression on model activations, but there are many more complex approaches like MLP probes. The standard form of probing is purely correlational, and does not causally verify that the probe has found a concept the model is actually using, so it’s a fairly low confidence technique (though more causal approaches have been explored \citep{elazar2021amnesic}). Notably, probing can be extremely cheap for both training and inference, e.g. monitoring a model with a linear probe can be much cheaper than using another LLM to analyse its output. In some sense, the main LLM is doing most of the work of processing the input into more useful representations already, and probes can extract these very cheaply, rather than needing to start from text. Probes are typically restricted to the domain they were trained on, though this can be broad, and sometimes they generalise surprisingly well \citep{marks2023geometry}.

\noindent \textbf{Dictionary learning:} When we apply sparse coding \citep{zhang2015survey,arora2018linear} techniques to model activations to break them down into sparse linear combinations of vectors from a fixed dictionary. Sparse coding is an unsupervised method of finding a dictionary (list of vectors), and mapping each activation to a sparse vector of coefficients for the dictionary that produce a good reconstruction to the activation. This is unsupervised in the sense that it only optimises for sparsity, in the hopes that this will find units corresponding to interpretable concepts, rather than baking in knowledge of what concepts to look for.

A particularly popular approach to sparse coding for interpretability is sparse autoencoders \citep{yun2021transformer, bricken2023monosemanticity, cunningham2023sparse, gao2024scaling, rajamanoharan2024improving, templeton2024scaling} (SAEs), where a single hidden layer autoencoder learns to reconstruct activations, with a sparsity penalty such as L1 on its latent activations. A high fraction of SAE units (aka latents or features) have been found to be interpretable \citep{bricken2023monosemanticity, rajamanoharan2024improving}, but not all. Sparse autoencoders have been shown to scale to frontier models \citep{templeton2024scaling, gao2024scaling} albeit at high cost, and trained on only a single layer. Other dictionary learning methods may be cheaper.

Dictionary learning is notable in that, rather than detecting a specific concept, it tries to find a complete decomposition of the activations into a sparse combination of vectors. A complete decomposition into interpretable concepts could enable some very high confidence conclusions \citep{olah2023interpretability}. Unfortunately, this is only an approximation, and there is some reconstruction error, which can correspond to concepts it didn’t have capacity to learn \citep{bussmann2024stitching} or error on concepts whose vectors were learned incorrectly \citep{rajamanoharan2024improving}. Further, the units are not always interpretable, and our interpretations may be incomplete \citep{anthropic2024sensitivity,huang2023rigorously}. More nuanced interpretability methods will likely be needed, such as an LLM interpretability agent~\citep{shaham2024multimodal} or using algorithm discovery techniques to contextualise subtle differences in features by connecting them to early features. Addressing these issues or bounding their effect seems vital for high confidence conclusions. 

Dictionary learning can naturally be used for understanding, but can also be used for control, by boosting or suppressing an interpretable unit to change the expression of that concept \citep{templeton2024scaling,farrell2024applying,goodfire2024llama,karvonen2024sieve}.

Dictionary learning will only identify concepts specific to the data distribution the dictionary is trained on. However, the data distribution can be very general, e.g. all data the model was ever trained on. Since every sophisticated concept in the model must have been useful for some training data point (otherwise it is unclear why the concept would have developed at all), dictionary learning could in principle find every concept in the model, assuming the strong linear hypothesis holds.

Another advantage of dictionary learning is that it’s very granular: the units of analysis are just directions in activation space, which are very localised, just like studying a single neuron. In contrast, essentially all behavioural approaches treat the unit of analysis as the entire model, extremely coarse. We speculate that coarse approaches to interpretability are more likely to have compelling behavioural analogues, while for localised approaches it may be harder to do anything remotely similar with behavioural approaches. So if there are uniquely impactful applications of interpretability, they may be more likely to occur with localised techniques.

\noindent \textbf{Steering vectors:} Steering vectors \citep{turner2023activation,larsen2016autoencoding,white2016sampling,li2024inference,zou2023representation} are a supervised technique for control, also inspired by the linear representation hypothesis. There are many variants, but in the simplest form a dataset of activations is collected with some concept (e.g. positive sentiment), a dataset of activations is collected without, and we take the mean difference to find the steering vector \citep{turner2023activation}. This can then be added to the model’s activations on another prompt to introduce that concept, typically with a coefficient found by a hyperparameter sweep. Steering vectors often seem to generalise out of the specific context that activations were taken \citep{zou2023representation}, but this is not yet well understood, and we cannot have high confidence that the technique will always work as expected.

Training data attribution (TDA): TDA \citep{koh2017understanding, pruthi2020estimating, grosse2023studying,park2023TRAK} tries to find which training datapoints most influenced the model to produce that output. These methods approximate the effect of “removing a sample from the training corpus”. While the model might learn the same information from other samples if a sample was removed, TDA methods are often used for a more practical question of “for a given sample z, which training datapoints increase the likelihood of z under the model parameters?”. 

Influence functions, in their original form proposed in \citet{koh2017understanding}, is often prohibitively expensive to compute for LLMs. As an alternative, methods that approximate Fisher Information Matrix \citep{ly2017tutorial,kunstner2019limitations} have been shown to scale to near frontier models \citep{grosse2023studying, chang2024scalable}. While they still remain expensive to run on large pretraining datasets, \citet{chang2024scalable} were able to scale their TrackStar algorithm to the entire C4 corpus \citep{raffel2020exploring} (160B tokens).

This is an unsupervised technique that can surface unexpected connections regarding which data contributed to different parts of the output. TDA has enabled practical applications such as removing noisy datapoints \citep{pruthi2020estimating}, tracing specific behaviors of models \citep{ruis2024procedural} and selecting datapoints for targeted fine-tuning \citep{xia2024less} (and DsDM \citep{engstrom2024dsdm}). TDA seems particularly useful with fine-tuning, which involves far smaller datasets than pre-training, and so is easy to re-finetune on edited data. TDA seems particularly useful for debugging unexpected failures, as it may point us to unexpected hypotheses.

TDA as it is currently used has several limitations. It surfaces data points that have influence, but doesn’t distinguish the type of influence: style, factual knowledge, grammar, reasoning, etc. Work thus far has focused on attribution to high-level metrics like average next token loss over an entire sample, and it is unclear how well it works on more refined metrics such as loss for a specific key token, or metrics based on model internals like why a neuron fires, but we believe these could be promising directions of future work.

\noindent \textbf{Auto-interp:} \citep{bills2023language,paulo2024automatically,choi2024scaling,shaham2024multimodal} Lots of tasks performed by interpretability researchers require flexible intelligence, but are fairly bounded. As LLMs become more capable they can do these tasks cheaper and faster, and so it is valuable to improve our ability to automate interpretability research, so we can accelerate and scale up current work.

An important example is interpreting a direction in a model’s activation space (e.g. a neuron or sparse autoencoder latent) - projecting the model’s activation vector onto this direction gives a scalar activation on each input. This is typically interpreted by looking at its max activating dataset examples, the inputs that most activate it from a sample of the training data, and look for a pattern \citep{olah2020zoom}. But spotting patterns in a sequence of data is a task that can be given to an LLM. There are many ways to refine this, and we expect to see more in future e.g. the LLM generating revisions \citep{bills2023language} or acting iteratively as an agent \citep{shaham2024multimodal}. This can also be automatically scored by having the LLM predict future activations from just its explanation \citep{bills2023language,paulo2024automatically,choi2024scaling}. These explanations cannot yet be relied upon\cite{huang2023rigorously}, and maximum activating examples can be misleading \citep{bolukbasi2021interpretability}, but in practice are often useful \citep{bricken2023monosemanticity,templeton2024scaling}. And if the data distribution is broad enough, the LLM may surface patterns corresponding to unexpected concepts.

\noindent \textbf{Component attribution techniques:} There are a number of techniques for attributing a model behaviour to specific components, e.g. heads, neurons, sparse autoencoder latents, layers, etc. These are valuable for localising a model behaviour, as this often corresponds to a sparse set of components. They typically involve studying a component on a specific prompt (or prompt pair) that induces a behaviour, and performing a causal intervention on it to see the change in the model’s behaviour. By default, this just gives a conclusion about that component on that prompt. But if this is done on a large dataset of prompts, we may be able to make more general statements. Three notable families of attribution techniques are:
\begin{itemize}
    \item Ablations (aka knockout): Replacing a component’s output with something meaningless, e.g. all zeros \citep{baan2019understanding,lakretz2019emergence,bau2020understanding}, its mean value (on some distribution) \citep{wang2022interpretability} or its value on a random other input \citep{chan2022causal}, and seeing the effect on the model’s output.
    \item Activation patching \citep{zhang2023towards,heimersheim2024use} (aka causal mediation analysis \citep{vig2020investigating}, causal tracing \citep{meng2022locating}, interchange interventions \citep{geiger2021causal,geiger2023causal}): A more refined form of ablations, where we instead replace a component’s output with its value on a different prompt, allowing us to localise just the effect of the differences in the prompts, rather than everything the component did. For example, prompt 1=``I hated that movie, it was”, prompt 2=``I loved that movie, it was”, varies the sentiment of the prompt, and we can patch a component to see its effect on whether the model says a positive or negative adjective, but control for whether the component is e.g. a bias term, or tracking grammatical structure.
    \item Direct Logit Attribution \citep{nostalgebraist2020interpreting, elhage2021mathematical}: How much a component’s output directly contributes to the correct output logit, i.e. how much they boosted the correct answer over the incorrect answer, ignoring the effect it has on any other components. 
\end{itemize}

\noindent \textbf{Circuit analysis:} \citep{wang2022interpretability,chan2022causal,hanna2024does,conmy2023towards,lieberum2023does} Circuit analysis is another term for algorithm discovery \citep{olah2020zoom}. One approach to circuit analysis is to frame the model as a computational graph of components, and to use causal interventions to find a subgraph that preserves model behaviour on a specific task. This is typically done with component attribution techniques like activation patching to localise key components, or edges between them. This has been done with a range of component granularity: layers \citep{meng2022locating,variengien2023look}, attention heads \citep{wang2022interpretability}, neurons \citep{nanda2023factfinding,ferrando2024similarity}, etc. Circuit finding has been shown to work on large models like Chinchilla 70B \citep{lieberum2023does}, though is expensive and only a partial circuit was found, though approximations like attribution patching \citep{nanda2023attribution,kramar2024atp} may speed things up.

Merely finding a sparse subgraph without knowing what components mean is not that helpful, so a particularly exciting recent direction has been circuit finding with sparse autoencoder latents as components \citep{marks2024sparse, kharlapenko2025scaling, dunefsky2024transcoders}, as these are often interpretable across the entire training data distribution.

A major issue with current circuit finding is that the circuits found seem quite messy and often aren’t that faithful to the original model behaviour \citep{miller2024transformer,chan2022causal}, limiting our confidence in the technique. This indicates, perhaps, that some of a model’s behaviour is explained by a sparse set of components, but that there’s a long tail of slightly helpful components.

This is typically done on a narrow task distribution, e.g. sentences that follow a specific template, which is helpful to do activation patching. While this limits the generality of the findings, it also makes the problem more tractable: on the full data distribution most model components will be polysemantic, used for many different things, while on a narrow distribution they may only be used for one thing. If we only care about the model's behavior on this narrow distribution, then ignoring the other behaviors is helpful. In a sense it’s a supervised technique, as we choose the task and distribution, but it may uncover the model doing things in unexpected ways.

Another approach to circuit analysis is to focus on the weights~\citep{nanda2023progress, olah2020zoom, gross2024compact, cammarata2021curve, elhage2021mathematical, mcdougall2023copy}.
This treats the model as a mathematical object, consisting of a bunch of matrix multiplications and a few non-linearities, and tries to reverse-engineer this to understand how the model's operation corresponds to an interpretable algorithm. If truly successful, this approach could give very high confidence, but progress thus far has largely been restricted to small or toy models, and has been limited or mistaken in various ways \citep{goldowsky2023localizing,stander2023grokking,wu2024unifying}. Overall, we consider this approach to be extremely ambitious and unlikely to scale, so do not focus on it here, but we would be very excited to see scalable progress on this front.

%% file: 06-addressing-misalignment/stress-tests.tex
\subsection{Alignment stress tests}\label{sec:alignment-stress-tests}

Alignment stress tests are a key source of empirical feedback about our alignment proposals. We can stress test entire proposals as well as individual assumptions within them.

While alignment stress tests can take many forms, they often proceed with the following steps:
\begin{enumerate}
    \item Start with the safety argument of some alignment scheme.
    \item Identify a key property on which the argument relies.
    \item Test if the key property holds, including adversarial tests of worst-case scenarios.
    \item If tests identify counterexamples, then the alignment scheme has vulnerabilities.
\end{enumerate}

Searching in this way for counterexamples to safety cases is a key type of alignment stress test. By nature of this search, the red team is free to adversarially choose any free parameter in the safety case. For example, consider the assumption that “if we train models using reinforcement learning from human feedback (RLHF), then they will be honest.” By leaving the deployment context unspecified, the assumption is implicitly making a claim for all deployment situations. Because stress tests are studying worst-case robustness, the red team adversarially looks for any deployment situation where the assumption fails.

Alignment stress tests typically do not provide any strong guarantees. A key risk in alignment stress testing is that there exists an unknown attack method that would have uncovered failure modes if only we had used it. When doing worst-case analysis, we can give the red team unrealistic advantages to produce a more conservative test. By doing so, we hope to evaluate against stronger attacks than any we would face in the real world. Nevertheless, we must bear in mind that stress tests have more power in establishing vulnerabilities in alignment plans than they do in establishing guarantees of alignment.
\subsubsection{Motivation}
Alignment stress tests aid the overall project of AGI safety in the following ways:

\textbf{Red teaming assumptions in alignment plans.} Ideally, we would have theoretical guarantees that our AI systems are beneficial. However, the current science of deep learning is not yet there. Moreover, it is unclear whether it is even possible to develop a mathematical theory that would apply to open-ended deployment contexts, which are complex systems. Absent theoretical guarantees, stress tests adversarially look for counterexamples where alignment proposals fail. If counterexamples are found, then they signify vulnerabilities in the proposed safety measure. If counterexamples are not found even when a well-incentivized, well-resourced team is searching for them, then this is some evidence indicating robustness of the proposal.

\textbf{Assurance for deployment decisions.} Pre-deployment tests are a crucial part of the deployment process for misuse (\Cref{sec:misuse-stress-tests}). Alignment stress tests can serve a similar function in the case of misalignment, for example by adversarially testing if deception remains after applying deception mitigations.

\textbf{Model organisms of misalignment.} Although we don’t yet have AGI, we would still like to study examples of models exhibiting misaligned behavior. Alignment stress tests can create examples of misalignment~\citep{greenblatt2024alignment} and how misalignment emerges, including the inductive biases in deep neural networks towards aligned versus misaligned behavior \citep{denison2024sycophancy}. After creating examples of misalignment, we can study our ability to monitor and mitigate misaligned behavior. For example, we can use deceptive models to study AI lie detectors.

\textbf{Demonstrations of risks.} In addition to its scientific benefits, demonstrating examples of misalignment is a powerful way to communicate AI risks to the world. As more stakeholders, including government leaders with diverse backgrounds, become involved in the project of AGI safety, having clear demonstrations of failure modes creates a common knowledge base of AI risk on which coordinated solutions can be built.

\subsubsection{Case studies}

\paragraph{Saliency maps.} Saliency maps are a widely-used technique for interpreting computer vision models \citep{baehrens2010explain,simonyan2013deep,shrikumar2016not,sundararajan2017axiomatic,springenberg2014striving,selvaraju2016grad,smilkov2017smoothgrad}. The goal of a saliency map is to illustrate how much each pixel of the input image affects the model’s output classification. \citet{adebayo2018sanity} stress test the assumption that visual inspection of saliency maps provides a faithful representation of what an image classifier has learned. For some saliency algorithms and some neural network layers, they find that randomizing all the weights in the layers doesn’t change the qualitative visual appearance of the saliency maps. \citet{adebayo2018sanity}’s finding demonstrates that it is unsafe to assume that these saliency algorithms contain information about what all the layers of an image classifier have learned.

\paragraph{Obfuscated gradients.} In the 2018 International Conference on Learning Representations, numerous papers used a variant of so-called “obfuscated gradients” to create defenses against white-box adversarial attacks on image classifiers. By the 2018 International Conference on Machine Learning, held just a few months later, \citet{athalye2018obfuscated} had designed an improved set of attacks that fully circumvented 6/7 methods (and partially circumvented the 1 remaining method) relying on obfuscated gradients. “Obfuscated gradients give a false sense of security,” \citet{athalye2018obfuscated} showed with their stress tests. On a meta level, this case study illustrates the importance of using strong attacks to evaluate potential alignment techniques.

\paragraph{Sleeper agents.} Variants of reinforcement learning  \citep{schulman2017proximal} are among the most popular post-training techniques for language models \citep{bai2022constitutional}. \citet{hubinger2024sleeper} stress test the assumption: “if there’s a set of inputs on which the model behaves badly, we can train away this bad behavior using standard reinforcement learning techniques.” To perform their stress test, \citet{hubinger2024sleeper} train coding models with a “backdoor behavior”: the models insert code vulnerabilities only when the input prompt says the current year is 2024. Then, \citet{hubinger2024sleeper} do standing reinforcement learning finetuning to teach the models to be helpful, honest, and harmless (HHH) \citep{askell2021general}. \citet{hubinger2024sleeper} find that even after HHH finetuning, the models still insert code vulnerabilities for input prompts with the year 2024. Thus, we can’t rely on existing HHH reinforcement learning techniques to remove secret backdoor behaviors.

\paragraph{Contrast consistent search.} \citet{burns2022how,burns2022discovering} propose that language models might represent the truth using just a few directions in their activation space. If so, we could identify all of them, and we could use them to evaluate what the model believes to be true.

Follow-up work has stress tested this hypothesis, with mixed results. On the one hand, \citet{farquhar2023challenges} present theoretical and empirical challenges to finding truth directions using unsupervised methods. On the other hand, \citet{marks2023geometry} find that truth directions can generalize to new distributions, and \citet{burger2024truth} find evidence that two dimensions might be enough to provide a generalization characterization of truth representations. It remains to be seen if methods such as \citet{burger2024truth}’s hold up to further stress tests.

\paragraph{Password-locked models.} ``Sandbagging'' refers to an AI intentionally displaying weak capabilities to pass safety evaluations. If an AI sandbags, the AI may then carry out highly capable bad actions once deployed, without us realizing that it could do so. \citet{greenblatt2024stress} test if techniques based on finetuning can elicit these sorts of hidden capabilities. To create a model organism of sandbagging, they train a more capable model to imitate a less capable model whenever a particular password is absent from the input prompt. Testing out finetuning techniques which don’t have access to the password, \citet{greenblatt2024stress} find encouraging results. Finetuning with expert demonstrations can often elicit the password-locked capability, even generalizing to different capabilities and different passwords.

\paragraph{Alignment faking.} ``Alignment faking'' involves an AI system mimicking desired values in order to mask conflicting, underlying goals which may surface later. \citet{greenblatt2024alignment} conduct stress tests with a model organism for one such behavior: instrumental goal preservation. They create scenarios where language models face a conflict between a pre-existing preference (harmlessness) and a new, hypothetical training pressure (complying with harmful requests). The models are informed that their behavior influences future training only under certain conditions (simulating monitored vs. unmonitored scenarios). Giving the model a scratchpad to observe its reasoning, \citet{greenblatt2024alignment} find models sometimes strategically comply with the harmful objective when monitored, reasoning in the scratchpad that this apparent alignment is necessary to avoid being modified by the training process. This stress test demonstrates that behavioral compliance achieved via RL training does not guarantee underlying preference alignment: instead, models might learn strategic deception motivated by goals such as preserving their current objectives.

\paragraph{Auditing games.} A key assumption underlying model evaluation is that our existing set of auditing techniques can detect misalignment. \citet{marks2025auditing} stress test this assumption by constructing an ``auditing game'': a red team deliberately trains a language model with a hidden objective (exploiting known reward model biases, termed ``reward model sycophancy'') and specifically trains it not to reveal this objective. Blinded blue teams are then tasked with auditing this model using various techniques, including behavioral tests, interpretability tools, and training data analysis. In \citet{marks2025auditing}'s tests, several teams successfully identified the hidden objective and the specific biases being exploited, but success relied heavily on access to samples of the training data; the team without data access failed.

%% file: 06-addressing-misalignment/safety-cases.tex
\subsection{Safety cases}\label{sec:alignment-safety-cases}
A safety case is an assessable argument that shows how risks have been minimised to an appropriate level. In this section, we focus on safety cases for risk from misalignment. We discuss two concrete ways of making safety cases \citep{clymer2024safety}: inability and control.

Note that not all risk-reducing actions are easily justified from a safety case perspective. For example, myopic optimization (\Cref{sec:limited-optimization}) can reduce risks by removing incentives for reward hacking that an overseer would not notice, but it is difficult to gather legible and assessable evidence for its utility in any given situation, since it is most helpful precisely in the settings where we wouldn't notice issues.

\subsubsection{Inability safety case} \label{sec:misalignment-inability-safety-case}
For particular pathways to harm, we can evaluate the AI system's capability to cause harm via that pathway (similarly to how we do for misuse, see \Cref{sec:misuse}). If we conclude the AI system is sufficiently unable to cause harm, that forms a solid safety case. Many arguments for inability to cause harm via misuse can be directly adapted to the misalignment case. For example, if a model is not capable enough at offensive cybersecurity to cause harm when used by a skilled human, it likely also cannot cause harm autonomously via its cybersecurity ability, even if misaligned.

Inability arguments can also address risks specific to misalignment. A large additional risk from misaligned AIs is that they might follow instrumental goals \citep{omohundro2008basic} such as power seeking or hiding misalignment \citep{carlsmith2022power,carlsmith2023scheming,greenblatt2024alignment}. For example a misaligned model might be incentivised to hide its misalignment during evaluation in order to ensure it will be deployed and can achieve it's misaligned goal~\citep{cotra2022without}. Following such instrumental subgoals requires the AI to have additional capabilities. For example, the AI needs to know about and understand its deployment scenario, sometimes called ``situational awareness'' \citep{ngo2024alignment,laine2024sad}; the AI needs to reason about or understand intuitively that it is beneficial to follow some instrumental subgoal \citep{meinke2024frontier}; and, the AI needs to act subtly enough to hide its misalignment or its power seeking motives from us \citep{benton2024sabotageevaluations}. If we can design evaluations for these or other relevant capabilities, we can make an inability safety case specific to an AI's ability to cause harm if misaligned \citep{balesni2024safetycases}. The benefit of these inability arguments for misalignment is that they are not risk-specific and do not require designing evaluations for all the ways a misaligned model can cause harm. The downside is that the evaluations will likely be passed earlier than risk-specific evaluations, meaning we cannot ultimately rely on these arguments alone but must combine them with risk-specific evaluations and inability arguments.

Potentially misaligned models result in additional challenges for performing dangerous capability evaluations and making inability safety cases. Specifically, if a model follows a misaligned goal, it will likely have an incentive to undermine the evaluation protocol. For example, the AI might be incentivized to hide capabilities from us, sometimes called “sandbagging” \citep{greenblatt2024stress,weij2025sandbagging}. \citet{meinke2024frontier} find that some AI models are already able to sandbag if this is incentivized in simple evaluation settings, while \citet{apollo2025claude} find that AI models sometimes express knowing that they are being evaluated, which could soon enable them to sandbag.

Sandbagging risk requires more care in setting up evaluation protocols \citep{benton2024sabotageevaluations}. Some capability elicitation techniques are likely more robust to sandbagging than others. For example, standard RL training can easily be subverted by a sandbagging agent simply by not exploring enough, sometimes called ``exploration hacking'' \citep{hubinger2023when}. However, it is significantly more difficult to subvert supervised fine-tuning, as this would require a sandbagging model to withstand or manipulate the gradient updates towards exhibiting good performance, sometimes called ``gradient hacking'' \citep{hubinger2019gradient}. This suggests benefit from using a variety of capability elicitation methods and comparing their results. AI developers should also monitor evaluation trajectories and closely examine scaling trends of dangerous capability evaluations, to detect possible signs of sandbagging.

We can reuse large parts of dangerous capability evaluations designed to assess misuse risks to make arguments about misaligned models, but we must additionally consider scheming ability and sandbagging. However, once models are capable of hide their misalignment and causing harm, we need a more focused assurance strategy for misalignment.

\subsubsection{Control}
Once models have the capabilities necessary to cause harm, we need a different approach to establishing a safety case. One promising approach involves implementing safeguards that prevent an AI from causing harm even if it was misaligned; this approach is typically called AI control~\citep{clymer2024safety, greenblatt2024control, korbak2025sketch}.

Making a control safety case requires ensuring the safety of all outputs produced by the AI system. The most straightforward way of constructing a control safety case is via a monitoring system that checks all outputs of the AI system, as described in \Cref{sec:monitoring}. Importantly, monitoring during deployment is distinct from supervision during training, which may employ similar methods but contributes to a different safety case.

To establish a robust control safety case based on monitoring, we need to detect undesired AI behavior. With current AI systems, humans can usually detect unsafe outputs easily, and the primary challenge lies in scaling supervision through automatic monitoring. Future AI systems might require techniques to enhance human oversight abilities (\Cref{sec:amplified-oversight}) use limited human supervision capacity to oversee large AI deployments~\citep{amodei2016concrete}.

Scalable control solutions might involve using other AI models to automate some of the supervision tasks. However, a control safety case using AI-automation can fail if a misaligned model, e.g., used for automatic monitoring, colludes with the deployed frontier model to undermine the control protocol. This can be mitigated by only using weaker, trusted models to supervise the outputs of larger, untrusted models~\citep{greenblatt2024control}, or by implementing specific measures to defeat collusion (\Cref{sec:judge-safety}).

A ``control evaluation''~\citep{greenblatt2024control} uses red teaming to assess a control protocol's ability to prevent unsafe AI outputs. During a control evaluation, a human red team controls all untrusted models within a scenario and attempts to induce safety violations that circumvent the control setup. For example, when testing control techniques for a coding assistant, the red team might try to modify the untrusted code model to insert backdoors into its output while the blue team might try to detect modifications the red-team made to the model~\citep{marks2025auditing} or use a weaker model to detect suspicious outputs. A reliable control safety case needs to be based on a well-designed control evaluation. For example, the control evaluation design needs to justify which models are trusted or untrusted and which affordances are given to the red and blue teams~\citep{korbak2025sketch}.

Defining undesired outcomes is crucial for establishing a control safety case. We focus on (1) avoiding catastrophic outcomes rather than building generally beneficial AI (\Cref{sec:intro}), and (2) using training targets ``with a safety margin'', such as human consultation during uncertainty (cf. \Cref{sec:safer-training-target}). This narrows the scope and simplifies defining undesired outcomes compared to the general case, though it remains challenging.

\subsubsection{Potential future safety cases}

While inability and control offer the currently most concrete approaches for constructing safety cases for potentially misaligned models, these arguments will likely prove insufficient for highly capable AI systems. This section outlines more speculative approaches to construct safety cases for future systems, which all require substantial further research.

\paragraph{Empirical arguments about model behaviour.} We might be able to make empirical arguments about how a model will behave. Reliable methods for assessing the degree of misalignment in a model are currently not available. However, future progress in alignment research and evaluation methods, particularly stress-testing alignment techniques (\Cref{sec:alignment-stress-tests}), may enable arguments about the likelihood of a particular model being misaligned. One approach involves placing the model in realistic ``honeypot'' scenarios where misaligned and aligned models would act differently. Crucially, honeypot scenarios should be difficult to distinguish from real deployments to avoiding misaligned models intentionally choosing not to engage with the honeypot.

A core challenge with this type of argument is that misaligned behavior may only arise for parts of the input distribution, which may not arise during testing. For example, \cite{greenblatt2024alignment} demonstrate that some AI systems will fake alignment when told that they will be finetuned to answer harmful requests, a fact that was presumably not known to the developers when the AI system was deployed. One approach is for honeypots to offer opportunities for achieving convergent instrumental subgoals~\citep{omohundro2008basic}, as these are likely useful regardless of the AI's misaligned goal. However, this would not handle cases in which the AI system only pursues a misaligned goal in a rare subset of inputs. Overall, achieving sufficient coverage of potential defection scenarios for a robust safety case may prove very challenging.

\paragraph{Empirical arguments about training incentives.} Concerns about misaligned AIs pursuing instrumental goals stem from the incentives present during training (\Cref{sec:misalignment-risks}). Demonstrating that a training process does not incentivize undesired instrumental goals could significantly bolster confidence in the resulting model's alignment~\citep{grosse2024three}. While this alone might not constitute a complete safety case, it could reduce reliance on other arguments, such as inability or control.

\paragraph{Understanding internal processes.} Given a sufficient understanding of the internal processes of an AI system, it would be possible to make safety case arguments that leverage this understanding. This approach would require significant progress in interpretability research (\Cref{sec:interpretability}), but could enable very powerful safety arguments. A basic example involves identifying an internal process or circuit that activates before undesirable actions, such as deception \citep{templeton2024scaling}. Monitoring this internal process could preemptively detect and prevent harmful outcomes, and result in a strong safety case, after sufficient validation.

\paragraph{(Provable) verification of model behaviour/internals.} An even stronger safety case could leverage (provable) verification of model behavior, internals, or training process \citep{dalrymple2024towards}. This could involve proving statements like ``the AI system will not do X'' across its entire input domain. While currently feasible only for small neural networks \citep{liu2021algorithms}, verifying properties of subsystems within larger AI systems might be an attainable goal.

%% file: 06-addressing-misalignment/discussion.tex
\subsection{Other areas}\label{sec:misalignment-other-areas}

Many other areas of research are relevant to misalignment, but do not feature in our approach. Typically, they are not yet mature enough that we are confident they will add value, and so as a matter of prioritization we focus on those approaches that we think will be useful in the near future, in line with our focus on ``anytime'' approaches (\Cref{sec:timelines}). We are excited for future work in these areas to demonstrate their utility and would be keen to incorporate them into our approach given sufficient evidence. Particular examples include agent foundations and science of deep learning and generalization.

%% file: 07-conclusion.tex
\section{Conclusion}

The transformative nature of AGI has the potential for both incredible benefits as well as severe harms. As a result, to build AGI responsibly, it is critical for frontier AI developers to proactively plan to mitigate severe harms.

In this paper, we outlined an approach towards technical mitigations for misuse and misalignment. For misuse, our approach centers around blocking malicious actors' access to dangerous capabilities. For misalignment, our approach has two prongs. First, we aim to better understand the reasons behind an AI system's actions, which can help us oversee it. Second, we aim to harden the environments in which AI systems act.

Many of the techniques described are still nascent and have many open research problems, and as such there is still much work to be done to mitigate severe risks. We hope that this paper can help the broader community to join us in enabling us to safely and securely access the potential benefits of AGI.